%% file: main.tex
\documentclass[10pt,twocolumn,letterpaper]{article}

\usepackage[pagenumbers]{wacv} %

\newbox\jsavebox
\newcommand{\jsubfig}[2]{%
	\sbox\jsavebox{#1}%
	\parbox[t]{\wd\jsavebox}{\centering\usebox\jsavebox\\#2}%
	}

\definecolor{lightcyan}{rgb}{0.878, 1,1} 
\definecolor{skyblue}{rgb}{0.529, 0.807, 0.92} 
\definecolor{hotpink}{rgb}{1, 0.411, 0.705} 
\definecolor{tomato}{rgb}{1, 0.388 0.278} 
\definecolor{lightpink}{rgb}{1, 0.713, 0.756} 
\definecolor{silver}{rgb}{0.752, 0.752, 0.752} 
\definecolor{lavenderblush}{rgb}{1, 0.94, 0.949} 
\definecolor{powderblue}{rgb}{0.69, 0.878, 0.9} 
\definecolor{royalblue}{rgb}{0.254, 0.411, 0.882} 
\definecolor{gray}{rgb}{0.5, 0.5, 0.5} 
\definecolor{aqua}{rgb}{0, 1, 1} 
\definecolor{lightpink}{rgb}{1, 0.714, 0.757} 
\definecolor{gray}{rgb}{0.5, 0.5, 0.5}
\definecolor{turquoise}{rgb}{0.251, 0.878, 0.816} 
\definecolor{blanchedalmond}{rgb}{1, 0.922, 0.804}
\definecolor{teal}{rgb}{0, 0.5, 0.5}
\definecolor{mintcream}{rgb}{0.961, 1, 0.980}
\definecolor{dimgray}{rgb}{0.412, 0.412, 0.412}
\definecolor{lightgray}{rgb}{0.827, 0.827, 0.827}
\definecolor{dodgerblue}{rgb}{0.118, 0.565, 1}
\definecolor{darkgreen}{rgb}{0, 0.392, 0}
\definecolor{darkviolet}{rgb}{0.58, 0, 0.827}
\definecolor{red235}{rgb}{0.922,0.129,0.129}
\definecolor{green235}{rgb}{0.129,0.922,0.129}
\definecolor{blue235}{rgb}{0.129,0.129,0.922}
\definecolor{yellow235}{rgb}{0.922,0.922,0.129}
\definecolor{lightblue}{rgb}{0.678, 0.847, 0.902} 
\definecolor{darkblue}{rgb}{0, 0, 0.545} 
\definecolor{palegreen}{rgb}{0.596,0.984,0.596}
\definecolor{kleinblue}{rgb}{0,0.184,0.655}
\definecolor{lavender}{rgb}{0.902,0.902,0.980}
\definecolor{cornflowerblue}{rgb}{0.392,0.584,0.929}
\definecolor{mintgreen}{rgb}{0.596,0.984,0.596}
\definecolor{burgundy}{rgb}{0.502,0.0,0.125}
\definecolor{beige}{rgb}{0.960,0.960,0.862}
\definecolor{slategray}{rgb}{0.439,0.502,0.447}
\definecolor{lightsteelblue}{rgb}{0.690,0.768,0.870}
\definecolor{ivory}{rgb}{1.0,1.0,0.941}
\definecolor{steelblue}{rgb}{0.274,0.510,0.706}
\definecolor{snow}{rgb}{1.0,0.98,0.98}

\usepackage[accsupp]{axessibility}
\usepackage{pgfplots}
\usepackage{pgfplotstable}
\usepackage{subcaption}
\usepackage{adjustbox}
\usepackage{tikz}
\usepackage{cuted}

\definecolor{wacvblue}{rgb}{0.21,0.49,0.74}
\usepackage[pagebackref,breaklinks,colorlinks,allcolors=wacvblue]{hyperref}

\usepackage[toc,page,titletoc]{appendix}

\def\wacvPaperID{849} %
\def\confName{WACV}
\def\confYear{2026}

\title{%
Color Bind: Exploring Color Perception in Text-to-Image Models}

\date{}
\author{{\centering Shay Shomer-Chai$^{1}$ \: Wenxuan Peng$^{2}$ \: Bharath Hariharan$^{2}$ \: Hadar Averbuch-Elor$^{2}$ 
        }
\\
{\parbox{0.9\textwidth}{\centering
$^1$Tel Aviv University \quad
        $^2$Cornell University  
       }
}
\\
\\
{\parbox{\textwidth}{\centering
\small{\url{https://tau-vailab.github.io/color-edit/}}       }
}
}

\begin{document}
\maketitle

\input{figures/teaser/teaser_SOTA/teaser_sw}

\input{sec/0_abstract}    
\input{sec/1_intro}

\input{sec/2_related}

\input{sec/3_dataset}

\input{sec/4_method}

\input{sec/5_results}

\input{sec/6_conclusion}
{
    \small
    \bibliographystyle{ieeenat_fullname}
    \bibliography{main}
}
\input{sec/X_suppl}

\end{document}

%% file: figures/teaser/teaser_SOTA/teaser_sw.tex
\begin{strip}
    \centering
    \includegraphics[width=\textwidth]{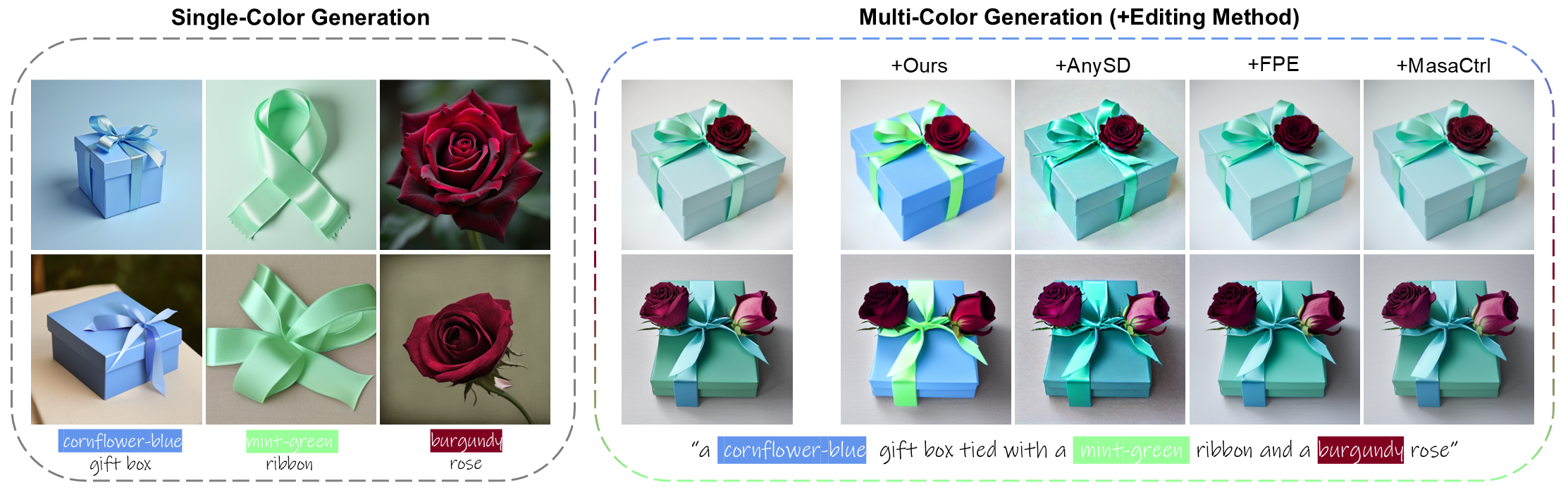}
    \vspace{-22pt}
    \captionof{figure}{
    Do text-to-image models such as FLUX~\cite{esser2024scalingrectifiedflowtransformers} (top row) and Stable-Diffusion-2.1~\cite{rombach2022highresolutionimagesynthesislatent} (bottom row) accurately perceive colors? As shown above, modern models can faithfully render even uncommon colors—\emph{e.g.}, cornflower-blue—in simple, single-object prompts. However, when faced with multi-color, multi-object prompts,  their performance degrades significantly. To evaluate this gap, we contribute a dedicated benchmark, along with an editing method that corrects these semantic errors. Our method consistently outperforms existing editing approaches, such as AnySD~\cite{yu2025anyeditmasteringunifiedhighqualityz}, FPE~\cite{liu2024understandingcrossselfattentionstable}, and MasaCtrl~\cite{Cao_2023_ICCV}, which struggle to resolve such misalignments.}
    \label{fig:teaser}
\end{strip}

%% file: sec/0_abstract.tex
\begin{abstract}
Text-to-image generation has recently seen remarkable success, granting users with the ability to create high-quality images through the use of text. However, contemporary methods face challenges in capturing the precise semantics conveyed by complex multi-object prompts. Consequently, many works have sought to mitigate such semantic misalignments, typically via inference-time schemes that modify the attention layers of the denoising networks. However, prior work has mostly utilized coarse metrics, such as the cosine similarity between text and image CLIP embeddings, %
or human evaluations, which are challenging to conduct on a larger-scale. In this work, we perform a case study on colors---a fundamental attribute commonly associated with objects in text prompts, which offer a rich test bed for rigorous evaluation. Our analysis reveals that pretrained models struggle to generate images that faithfully reflect multiple color attributes—far more so than with single-color prompts—and that neither inference-time techniques nor existing editing methods reliably resolve these semantic misalignments.  Accordingly, we introduce a dedicated image editing technique, mitigating the issue of multi-object semantic alignment for prompts containing multiple colors. We demonstrate that our approach significantly boosts performance over a wide range of metrics, considering images generated by various text-to-image diffusion-based techniques. Our code, benchmark and evaluation protocol is publicly available on our project webpage.

\end{abstract}

%% file: sec/1_intro.tex
\section{Introduction}
\label{sec:intro}

Hans Hofmann, perhaps one of the most influential painters of the 20th century, famously stated that: ``In nature, light creates the color. In the picture, color creates the light." According to Hofmann, our natural world can be described in its fullest expression through color. And indeed, colors---and particularly how they are perceived and described by humans---have been extensively studied in cognitive science~\cite{kay1978linguistic,berlin1991basic,regier2007color}. The recent rise of foundational computational models has invited parallel studies~\cite{abdou2021languagemodelsencodeperceptual,loyola2023perceptualstructureabsencegrounding} that explore whether colors are encoded within these models similarly to the human brain. 
In this work, we are interested in exploring color perception in the context of modern text-to-image models. These models are capable of turning our imagination into stunning high-quality images, perhaps analagous to Hofmann's ``picture", begging the question: \emph{Do these models perceive colors just like us?} %

Unfortunately, the answer might be no. Evidence suggests that these models struggle in faithfully generating the semantics conveyed by complex multi-subject prompts~\cite{chefer2023attendandexciteattentionbasedsemanticguidance,NEURIPS2023_0b08d733,dahary2024yourselfboundedattentionmultisubject}. Consider the results depicted in Figure \ref{fig:teaser}. While text-to-image models mostly succeed in generating a ``cornflower-blue gift box", they are challenged by the more complicated prompt that also contains additional colored entities. We observe that a case study on colors offers a unique opportunity for performing a rigorous evaluation over text-to-image generative models. By contrast to coarse CLIP-based metrics, entangled VQA-based scores~\cite{lin2024evaluatingtexttovisualgenerationimagetotext}, or small-scale human evaluations conducted by prior work, performance can be measured \emph{explicitly} using perceptual color differences.  

Our contribution is twofold.  First, we conduct a set of controlled experiments, comparing results generated using a single object (and its associated color) and results generated using multiple objects (and multiple colors). We also separately consider perceptually-close colors and distant colors, and contribute a benchmark for color understanding in image generation. Second, based on insights from this analysis, we introduce a flexible and model-agnostic approach that addresses these failures not by retraining, but by \emph{refining the generated outputs} themselves. Concretely, we formulate this as an editing framework that modifies images by accurately binding colors their intended objects while preserving all other attributes (Fig.~\ref{fig:teaser}).

Our analysis reveals that pretrained text-to-image models are significantly more successful in representing color attributes in the single subject setting. Inference-time methods that explicitly target the problem of \emph{semantic leakage}—where attributes intended for one object instead influence unrelated objects—cannot robustly bridge this gap, and also often compromise performance over single subject prompts. We also observe a clear performance gap between close and distant color pairs, demonstrating that in most cases distant color pairs yield worse performance, due to possible leakage which more severely affects performance. %

Accordingly, we introduce a training-free approach that utilizes attention-based diffusion models for editing existing images to match color specifications. In particular, we propose two objectives for guiding our optimization scheme:  an \emph{attention} loss for binding the colors to the right object and a \emph{color} loss that forces objects to have the right color. A key insight enabling our proposed attention loss is that a simplified color-less reference text prompt can provide supervision for grounding noisy attention maps. Inspired by the psychological Stroop effect~\cite{macleod1991half} which suggests that incongruent stimuli such as color words printed in differing colors (e.g. “blue” printed in red) have a stronger interference effect, our loss is built on the premise that misalignments in attribute binding leads to errors in the attention maps. Hence, we propose a loss that encourages the cross-attention maps of the full prompt to be more congruent to the simplified reference prompt, thus allowing for binding the colors to the correct image regions.

We conduct experiments that evaluate generation performance, with and without our approach. Our evaluation shows that our method consistently boosts performance, considering various text-to-image models and metrics. We also compare against alternative image editing baselines and demonstrate that prior editing models struggle at resolving the semantic misalignments of the generated images.

%% file: sec/2_related.tex
\section{Related Work}
 \noindent 
 \textbf{Misalignment Mitigation in Text-to-image Models}. 
Pretrained text-to-image diffusion models~\cite{rombach2022highresolutionimagesynthesislatent,xue2024raphael,cao2024controllable,  esser2024scalingrectifiedflowtransformers} have garnered
significant attention due to their remarkable ability to generate diverse high-quality images. 
However, a critical shortcoming of existing text-to-image models pertains to their limited capacity in representing the exact semantics, particularly in complicated multi-subject prompts. 

Prior work~\cite{chefer2023attendandexciteattentionbasedsemanticguidance,NEURIPS2023_0b08d733} has analyzed and identified typical misalignments, such as catastrophic neglect (\emph{i.e.}, failure to generate one or more mentioned subjects) and incorrect attribute binding (\emph{i.e.}, failure to match between attributes and subjects). Various mitigation schemes have been proposed. These increase semantic alignment by modifying the text embeddings~\cite{feng2022training,tunanyan2023multi} or by better aligning the cross-attention maps~\cite{chefer2023attendandexciteattentionbasedsemanticguidance,NEURIPS2023_0b08d733}. Several techniques mitigating semantic misalignments also utilize additional inputs, such as fine-grained segmentation maps~\cite{kim2023dense}, coarse spatial layouts~\cite{phung2024grounded,dahary2024yourselfboundedattentionmultisubject} or through the use of compositional generation~\cite{liu2023compositionalvisualgenerationcomposable}. In this work, we focus on the problem of incorrect attribute binding, focusing specifically on colors. We introduce a new benchmark for studying such semantic misalignments, analyzing the performance of existing text-to-image models on our proposed benchmark. 

\smallskip \noindent 
\textbf{Color Perception in Computational Models}. 
Understanding color perception and its relation with associated color names in language is a longstanding goal in cognitive science~\cite{kay1978linguistic,berlin1991basic,regier2007color}. With the rise of foundational computational models, one line of work has explored the question of whether this alignment is also reflected in these models. In particular, prior work has analyzed color perception in text-only language models. Abdou et al.~\cite{abdou2021languagemodelsencodeperceptual} propose probing techniques for measuring semantic alignment of basic color terms. Loyola et al.~\cite{loyola2023perceptualstructureabsencegrounding} extend this methodology to more abstract and complex color descriptions, such as ``NYC Taxis". In our work, we are interested in exploring color perception in multimodal text-to-image models, which are grounded in the visual world. Hence, we can measure this semantic alignment directly, without the need to compare high-dimensional language encodings with their corresponding CIELAB color values. 

Another line of work has explored color understanding for comparing multimodally-trained models with text-only models. Alper et al.~\cite{alper2023bertblindexploringeffect} demonstrated that vision-and-language models can better associate objects with colors, in comparison to text only models, like BERT~\cite{devlin2018bert}.  Grimal et al.~\cite{grimal2024tiammetricevaluating} proposed a new metric for evaluating alignment in text-to-image models, highlighting issues of color leakage and significant performance drops when prompts involve two colors. Several work~\cite{paik2021worldoctopusreportingbias, liu2022largeroctopiamplifyreporting} have studied the effect of reporting bias on language models, demonstrating that they are able to connect object and colors despite the reporting bias in text~\cite{misra2016seeinghumanreportingbias}. Monroe et al.\cite{monroe2017colorscontextpragmaticneural} conducted a study on colors in context, training classifers to predict colors given multiple colors as reference, demonstrating that color term have a broader meaning when in context with other colors. Kim et al.~\cite{kim2024attributebasedinterpretableevaluation} noted that diffusion models are less effective at modeling color-related attributes compared to shape-related attributes.  Butt et al~\cite{butt2024colorpeelcolorpromptlearning} also focus on colors in the context of text-to-image generative models. They propose an approach for optimizing tokens representing target colors. In this work, we study to what extent existing models can depict target colors, considering single and multiple objects and colors and propose an editing framework which does not require training per RGB color value.

\smallskip \noindent 
\textbf{Text-guided Image Editing}. The generative power of text-to-image diffusion models has also fueled increasing interests in text-guided image editing~\cite{hertz2022prompttopromptimageeditingcross,Cao_2023_ICCV,Patashnik_2023_ICCV,liu2024understandingcrossselfattentionstable,yu2025anyeditmasteringunifiedhighqualityz,brooks2023instructpix2pixlearningfollowimage, zhang2024magicbrushmanuallyannotateddataset}. To edit input images, existing methods typically manipulate the internal representations of the denoising networks, in particular the self and cross attention maps~\cite{tumanyan2023plug,alaluf2024cross} or employs instruction-based editing. To mitigate color misalignments in text-to-image models, we propose an editing method that builds upon the attention-based framework presented in Liu et al.~\cite{liu2024understandingcrossselfattentionstable}. We adapt it to our problem setting, as further detailed in Section \ref{sec:method}. %

%% file: sec/3_dataset.tex
\input{figures/clip_scores/clip_scores}

\section{The CompColor Benchmark}%
\label{sec:dataset}
Despite recent advances in text-to-image generation, it remains unclear whether current models can reliably follow object-level specifications—particularly when prompts involve multiple objects with distinct attributes.
For example, when asked to generate an image of ``a cornflower-blue gift box with a mint-green ribbon,” we observe that while leading models can correctly render each color when described in isolation, they often fail when multiple colored objects are described together—producing results where the box and ribbon blend into a mixed hue, such as greenish blue, instead of preserving their distinct colors (Figure~\ref{fig:teaser}). %

We aim to better understand the limitations of current image generation models in responding to complex prompts with multiple fine-grained attributes.
However, a key challenge here is the lack of quantitative metrics that can accurately capture such fine-grained semantic alignment.
Prior work typically relies on coarse metrics, such as the cosine similarity between the text prompt and the generated image in CLIP embedding space (denoted as $\text{CLIP}_\text{Sim}$), 
or the visual question-answering
capabilities of vision-and-language models (\emph{e.g.},  $\text{VQA}_\text{Score}$~\cite{lin2024evaluatingtexttovisualgenerationimagetotext}).
As shown in Fig \ref{fig:cs_vqascore_fail}, these metrics often fail in multi-object, multi-color scenarios—scoring misaligned generations as highly relevant.

To directly evaluate fine-grained semantic alignment in generated images, we introduce a benchmark composed of complex, multi-object prompts with carefully specified color attributes, which we refer to as \textbf{CompColor}. We focus on color as a first diagnostic axis for compositionality: it is a commonly described visual property, closely tied to object identity, and uniquely suited for precise, perceptually grounded quantification—having been extensively studied in cognitive science and computer vision~\cite{10.1007/978-3-319-10602-1_7,abdou2021languagemodelsencodeperceptual, loyola2023perceptualstructureabsencegrounding}. We can therefore use Euclidean distance in the CIELAB color space, a perceptually uniform metric, to explicitly quantify color-object alignment in generated images~\cite{abdou2021languagemodelsencodeperceptual,10.1007/978-3-319-10602-1_7,butt2024colorpeelcolorpromptlearning}. This allows for a systematic and scalable analysis of models' compositional performance.
We next describe our benchmark construction; additional details and visualizations are provided in the supplementary material.

\smallskip
\noindent\textbf{Selecting the Color Set.}
We start by identifying a robust color set that text-to-image models can consistently recognize and generate. This ensures that our benchmark focuses on evaluating compositional ability—rather than testing the model’s ability to understand obscure or rarely used color terms. We begin with the 140 named colors from the HTML color names table\footnote{\url{https://en.wikipedia.org/wiki/Web_colors\#HTML_color_names}}. This set includes both basic colors (e.g., Red, Blue) and extended colors (e.g., Lavender, Cyan), providing comprehensive coverage. 
We evaluate each color using simple prompts (e.g., ``a \textit{{color}} colored \textit{{object}}'') across 5 objects randomly selected from our object set and 10 seeds. Colors consistently recognized by baseline text-to-image models (Stable Diffusion 1.4 and 2.1)—determined by low average color differences in CIELAB space—are retained, resulting in a final set of 35 colors.

\smallskip \noindent 
\textbf{Forming Color Pairs.}  
To evaluate color fidelity with compositional prompts, we build our benchmark by creating pairs of colors for prompts structured as ``a \textit{\{color1\}} colored \textit{\{object1\}} and a \textit{\{color2\}} colored \textit{\{object2\}}." 
A key factor in constructing color pairs is the perceptual similarity of the two colors.
On the one hand, if the two colors are similar, then the model may still succeed at a reasonable generation even if it mixes the colors.
On the other hand, similar perceptual colors may pose a challenge for methods which try to ensure that the generated objects have distinct attributes.
As such, we classify the pairs into two types: \textit{close} and \textit{distant}, based on their perceptual similarity in the CIELAB color space. \textit{Close} colors are those that appear visually similar to the human eye, low LAB distance (e.g., \colorbox{skyblue}{SkyBlue} vs. \colorbox{lightcyan}{LightCyan}), while \textit{distant} colors are distinctly different, high LAB distance (e.g., \colorbox{skyblue}{SkyBlue} vs. \colorbox{hotpink}{HotPink}). Examples of these pairings are shown in the supplementary. %
For each color in the 35-color set, we identify three \textit{close} colors and three \textit{distant} colors to form diverse pairs for evaluation.

\smallskip \noindent 
\textbf{Object Selection for Paired Prompts.}  
We follow prior works on attribute leakage \cite{chefer2023attendandexciteattentionbasedsemanticguidance,Ge_2023_ICCV,NEURIPS2023_0b08d733} and take the union of objects used in \textit{Attend-and-Excite}, \textit{Rich Text}, and \textit{SynGen}. From this union, we carefully remove: (i) Objects with dominant intrinsic colors (e.g., bananas are mostly yellow).
(ii) Objects with small or irregular surfaces (e.g., glasses only have border color). %
For each color pair, we randomly select two objects and generate five composition prompts. We also conduct experiments with prompts containing more than two colored objects, included in the supplementary.

\subsection{Evaluation Metrics}
To evaluate the performance of the model in adhering to color specifications, we introduce an explicit color evaluation pipeline. First, we segment the target object using SAM~\cite{ravi2024sam2}. To reduce noise and capture dominant colors, we quantize the colors on the segmented object by performing k-means clustering and find the cluster closest to the ground-truth color specified in the prompt.

We primarily rely on an \textbf{Accuracy} metric, which assigns a score of 1 if the LAB distance between the predicted and target color is below a perceptual threshold $\tau$, and 0 otherwise. This provides a clear, binary indicator of whether the generated object appears perceptually close to the intended color. In all experiments, we set $\tau=10$; see the supplementary material for additional details and explanations.%

To supplement this with a more fine-grained evaluation, we also compute two additional metrics:
(1) \textbf{LAB L2 Distance}, measuring Euclidean distance between the matched cluster color and the ground-truth color in the 3D CIELAB color space, a perceptually uniform color representation, similarly to~\cite{abdou2021languagemodelsencodeperceptual}.
(2) \textbf{RGB L2 Distance} Euclidean distance between the matched cluster color and the ground truth color in the RGB space, similarly to~\cite{Ge_2023_ICCV}. 
Extended results over these metrics are provided in the supplementary.

\subsection{Baselines} 
We next use our benchmark to evaluate contemporary diffusion models, separately considering pretrained models and inference-time techniques that explicitly focus on generating complicated multi-object prompts. %
\subsubsection*{Pretrained Models}
\begin{enumerate}
\item \textbf{Stable Diffusion (SD)}~\cite{rombach2022highresolutionimagesynthesislatent}: A widely used UNet-based text-to-image diffusion model. We use versions \textbf{1.4, 1.5,} and \textbf{2.1} without further modifications.
\item \textbf{FLUX}~\cite{esser2024scalingrectifiedflowtransformers} is a rectified flow based method that integrates a transformer backbone, aiming to capture global text–image context more effectively. We use FLUX-dev for evaluation.
\end{enumerate}
\subsubsection*{Inference-time Models}
\begin{enumerate}
\item \textbf{Attend-and-Excite (A\&E)}~\cite{chefer2023attendandexciteattentionbasedsemanticguidance},  optimizes the latents by encouraging to maximize the cross-attention maps values for each subject token. It is based on SD 1.4.
\item \textbf{Structred-Diffusion (StructDiff)}~\cite{feng2022training},  incorporates linguistic structures with the diffusion guidance process based on the controllable properties of manipulating cross-attention layers. It is based on SD 1.4.
\item \textbf{SynGen}~\cite{NEURIPS2023_0b08d733}  uses language-driven cross-attention losses to force the cross-attention map of the modifier to largely overlap with the cross-attention map of the noun, while remaining largely disjoint with the maps corresponding to other nouns and modifiers. It is based on SD 1.4.
\item \textbf{RichText}~\cite{Ge_2023_ICCV} enables precise color control of generated objects by enforcing a color loss on the intermediate latents masked by the object mask. It is based on SD 1.5. %
\item \textbf{Bounded-Attention (BA)}~\cite{dahary2024yourselfboundedattentionmultisubject} is a layout conditional generation model based on SD 1.5. 
\end{enumerate}

\input{tables_cvpr/benchmark-results}
\subsection{Results of Prior Art}
\label{sec:baseline-results}
We show the results from the baselines in Table~\ref{tab:baseline_comparison}.

\smallskip \noindent \textbf{Pretrained Models}. All pretrained models achieve significantly lower errors when generating single objects, in comparison to multi-object multi-color prompts. This suggests that composing a multi-object image with precise colors is indeed a challenge for image generative models, especially when the prompt specifies distant colors for the different objects.
We observe that this is because the generative model tends to switch or leak colors between the objects. For example, when asked to generate a \colorbox{hotpink}{hot-pink} colored suitcase and a \colorbox{tomato}{tomato} colored hat, it generates an \colorbox{tomato}{tomato} colored suitcase (see the leftmost example on the middle row of Figure~\ref{fig:comparison_against_reference}). 
To further investigate color leakage, we repeated our evaluation under a more lenient criterion: 
for each object, the generated color is compared against both colors specified in the prompt, and the closest one is used for scoring.
This effectively gives the model credit even if colors are swapped or leaked much across objects, thereby isolating and quantifying the impact of color leakage/swap as a common failure mode. 
Indeed, performance improves significantly (e.g., accuracy of SD 2.1 doubled from 0.26 to 0.5 over distant pairs), demonstrating that leakage is a major factor affecting performance of pretrained models; additional details are provided in the supplementary material. %

 \smallskip \noindent \textbf{Inference-time Models}. We observe that inference-time models are also challenged by our  benchmark, despite being explicitly designed to improve attribute binding for generative models. These models achieve similar performance to pretrained models, showing better results on prompts with close colors while still struggling with distant colors.
The one exception is SynGen, which significantly outperforms the other baselines on distant colors. This can be attributed to its inference-time loss that forces attention maps of attributes to align with their corresponding objects while remaining distinct from other objects. However, this strict separation exaggerates differences even when objects are required to have similar colors, leading to higher errors in the close-color setting.

Intriguingly, we find that all five inference-time models \emph{degrade} single-object color fidelity, performing worse in terms of averaged LAB distance than their respective base Stable Diffusion models (1.4 or 1.5). This suggests that 
the \emph{additional loss terms, while designed to improve compositional control, can inadvertently harm performance on simpler prompts where no composition is needed}.

These observations highlight that, despite targeted architectural changes and inference-time interventions, existing approaches still struggle to enforce reliable object-color bindings in compositional prompts. Moreover, several methods even introduce trade-offs, compromising fidelity in simple settings to improve performance in complex ones. %
This motivates the need for a flexible, model-agnostic solution that can correct color bindings post hoc, without retraining the model or compromising single-object fidelity. Next, we address this need through a training-free editing approach that complements our benchmark. Together, these contributions offer both a diagnostic framework and a corrective mechanism for improving compositional color grounding in text-to-image diffusion models.

%% file: figures/clip_scores/clip_scores.tex
\begin{figure}[t]
    \centering
    \jsubfig{\fcolorbox{blanchedalmond}{blanchedalmond}{\includegraphics[height=3.7cm]{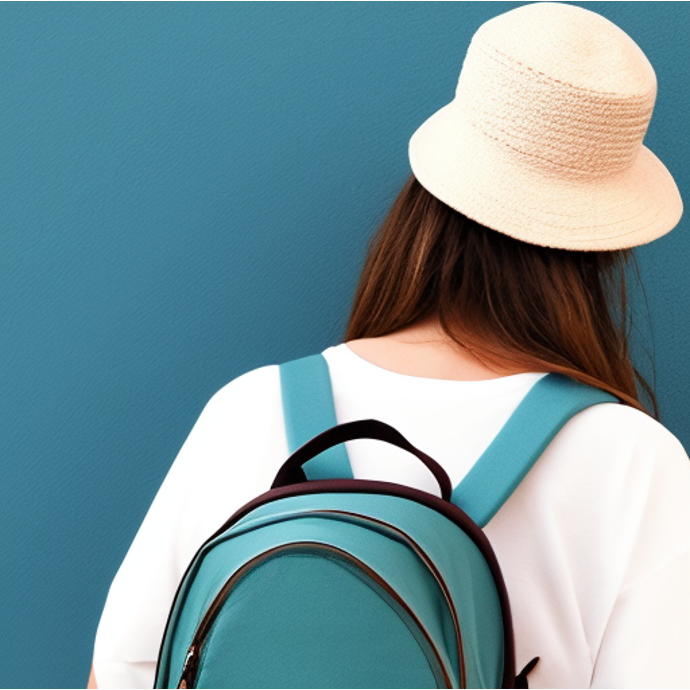}}}{\scriptsize{$\text{CLIP}_\text{Sim}(\text{\colorbox{teal}{$P_1$}})>\text{CLIP}_\text{Sim}(\text{\colorbox{blanchedalmond}{$P_2$}})$}}
    \jsubfig{\fcolorbox{dimgray}{dimgray}{\includegraphics[height=3.7cm]{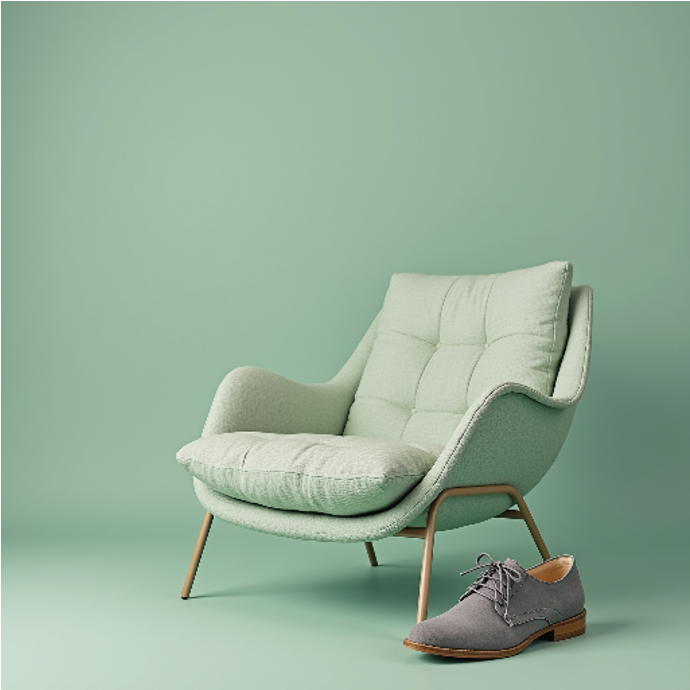}}}{\scriptsize{$\text{VQA}_\text{Score}(\text{\colorbox{mintcream}{$P_3$}})>\text{VQA}_\text{Score}(\text{\colorbox{dimgray}{$P_4$}})$}} \\
\begin{flushleft}
\footnotesize{
\text{\colorbox{teal}{$P_1$}}: ``a \colorbox{teal}{teal} colored hat and a blanched-almond colored backpack".\\
\text{\colorbox{blanchedalmond}{$P_2$}}: ``a \colorbox{blanchedalmond}{blanched-almond} colored hat and a teal colored backpack".\\
\text{\colorbox{mintcream}{$P_3$}}: ``a \colorbox{mintcream}{mint-cream} colored shoe and a mint-cream colored 
chair".\\
\text{\colorbox{dimgray}{$P_4$}}: ``a \colorbox{dimgray}{dim-gray} colored shoe and a mint-cream colored chair".}
\end{flushleft}
\vspace{-15pt}
 \caption{Existing quantitative metrics, such as $\text{CLIP}_\text{Sim}$ and $\text{VQA}_\text{Score}$ depicted above, struggle to accurately capture fine-grained semantic alignment. This is illustrated by the two examples above where misaligned text prompts achieve higher scores in comparison to the text that correctly describes the image. }
 \vspace{-0.5cm}
    \label{fig:cs_vqascore_fail}
\end{figure}

%% file: tables_cvpr/benchmark-results.tex
\begin{table}[t]
\centering
\resizebox{\columnwidth}{!}{
\begin{tabular}{lccc}
\hline
\textbf{Method} & \textbf{Single} & \textbf{Pair-Close} & \textbf{Pair-Distant} \\
\hline
\multicolumn{4}{c}{\textbf{Pretrained Models}} \\
\hline
SD 1.4 & \textbf{12.03\textbackslash 81.47\textbackslash 0.51} & 16.80\textbackslash 98.07\textbackslash 0.38  & 22.02\textbackslash 126.35\textbackslash 0.28 \\
SD 1.5 & \textbf{11.91\textbackslash 79.48\textbackslash 0.53} & 16.53\textbackslash 90.71\textbackslash 0.36  & 22.54\textbackslash 127.35\textbackslash 0.30 \\
SD 2.1 & \textbf{10.74\textbackslash 71.21\textbackslash 0.61} & 18.09\textbackslash 99.11\textbackslash 0.33  & 25.38\textbackslash 135.04\textbackslash 0.26 \\
FLUX & \textbf{11.45\textbackslash 74.13\textbackslash 0.56} & 12.50\textbackslash 71.41\textbackslash 0.54  & 15.23\textbackslash 79.50\textbackslash 0.49 \\
\hline
\multicolumn{4}{c}{\textbf{Inference-time Models}} \\
\hline
 A\&E & \textbf{14.06\textbackslash 76.18\textbackslash 0.55} & 17.02\textbackslash 81.51\textbackslash 0.46  & 20.36\textbackslash 103.47\textbackslash 0.38 \\
 StructDiff & \textbf{12.94\textbackslash 73.70\textbackslash 0.58} & 14.46\textbackslash 80.01\textbackslash 0.39  & 19.79\textbackslash 101.06\textbackslash 0.39 \\
SynGen & 13.34\textbackslash 82.57\textbackslash 0.45 & 12.56\textbackslash 81.82\textbackslash 0.49  & \textbf{11.13\textbackslash 67.21\textbackslash 0.56} \\
RichText & 22.23\textbackslash 76.89\textbackslash 0.29 & \textbf{18.60\textbackslash 78.40\textbackslash 0.35}  & 21.66\textbackslash 84.99\textbackslash 0.30 \\
BA & 15.74\textbackslash 90.45\textbackslash 0.43 & \textbf{13.10\textbackslash 79.73\textbackslash 0.55}  & 21.77\textbackslash 118.13\textbackslash 0.29 \\
\hline
\end{tabular}
}
\vspace{-8pt}
\caption{Performance of prior art on the benchmark, evaluated with the LAB($\downarrow$)\textbackslash RGB($\downarrow$)\textbackslash ACC($\uparrow$) metrics, over text prompts containing a single object and color attribute (Single) and multiple objects and colors (Pair-Close and Pair-Distant, denoting whether or not the colors are visually similar).}
\vspace{-0.5cm}
\label{tab:baseline_comparison}
\end{table}

%% file: sec/4_method.tex
\input{figures/overview/overview}
\section{Method}
\label{sec:method}
\paragraph{Problem setup:}

We aim to address this problem so as to allow users to precisely control the color of the generated objects.
For the sake of generality and to allow our approach to be relevant for future generative models as well, we frame our task as an \emph{editing} task: the input to our system is a prompt $P$ with color attributes (e.g., ``A teal cup and a pink spoon") and a source image $I_s$ with the right objects but potentially the wrong colors.
This also allows us to edit real images instead of generated ones.
To further enable precise control, we assume that the colors come with specified RGB values (e.g., ``teal" = (0,128,128) ).
Our goal is to edit the image $I_s$ so that it faithfully reflects the specified color while preserving its overall appearance and structure and still producing a high quality image; see Figure \ref{fig:overview}.

\subsection{Preliminaries}
We begin by providing brief background about text-to-image diffusion models, which will form the backbone of our approach.
We use text-to-image diffusion models for generating an initial image. That is, given a prompt $P$, these diffusion models generate an image $I$ that matches the prompt.
Modern models operate in the latent space of an autoencoder (typically a fixed VQ-VAE~\cite{rombach2022highresolutionimagesynthesislatent}): they generate a latent tensor $Z$ that can be decoded into $I$.

Diffusion models generate in an iterative manner by gradually denoising a latent tensor $Z_T$ sampled from a Gaussian prior, thus producing a sequence of intermediate latents $Z_{T-1}, \ldots, Z_0$.
This sequence is typically thought of as the ``backward'' counterpart of a forward diffusion process that gradually adds noise to an image.
Each step of this sequence utilizes a neural network $\theta$ that takes the time step $t$, $Z_{t}$ and the prompt $P$ and predicts the noise $\epsilon(Z_{t}, t, P)$.

For the exposition that follows, it will be important that this neural network conditions on the prompt $P$ through a cross-attention mechanism: every spatial location in $Z_t$ attends to tokens in the text.
Thus, for any token in the text, we can extract a heatmap showing which image locations attend to---and are influenced by---that token.
Additionally, this neural network also uses self-attention layers between the image locations themselves.

\subsection{Proposed Approach}
We wish to use off-the-shelf text-to-image models to address the task of editing object colors posed above without any further training.
This requires us to solve three challenges.
First, we want the colors to be matched to the right objects: for example, in the prompt ``a teal cup and a pink spoon", we want the color ``pink" to bind to the spoon and not to the cup or the background.
Second, we want the color to match the provided RGB values.
Importantly, we want to make these edits while preserving the structure of the image to the extent possible.
We address these challenges below.

\subsubsection{Binding colors to the right object}
\label{sec:color-bind}
As previously discussed, modern generative models struggle with binding the colors to the right object.
This corroborates past work that has demonstrated this binding problem for general attributes object~\cite{chefer2023attendandexciteattentionbasedsemanticguidance,NEURIPS2023_0b08d733}.
The reason for this difficulty is apparent from the cross-attention maps (see Figure~\ref{fig:overview}): the color tokens (e.g., ``pink") attend to the wrong parts of the image (in this case the flowers instead of the spoon).
Thus, intuitively we can ensure proper binding of colors/attributes to the right objects if we ``correct" these cross-attention maps. But how do we make this correction?

We propose an approach that constructs pseudo-ground-truth for the cross attention maps, different from the contrastive loss proposed by prior work~\cite{NEURIPS2023_0b08d733} which can degrade performance when objects are close or share similar attributes (as shown in Section~\ref{sec:dataset}).
We take inspiration from \emph{Stroop probing}~\cite{alper2023bertblindexploringeffect}, which is an approach for probing language representations.
Stroop probing analyzes language models by comparing the representations of a reference prompt (e.g., "a photo of a banana") to a prompt with colors specified (e,g, "a photo of a blue banana").
Here we take inspiration from this idea and propose to compare the \emph{cross-attention maps} of our prompt with color (e.g., "a photo of a blue banana") with the cross-attention maps of a \emph{simplified prompt} (" a photo of a banana").
Ideally, we would want the cross-attention maps of "blue" and "banana" from the former to match the cross-attention maps of "banana" from the latter.
Thus, the cross-attention maps from the simplified prompt provide \emph{pseudo-ground-truth} cross-attention maps for all the objects and attributes.

Concretely, we construct our simplified prompt $P_{simp}$ by stripping colors from our original prompt. We then perform DDIM inversion using the provided image and the simplified prompts to obtain a latent $Z_T$, and then perform backward diffusion to  recover intermediate cross attention maps corresponding to the simplified prompt.
Denote these pseudo-GT cross-attention map from the simplified prompt as \(A^{simp}_{object}\). 

Then to edit the image, we start from the same latent $Z_T$, but now perform the backward diffusion with the full prompt. 
Denote the cross-attention maps of the full-prompt process as \(A^{full}_{color}\) and \(A^{full}_{object}\).
Then, from the first iteration of the backward process, we force the color cross-attention map \(A^{full}_{color}\) and object cross-attention \(A^{full}_{object}\) to be as congruent as possible to the object's pseudo-GT cross-attention map \(A^{simp}_{object}\) extracted on the first branch using \(P_{simp}\); see Figure \ref{fig:overview}. 
We use a Jensen–Shannon divergence ($\text{JSD}$), a symmetric version of Kullback-Leibler divergence, for defining a loss over each element as follows:
\begin{align}
    L_{Attention} = \sum_{i=1}^{2} \text{JSD}(A^{simp}_{{object }\,i},A^{full}_{{object }\,i})\\ + \text{JSD}(A^{simp}_{{object }\,i},A^{full}_{{color }\,i})
\end{align}
We use this \emph{Stroop attention loss} to update the latent \(J_{AL}\) times in every iteration of the diffusion process: %
\begin{align}
     Z_t = Z_t - \gamma_{AL}\cdot\frac{dL_{Attention}}{dZ_t} 
\end{align}
where $\gamma_{AL}$ is a predefined balancing coefficient. 
This loss encourages the cross-attention maps constructed from the full prompt to be more congruent to those of the simplified prompt without any color specifications, reducing the discrepancy between their representations.

\subsubsection{Getting the right colors}
\label{sec:getting-the-right-colors}
Apart from binding the colors to the right objects, a second challenge is getting the precise color correct.
This is important because color is a fine-grained attribute with many nuances.
We have as input precise color values in the form of RGB tuples for each object.

To force each object to have the right color, we introduce a color loss.
The color loss takes the current latents \(Z_{t}\) and decodes them into intermediate noisy image (see Figure \ref{fig:overview}).
We then mask this decoded image using \(A^{full}_{color}\) and calculate the \(L_{2}\) distance between the average RGB value of the relevant object and the target RGB value, denoted as $object_{RGB}$:

\begin{align}
        L_{Color} = MSE(\frac{\sum_{p}A^{full}_{color}\cdot VAE(Z_t)}{\sum_{p}A^{full}_{color}},object_{RGB})
\end{align}

We optimize the current latents using the above loss as before:
\begin{align}
     Z_t = Z_t - \lambda_{CL}\cdot{M^{simp}_{object}}\cdot\frac{dL_{Color}}{dZ_t} 
\end{align}
where $\lambda_{CL}$ is a predefined balancing coefficient and
\(M^{simp}_{object}\) is a segmentation map extracted for each object from an off-the-shelf segmentation model~\cite{ravi2024sam2}. In the supplementary material, we show that these masks can also be extracted from the cross-attention maps directly. %

\smallskip
Finally, we want to ensure that our modifications do not alter the structure of the original image. To this end, we adopt existing strategies from prior diffusion-based editing methods~\cite{hertz2022prompttopromptimageeditingcross,Cao_2023_ICCV,liu2024understandingcrossselfattentionstable,Patashnik_2023_ICCV}. Specifically, we replace self-attention maps with those from the simplified prompt and apply background blending during the backward process; additional details are provided in the supplementary.

%% file: figures/overview/overview.tex
\begin{figure*}[t]
    \centering
    \includegraphics[width=0.98\textwidth]{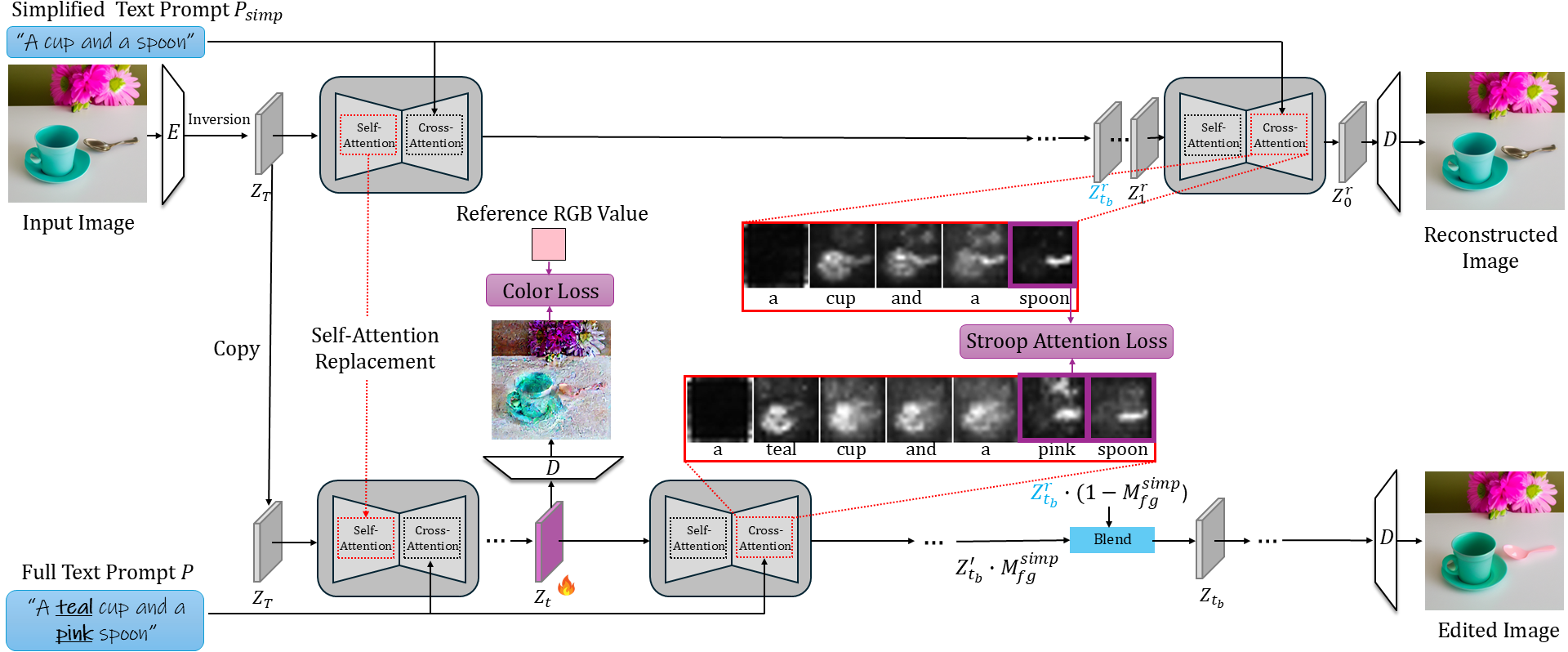}
    \caption{
    \textbf{Method architecture.} Given an input image and a target prompt $P$ containing multiple color attributes, we present an approach for editing the image to match the color specification while preserving all other attributes. Our final edited result is illustrated on the bottom right. Two objectives guide our inference-time optimization procedure: an  attention loss that binds colors to the right object (Section \ref{sec:color-bind}) and a color loss that gets the right colors (Section \ref{sec:getting-the-right-colors}); for simplicity we only demonstrate these for the ``pink spoon" in the visualization above. The notations used above are discussed in more details throughout Section \ref{sec:method}.
    }
    \vspace{-0.5cm}
    \label{fig:overview}
\end{figure*}

%% file: sec/5_results.tex
\input{tables_cvpr/editing_comparison}

\subsection{Evaluation }
\label{sec:results }

\input{figures/comparisons/comparison_against_reference}

To assess the effectiveness of our approach, we compare against several representative editing methods: (i) \emph{instruction-following} editors (\textbf{MagicBrush}~\cite{zhang2024magicbrushmanuallyannotateddataset},
\textbf{InstructPix2Pix}~\cite{brooks2023instructpix2pixlearningfollowimage}, \textbf{AnySD}~\cite{yu2025anyeditmasteringunifiedhighqualityz}) trained on paired instruction–image edits, and 
(ii) \emph{attention-manipulation} editors (\textbf{MasaCtrl}~\cite{Cao_2023_ICCV}, 
\textbf{FPE}~\cite{liu2024understandingcrossselfattentionstable}) that steer self/cross-attention at inference.
Implementation details for each method are provided in the supplementary, as well as runtime comparisons, ablations of each module (Stroop loss, color loss, simplified prompt), and results over various metrics that quantify additional aspects, such as semantic alignment and image quality. We also provide interactive visualizations and comparisons over random samples in our benchmark.

\smallskip \noindent
\textbf{Quantitative results.}
Quantitative results are reported in Table~\ref{tab:edited_vs_improvement_separate}. As can be seen in the table, our method consistently improves color accuracy across all base models and settings.  %
The largest improvements are observed when editing SynGen generations; FLUX generations also benefit substantially, especially for close-color prompts. Statistical significance (Wilcoxon signed-rank) and per-model breakdowns are reported in the supplementary.

As can be observed in the table, the improvements obtained by our approach substantially exceed those achieved by existing editing methods. Our experiments demonstrate \emph{instruction-following} editors 
are not well designed for multi-color, multi-object binding as they lack a mechanism to explicitly ensure color-object correspondence. In particular, earlier work (\emph{i.e.}, InstructPix2Pix) often fail on this task, producing images that “over-edit” the scene. This is particularly apparent for distant-color pairs, where the model cannot reliably disambiguate which object each color belongs to. \emph{Attention-manipulation} editors can better preserve the object structure and realism. However, as they don't directly enforces attribute binding, 
they frequently mis-bind colors across objects—swapping or blending them—leading to a measurable drop in Accuracy on multi-color prompts. For example, in our SynGen and FLUX comparisons, these methods maintained object layout but assigned the wrong colors to other objects or background regions.

By contrast, our approach introduces two targeted components. The \emph{Stroop attention loss} explicitly aligns color tokens with the correct object regions by leveraging a color-less pseudo-reference, directly addressing semantic leakage. The \emph{color loss} further enforces the exact RGB values on the segmented object. Together, these allow our method to outperform baselines across paradigms, delivering edits that are both accurate and structurally consistent.

\smallskip \noindent
\textbf{Qualitative results.}
Figure~\ref{fig:comparison_against_reference} shows that our edits faithfully change target colors while preserving object shape and background.
For SD~2.1, ours is the only method that performs the intended edit without collateral artifacts.
On SynGen, baselines distort object details (e.g., hat shape/ribbon) whereas our edits remain localized and faithful.
For FLUX, baselines often tint background regions, while our edits stay confined to the target objects.

%% file: tables_cvpr/editing_comparison.tex
\captionsetup[table]{font=footnotesize}
\begin{table}[t]
\scriptsize
\centering
\resizebox{\columnwidth}{!}{
\begin{tabular}{l|cccccc}
\toprule
\textbf{Method} & \textbf{Ours}& \textbf{AnySD} & \textbf{MagicBrush} & \textbf{InstructPix2Pix} & \textbf{FPE} & \textbf{MasaCtrl} \\
\midrule
\multicolumn{7}{c}{\textbf{Close}} \\
\midrule
SD 1.4       &  \textbf{59.00 \textcolor{darkgreen}{21.00\%$\uparrow$}} & 41.95 \textcolor{darkgreen}{3.41\%$\uparrow$} & 51.12 \textcolor{darkgreen}{10.11\%$\uparrow$} & 38.83 \textcolor{darkgreen}{0.00\%$\uparrow$} &  41.43 \textcolor{darkgreen}{2.86\%$\uparrow$} &  38.54 \textcolor{darkgreen}{0.98\%$\uparrow$} \\
SD 1.5        &  \textbf{54.44 \textcolor{darkgreen}{18.33\%$\uparrow$}} & 34.07 \textcolor{red}{-1.1\%$\downarrow$} & 47.06 \textcolor{darkgreen}{10.46\%$\uparrow$} & 39.41 \textcolor{darkgreen}{3.53\%$\uparrow$} &  37.39 \textcolor{darkgreen}{2.63\%$\uparrow$} &  33.33 \textcolor{red}{-1.64\%$\downarrow$}\\
SD 2.1     &  \textbf{55.36 \textcolor{darkgreen}{22.32\%$\uparrow$}}  & 35.98 \textcolor{darkgreen}{2.93\%$\uparrow$} & 42.19 \textcolor{darkgreen}{11.46\%$\uparrow$} & 37.50 \textcolor{darkgreen}{4.46\%$\uparrow$} & 35.98 \textcolor{darkgreen}{2.93\%$\uparrow$} & 30.00 \textcolor{red}{-1.74\%$\downarrow$} \\
FLUX     &  \textbf{69.65 \textcolor{darkgreen}{15.97\%$\uparrow$}}  & 57.42 \textcolor{darkgreen}{3.87\%$\uparrow$} & 53.82 \textcolor{darkgreen}{1.91\%$\uparrow$} & 36.18 \textcolor{red}{-18.43\%$\downarrow$} & 60.58 \textcolor{darkgreen}{7.05\%$\uparrow$} & 56.03 \textcolor{darkgreen}{1.95\%$\uparrow$}  \\
A\&E        &  \textbf{60.73 \textcolor{darkgreen}{14.80\%$\uparrow$}} & 44.14 \textcolor{red}{-0.3\%$\downarrow$} & 53.91 \textcolor{darkgreen}{11.33\%$\uparrow$} & 39.08 \textcolor{red}{-5.28\%$\downarrow$} & 45.80 \textcolor{darkgreen}{1.74\%$\uparrow$} &  41.69 \textcolor{red}{-3.93\%$\downarrow$} \\
StructDiff     &  \textbf{62.56 \textcolor{darkgreen}{23.08\%$\uparrow$}} & 43.69 \textcolor{darkgreen}{3.40\%$\uparrow$} & 50.59 \textcolor{darkgreen}{13.53\%$\uparrow$} & 48.39 \textcolor{darkgreen}{6.99\%$\uparrow$} &  45.07 \textcolor{darkgreen}{6.10\%$\uparrow$} & 41.63 \textcolor{darkgreen}{2.39\%$\uparrow$} \\
SynGen     &  \textbf{71.32 \textcolor{darkgreen}{22.43\%$\uparrow$}} & 44.69 \textcolor{red}{-3.66\%$\downarrow$} & 57.28 \textcolor{darkgreen}{9.86\%$\uparrow$} & 46.12 \textcolor{red}{-1.29\%$\downarrow$} & 63.41 \textcolor{darkgreen}{15.94\%$\uparrow$}  & 56.32 \textcolor{darkgreen}{7.94\%$\uparrow$} \\
RichText &  \textbf{57.64 \textcolor{darkgreen}{22.92\%$\uparrow$}} & 45.07 \textcolor{darkgreen}{10.56\%$\uparrow$} & 55.05 \textcolor{darkgreen}{15.60\%$\uparrow$}  & 42.86 \textcolor{darkgreen}{5.26\%$\uparrow$} & 50.67 \textcolor{darkgreen}{15.33\%$\uparrow$}  & 42.67 \textcolor{darkgreen}{7.33\%$\uparrow$} \\
BA     &  \textbf{66.39 \textcolor{darkgreen}{11.48\%$\uparrow$}} & 53.39 \textcolor{red}{-2.71\%$\downarrow$} & 56.00 \textcolor{darkgreen}{0.57\%$\uparrow$} & 47.56 \textcolor{red}{-8.54\%$\downarrow$} &  57.53 \textcolor{darkgreen}{1.88\%$\uparrow$} & 55.05\textcolor{red}{-1.30\%$\downarrow$} \\

\midrule
\multicolumn{7}{c}{\textbf{Distant}} \\
\midrule
SD 1.4        &  \textbf{47.78 \textcolor{darkgreen}{19.70\%$\uparrow$}} & 31.86 \textcolor{darkgreen}{2.94\%$\uparrow$} & 43.92 \textcolor{darkgreen}{14.86\%$\uparrow$} & 18.18 \textcolor{red}{-11.52\%$\downarrow$} & 27.49 \textcolor{red}{-0.47\%$\downarrow$} &  25.49 \textcolor{red}{-2.45\%$\downarrow$} \\
SD 1.5        &  \textbf{48.43 \textcolor{darkgreen}{18.83\%$\uparrow$}} & 34.80 \textcolor{darkgreen}{5.73\%$\uparrow$} & 42.77 \textcolor{darkgreen}{14.45\%$\uparrow$} & 20.21 \textcolor{red}{-8.29\%$\downarrow$} &  27.95 \textcolor{darkgreen}{0.00\%$\uparrow$} &  27.48 \textcolor{red}{-2.25\%$\downarrow$}\\ 
SD 2.1     &  \textbf{49.63 \textcolor{darkgreen}{23.88\%$\uparrow$}}  & 30.83 \textcolor{darkgreen}{6.77\%$\uparrow$} & 40.57 \textcolor{darkgreen}{20.28\%$\uparrow$} & 18.61 \textcolor{red}{-5.63\%$\downarrow$} & 28.42 \textcolor{darkgreen}{2.88\%$\uparrow$} & 24.23 \textcolor{darkgreen}{0.38\%$\uparrow$} \\
FLUX     &  \textbf{62.70 \textcolor{darkgreen}{14.02\%$\uparrow$}}  & 55.61 \textcolor{darkgreen}{6.42\%$\uparrow$} & 47.73 \textcolor{red}{-2.27\%$\downarrow$} & 19.23 \textcolor{red}{-29.23\%$\downarrow$} &  53.74 \textcolor{darkgreen}{4.55\%$\uparrow$}  &  53.31 \textcolor{darkgreen}{3.46\%$\uparrow$} \\
A\&E        &  \textbf{61.32 \textcolor{darkgreen}{22.89\%$\uparrow$}}  & 40.37 \textcolor{darkgreen}{2.37\%$\uparrow$} & 50.78 \textcolor{darkgreen}{12.79\%$\uparrow$} & 24.03 \textcolor{red}{-12.66\%$\downarrow$} &  42.24 \textcolor{darkgreen}{3.82\%$\uparrow$}  &  36.69 \textcolor{red}{-0.78\%$\downarrow$} \\
StructDiff     &  \textbf{56.82 \textcolor{darkgreen}{17.73\%$\uparrow$}} & 37.33 \textcolor{red}{-2.22\%$\downarrow$} & 40.11 \textcolor{darkgreen}{3.95\%$\uparrow$} & 22.92 \textcolor{red}{-16.15\%$\downarrow$} & 36.40 \textcolor{red}{-2.63\%$\downarrow$} & 35.55 \textcolor{red}{-2.67\%$\downarrow$}\\
SynGen    &  \textbf{72.78 \textcolor{darkgreen}{16.57\%$\uparrow$}}  & 60.24 \textcolor{darkgreen}{4.15\%$\uparrow$} & 56.97 \textcolor{darkgreen}{1.23\%$\uparrow$} & 32.10 \textcolor{red}{-25.46\%$\downarrow$} &  67.83 \textcolor{darkgreen}{11.88\%$\uparrow$}  &  63.69 \textcolor{darkgreen}{8.93\%$\uparrow$} \\
RichText &  \textbf{53.80 \textcolor{darkgreen}{23.98\%$\uparrow$}}  & 38.15 \textcolor{darkgreen}{9.25\%$\uparrow$} & 40.91 \textcolor{darkgreen}{15.91\%$\uparrow$} & 21.74 \textcolor{red}{-4.35\%$\downarrow$} &  38.51 \textcolor{darkgreen}{9.77\%$\uparrow$} & 31.74 \textcolor{darkgreen}{1.80\%$\uparrow$} \\
BA     &  \textbf{47.47 \textcolor{darkgreen}{18.40\%$\uparrow$}}  & 35.43 \textcolor{darkgreen}{5.51\%$\uparrow$} & 38.31 \textcolor{darkgreen}{12.34\%$\uparrow$} & 16.39 \textcolor{red}{-15.38\%$\downarrow$} &  28.31 \textcolor{red}{-1.04\%$\downarrow$}&  25.40 \textcolor{red}{-3.81\%$\downarrow$} \\
\bottomrule
\end{tabular}
}
\vspace{-8pt}
\caption{Color editing evaluation, comparing our approach to AnySD, MagicBrush, InstructPix2Pix, FPE and MasaCtrl. In addition to the absolute ACC value, we report the improvement (relative to the source image) in colors (\textcolor{darkgreen}{green} denotes improvements and \textcolor{red}{red} denotes degradations). %
}
\vspace{-0.5cm}
\label{tab:edited_vs_improvement_separate}

\end{table}

%% file: figures/comparisons/comparison_against_reference.tex
\definecolor{firebrick}{rgb}{0.698, 0.1353 0.1353} 
\definecolor{tan}{rgb}{0.823, 0.705 0.549} 
\definecolor{hotpink}{rgb}{1, 0.411, 0.705} 
\definecolor{tomato}{rgb}{1, 0.388 0.278} 
\definecolor{ghostwhite}{rgb}{0.972, 0.972 1} 
\definecolor{paleturquoise}{rgb}{0.686, 0.933, 0.933} 
\definecolor{silver}{rgb}{0.752, 0.752, 0.752} 
\definecolor{lavenderblush}{rgb}{1, 0.94, 0.949} 
\definecolor{cornflowerblue}{rgb}{0.392, 0.584, 0.929} 
\definecolor{linen}{rgb}{0.98, 0.94, 0.90} 
\definecolor{powderblue}{rgb}{0.69, 0.878, 0.9} 
\definecolor{teal}{rgb}{0, 0.5, 0.5} 
\definecolor{aqua}{rgb}{0, 1, 1} 
\definecolor{blanchedalmond}{rgb}{1, 0.92, 0.8} 
\definecolor{whitesmoke}{rgb}{0.96, 0.96, 0.96} 
\definecolor{lightblue}{rgb}{0.678, 0.847, 0.9} 

\begin{figure*}[htbp]
    \raggedright
       \hspace{-0.17cm} 
      \raisebox{-0.55cm}{\rotatebox{90}{\textcolor{black}{\bfseries SD 2.1}}} %
    \begin{minipage}{0.135\textwidth}
        \centering
    \raisebox{0.1cm}{\textbf{Source}} %
        \includegraphics[width=\textwidth]{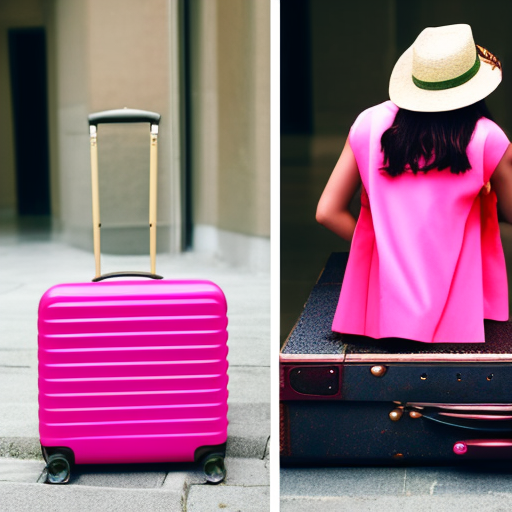} 
    \end{minipage}
    \hspace{0.1cm} 
    \begin{minipage}{0.135\textwidth}
        \centering
       \raisebox{0.1cm}{\textbf{+Ours}} %
        \includegraphics[width=\textwidth]{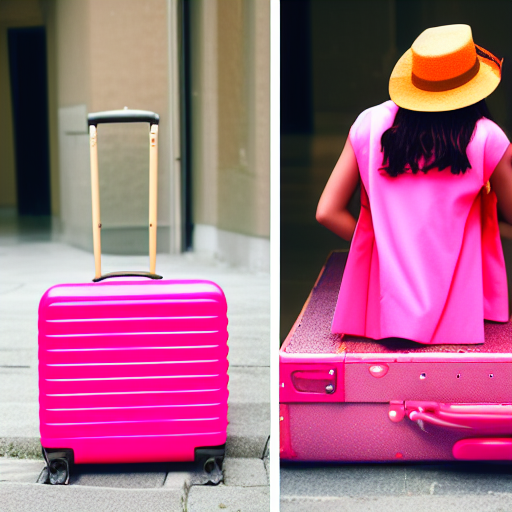}
    \end{minipage}
    \begin{minipage}{0.135\textwidth}
        \centering
        \textbf{+AnySD}
        \vspace{-0.04cm}
        \includegraphics[width=\textwidth]{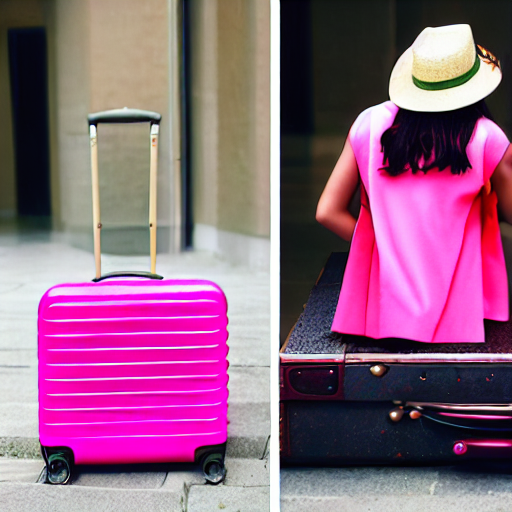}
    \end{minipage}
    \begin{minipage}{0.135\textwidth}
        \centering
        \textbf{+MagicBrush}
        \vspace{-0.04cm}
        \includegraphics[width=\textwidth]{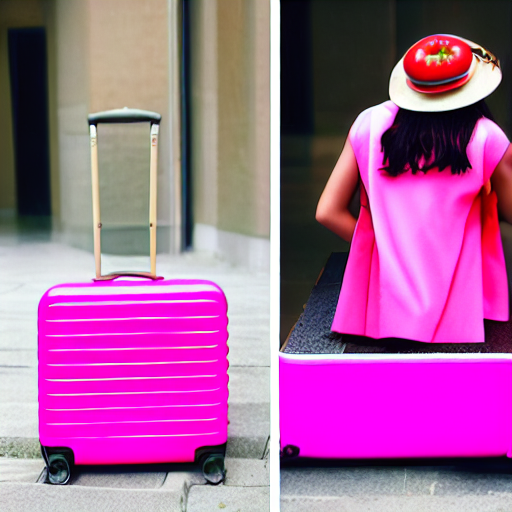}
    \end{minipage}
    \begin{minipage}{0.135\textwidth}
        \centering
        \raisebox{0.1cm}{\textbf{+InstPix2Pix}}
        \includegraphics[width=\textwidth]{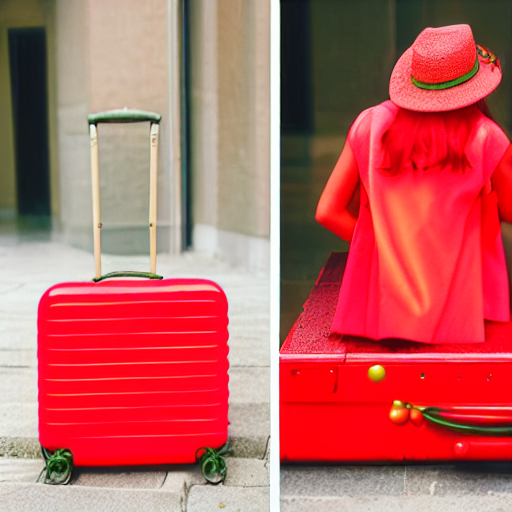}
    \end{minipage}
    \begin{minipage}{0.135\textwidth}
        \centering
        \raisebox{0.1cm}{\textbf{+FPE} }      
        \includegraphics[width=\textwidth]{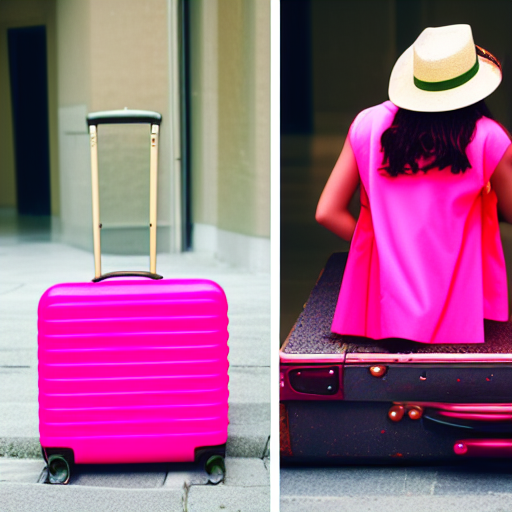}
    \end{minipage}
    \vspace{-0.01cm}
    \begin{minipage}{0.135\textwidth}
        \centering
        \raisebox{0.1cm}{\textbf{+MasaCtrl}}
        \includegraphics[width=\textwidth]{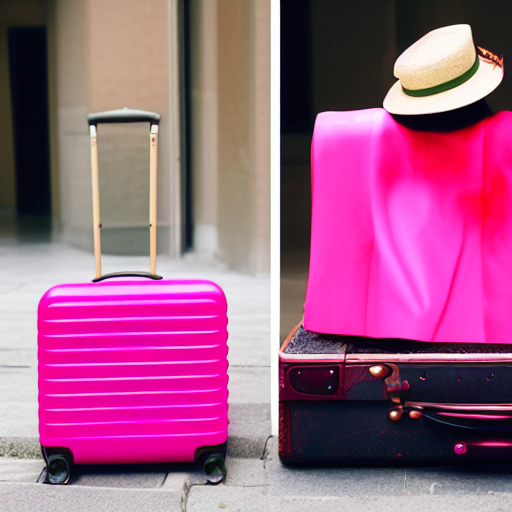}
    \end{minipage}
    \vspace{-0.015cm}
    \hspace{-0.165cm} 
    \raisebox{-0.4cm}{\rotatebox{90}{\textcolor{black}{\bfseries FLUX}}} %
    \begin{minipage}{0.135\textwidth}
        \centering
        \includegraphics[width=\textwidth]{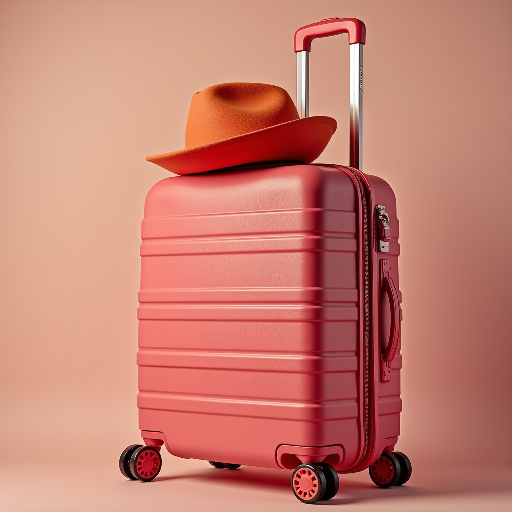} 
    \end{minipage}
        \hspace{0.1cm} 
    \begin{minipage}{0.135\textwidth}
        \centering
        \vspace{+0.05cm}
        \includegraphics[width=\textwidth]{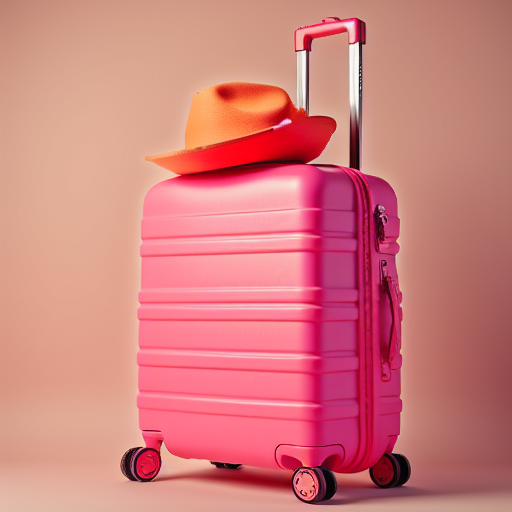} 
    \end{minipage}
    \begin{minipage}{0.135\textwidth}
        \centering
        \vspace{+0.05cm}
        \includegraphics[width=\textwidth]{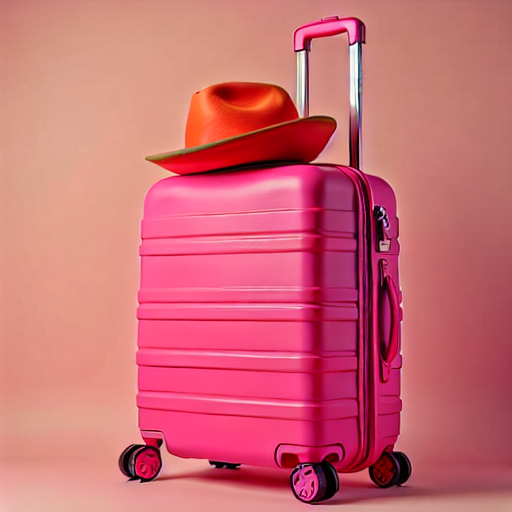}
    \end{minipage}
    \begin{minipage}{0.135\textwidth}
        \centering
        \vspace{+0.05cm}
        \includegraphics[width=\textwidth]{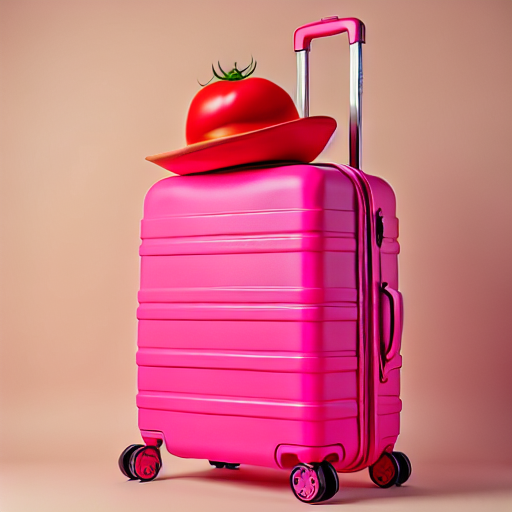}
    \end{minipage}
    \begin{minipage}{0.135\textwidth}
        \centering
        \vspace{+0.05cm}
        \includegraphics[width=\textwidth]{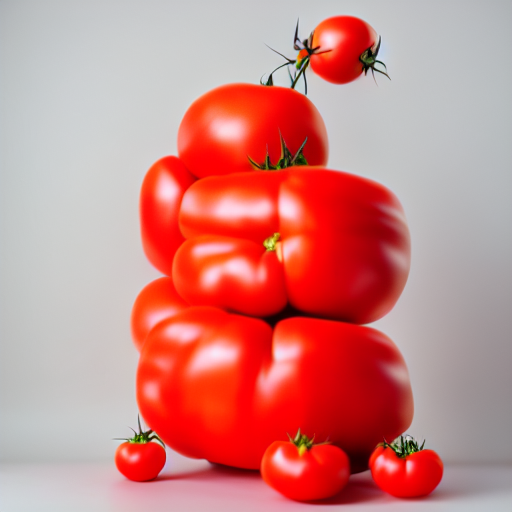}
    \end{minipage}
    \begin{minipage}{0.135\textwidth}
        \centering
        \vspace{+0.05cm}
        \includegraphics[width=\textwidth]{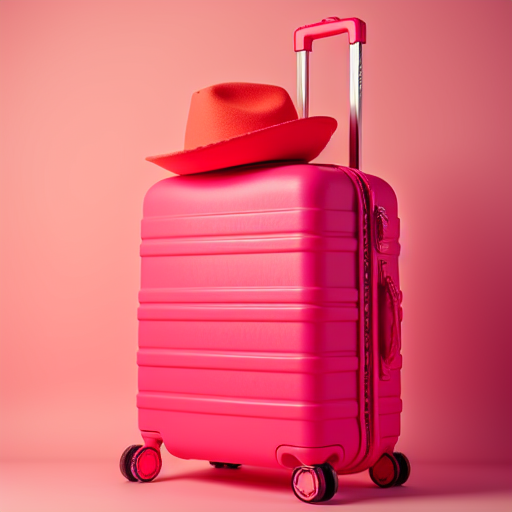} 
    \end{minipage}
    \begin{minipage}{0.135\textwidth}
        \centering
        \vspace{+0.05cm}
        \includegraphics[width=\textwidth]{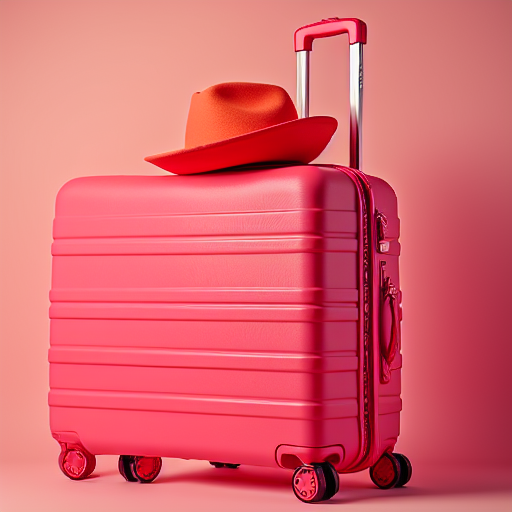}
    \end{minipage}
    \vspace{-0.015cm}
    \hspace{-0.23cm} 
      \raisebox{-0.45cm}{\rotatebox{90}{\textcolor{black}{\bfseries SynGen}}} %
    \begin{minipage}{0.135\textwidth}
        \centering
        \vspace{+0.05cm}
        \includegraphics[width=\textwidth]{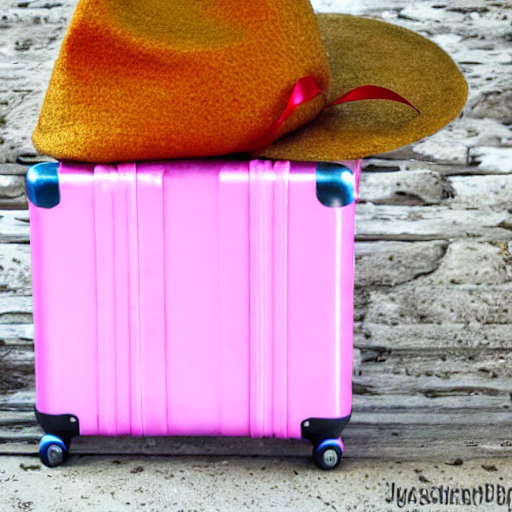}
    \end{minipage}
        \hspace{0.1cm} 
    \begin{minipage}{0.135\textwidth}
        \centering
        \vspace{+0.05cm}
        \includegraphics[width=\textwidth]{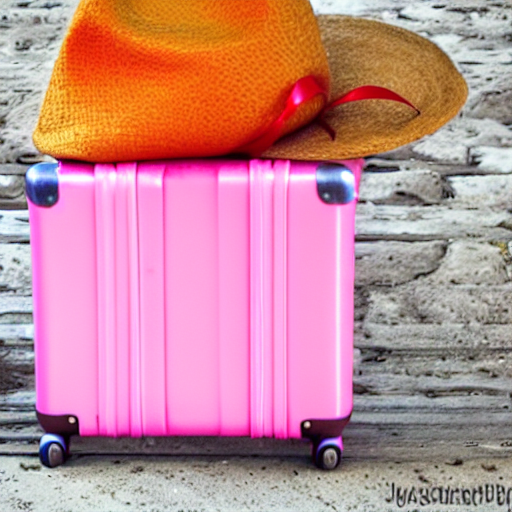}
    \end{minipage}
    \begin{minipage}{0.135\textwidth}
        \centering
        \vspace{+0.05cm}
        \includegraphics[width=\textwidth]{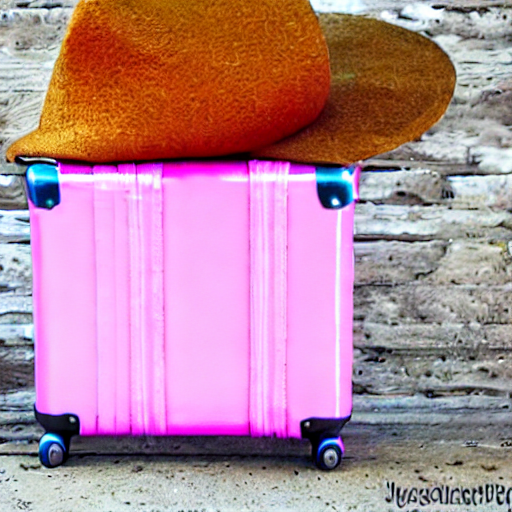}
    \end{minipage}
    \begin{minipage}{0.135\textwidth}
        \centering
        \vspace{+0.05cm}
        \includegraphics[width=\textwidth]{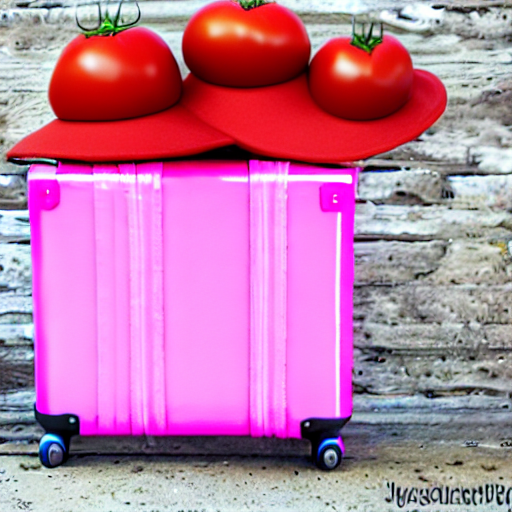}
    \end{minipage}
    \begin{minipage}{0.135\textwidth}
        \centering
        \vspace{+0.05cm}
        \includegraphics[width=\textwidth]{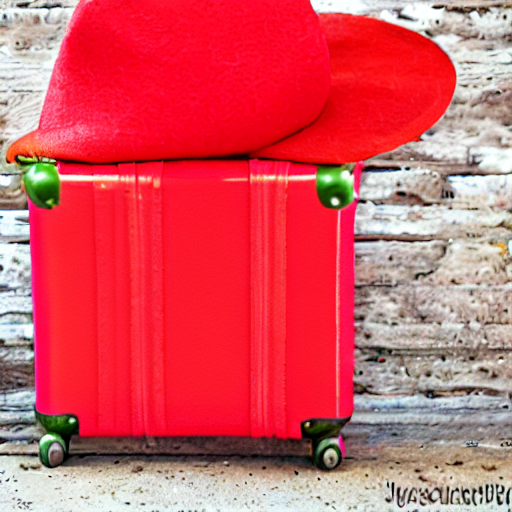}
    \end{minipage}
    \begin{minipage}{0.135\textwidth}
        \centering
        \vspace{+0.05cm}
        \includegraphics[width=\textwidth]{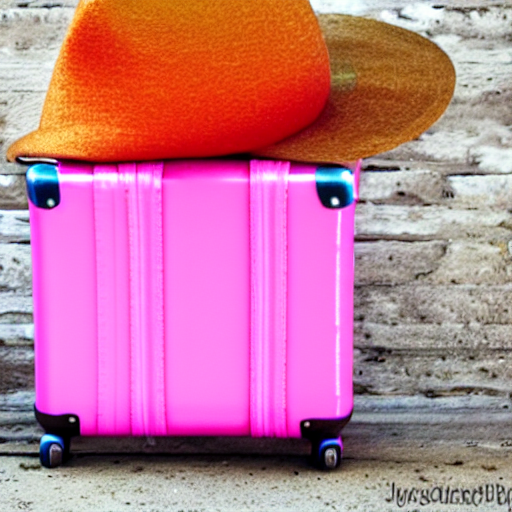}
    \end{minipage}
    \begin{minipage}{0.135\textwidth}
        \centering
        \vspace{+0.05cm}
        \includegraphics[width=\textwidth]{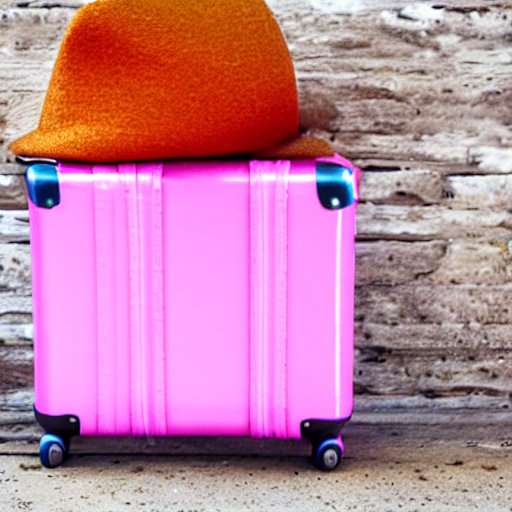}
    \end{minipage}
    \vspace{0.1cm}

    \centering
    {\footnotesize {"a \colorbox{hotpink}{\textcolor{white}{hot-pink}} colored suitcase and a \colorbox{tomato}{\textcolor{white}{tomato}} colored hat"}} %
    \vspace{-6pt}
    \caption{Images generated using the baselines, before and after being edited by our approach, AnySD~\cite{yu2025anyeditmasteringunifiedhighqualityz}, MagicBrush~\cite{zhang2024magicbrushmanuallyannotateddataset}, InstructPix2Pix~\cite{brooks2023instructpix2pixlearningfollowimage}, FPE~\cite{liu2024understandingcrossselfattentionstable} and MasaCtrl~\cite{Cao_2023_ICCV}. We illustrate close color pair in the left images. Many additional examples are provided in the supplementary material. Our method consistently outperforms existing editing approaches, we correctly edit the colors and maintain the objects' shape and background color perfectly aligned with the source image.}
    \label{fig:comparison_against_reference}
\end{figure*}

%% file: sec/6_conclusion.tex
\section{Conclusion}
In this work, we introduced a benchmark for rigorously  evaluating color perception in text-to-image models. We analyzed performance across multiple methods, conducting a series of controlled experiments over prompts containing one or two subjects paired with color attributes. Additionally, we proposed a training-free approach for accurately binding colors to their associated objects in existing images, capable of correcting color-attribute leakage. Creative design tools that require precise control over generated content could potentially benefit from our approach, which enables users to perform fine-grained, color-based edits through an intuitive, text-guided interface.

While methods addressing semantic misalignments in generative models often target more than color-based attributes, we believe that 
color provides a uniquely measurable testbed, 
hence our evaluation framework may be useful even for approaches that do not explicitly target color.
Touching back on Hofmann's quote, an interesting avenue for future work is accounting for lighting within the evaluation framework.  
Finally, we hope that our framework could be extended beyond colors, enabling controlled experiments that can carefully analyze the performance of text-to-image generative models over many additional dimensions. 

%% file: sec/X_suppl.tex
\clearpage

\appendix
\renewcommand{\thesection}{\Alph{section}} %
\renewcommand{\theequation}{\Alph{section}.\arabic{equation}} %
\renewcommand{\thefigure}{\Alph{section}.\arabic{figure}} %
\renewcommand{\thetable}{\Alph{section}.\arabic{table}} %

\twocolumn[ 
    \begin{center}
        \huge \textbf{Supplementary Material} %
    \end{center}
    \vspace{18pt} %
]
\setcounter{figure}{0} %
\setcounter{table}{0}  %
\setcounter{equation}{0}  %

\renewcommand{\theequation}{S\arabic{equation}} %
\renewcommand{\thefigure}{S\arabic{figure}} %
\renewcommand{\thetable}{S\arabic{table}} %
\setcounter{equation}{0}
\setcounter{figure}{0}
\setcounter{table}{0}

We refer readers to the interactive visualizations at the accompanied interactive\_visualizations.html which shows results over randomly selected objects and colors. In this document, we provide  details regarding our benchmark in Section \ref{sec:dataset-supp_}. Details regarding our evaluation metric are provided in in Section \ref{sec:Evaluation Metric}. Implementation details are presented in Section \ref{sec:imp_supp}. Additional experiments and ablations are provided in Section \ref{sec:result-supp}. %

\input{sec/supp_benchmark}

\input{sec/supp_details}

\input{sec/supp_results}

%% file: sec/supp_benchmark.tex
\section{Additional Details of our Benchmark}
\label{sec:dataset-supp_}

\textbf{Overview.}  
As mentioned in Section 3 of the main paper, our benchmark systematically evaluates the ability of text-to-image models to perceive and generate fine-grained color attributes across single-object and multi-object prompts. Below, we provide additional implementation details beyond the scope of the main paper:

\textbf{a. Selecting the Color Set:}  
As outlined in the main paper, we start with 140 named colors from the HTML colors table. To ensure robustness in our analysis, we finalize a set of 35 colors that are reliably recognized by both Stable Diffusion 1.4 and 2.1. Here, we provide more implementation details for the filtering process: 
\begin{itemize} 
    \item \textbf{Evaluation of Color Accuracy:} For each color-object pair, we generated images using the prompt ``a \textit{{color}} colored \textit{{object}}'' across the 5 selected objects (\textit{backpack}, \textit{bench}, \textit{car}, \textit{pillow}, and \textit{truck}) and 10 random seeds. 
    
    \item \textbf{Filtering Threshold:} We computed the mean CIELAB distance(we used $\Delta E_{CMC}$ which is more sensitive to subtle differences between similar colors) for each generated image across seeds and objects, retaining only those with distances below a threshold of 15 for both models. This threshold was chosen based on perceptual experiments to balance inclusiveness with accuracy. The finalized set of 35 selected colors is shown below:

\begin{center}
\scriptsize%
\begin{verbatim}
COLORS_final = {
    'HotPink', 'LightPink', 'Tomato', 'BlanchedAlmond', 
    'RoyalBlue', 'SteelBlue', 'CornflowerBlue', 
    'SkyBlue', 'LightSkyBlue', 'LightSteelBlue', 
    'LightBlue', 'PowderBlue', 'Teal', 'MediumTurquoise', 
    'Turquoise', 'Aqua', 'Cyan', 'PaleTurquoise', 
    'LightCyan', 'Linen', 'WhiteSmoke', 'LavenderBlush', 
    'GhostWhite', 'FloralWhite', 'MintCream', 'Snow', 
    'Ivory', 'White', 'Black', 'DimGray', 'SlateGray', 
    'Gray', 'LightSlateGray', 'LightGray', 'Silver'
}
\end{verbatim}
\end{center}
\end{itemize}
    
\textbf{b. Color Pairing:}  
To evaluate how perceptual similarity between colors influence color fidelity and attribute leakage, we pair each color in the finalized set with three \textbf{close} colors (perceptually similar) and three \textbf{distant} colors (perceptually distinct) from the same set. Specifically:
For each color in the finalized set, we compute the CIELAB distances(we used $\Delta E_{ab}$, as we are specifically interested in maximizing perceptual distinctness between the color and all other colors. Based on these distances, we define the following pair types:

\begin{itemize}
    \item \textbf{Close Pairs:}  
    Close pairs are colors with perceptual similarity to the reference color, defined as having a CIELAB distance in the range [5, 30] while also belonging to a different color group as specified in the \href{https://en.wikipedia.org/wiki/Web_colors#Extended_colors}{HTML color definitions}.

    \item \textbf{Distant Pairs:}  
    Distant pairs are colors with clear perceptual differences from the reference color, having a CIELAB distance greater than 60.
\\

Example Pairing for ``SkyBlue" and ``LavenderBlush" in Figure~\ref{fig:color_pairs}:

\begin{figure}[h]
    \centering
    \includegraphics[width=1.\linewidth]{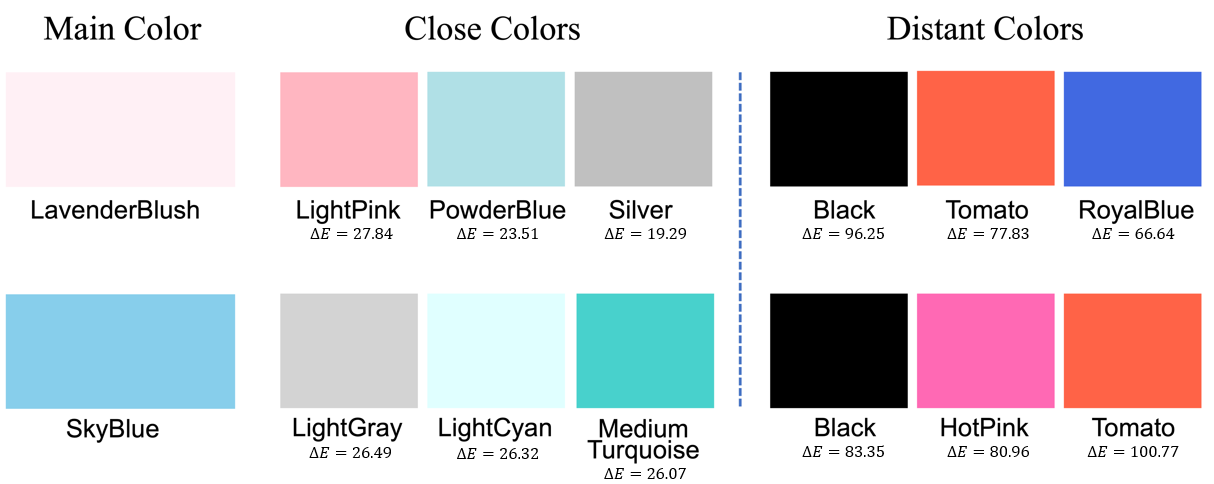}
    \caption{Examples of close and distant color pairs. The leftmost column in each row shows the main color (e.g., LavenderBlush or SkyBlue). The middle column illustrates \textit{close} colors, which are perceptually similar, lower LAB distance (e.g., LightPink, PowderBlue and {Silver}), while the right column shows \textit{distant} colors, which are distinctly different higher LAB distance (e.g., Black, Tomato and RoyalBlue). }
    \label{fig:color_pairs}
\end{figure}
\end{itemize}

\textbf{c. Object Selection:}  
Objects are chosen from a union set derived from Attend-and-Excite, Rich Text, and SynGen. To ensure a fair and unbiased evaluation, objects with dominant intrinsic colors (e.g., bananas) or irregular surfaces (e.g., glasses) are excluded. The selected objects are:

\scriptsize
\begin{verbatim}
objects_final = [
    'chair', 'backpack', 'shirt', 'flower', 'suitcase', 
    'shoe', 'pillow', 'handbag', 'umbrella', 'car', 
    'bow', 'surfboard', 'crown', 'guitar', 'bench', 
    'bicycle','balloon', 'hat', 'bowl'
]
\end{verbatim}
\normalsize

\textbf{d. Prompt Design for Benchmark:}  

\begin{itemize}
    \item \textbf{Pair-Color Prompts:}  
    For each color pair, we randomly sample 2 objects from the selected set to create prompts of the format:  
    ``a \textit{\{color1\}} colored \textit{\{object1\}} and a \textit{\{color2\}} colored \textit{\{object2\}}''.  
    We generate 5 random prompts for each color pair to ensure diversity. Examples of pair-color prompts include:  
    \scriptsize
    \begin{verbatim}
    prompts_for_{SkyBlue,HotPink} = [
         'a skyblue colored backpack 
         and a hotpink colored chair.',
         'a skyblue colored flower 
         and a hotpink colored bench.',
         'a skyblue colored surfboard 
         and a hotpink colored bench.',
         'a skyblue colored hat 
         and a hotpink colored pillow.',
         'a skyblue colored balloon
         and a hotpink colored bicycle.'
    ]
    \end{verbatim}
    \normalsize

    \item \textbf{Single-Color Prompts:}  
    To create a comparison set, single-color prompts are constructed by removing the paired object and color from the pair-color prompts. This enables a controlled comparison between single-color and pair-color scenarios. Examples of single-color prompts include:  
    \scriptsize
    \begin{verbatim}
    prompts_for_{SkyBlue,remove_HotPink} = [
         'a skyblue colored backpack.',
         'a skyblue colored flower.',
         'a skyblue colored surfboard.',
         'a skyblue colored hat.',
         'a skyblue colored balloon.'
    ]
    \end{verbatim}
    \normalsize
\end{itemize}

\section{Evaluation Protocol and Metrics}
\label{sec:Evaluation Metric}

To evaluate model performance in adhering to color specifications, we design a three-step process to isolate dominant object colors while minimizing background and contextual influence. This process ensures accurate measurement of the specified object color by identifying the best-matching dominant color for comparison with the target color. The process consists of three main steps, as illustrated in Figure~\ref{fig:evaluation_metric_pipeline}:

\begin{figure}[t]
    \centering
    \includegraphics[width=\linewidth]{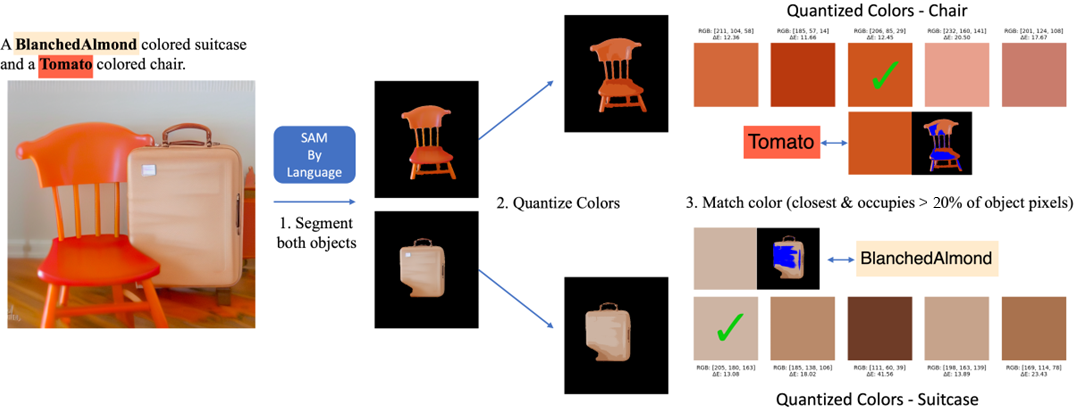} 
    \caption{Illustration of the evaluation metric pipeline. The process involves segmenting the target objects, quantizing their colors, and matching the closest dominant color to the prompt's specified color.}
    \label{fig:evaluation_metric_pipeline}
\end{figure}

\begin{enumerate}
    \item \textbf{Object Segmentation:}  
    The target objects are segmented from the generated image using the SAM. This step eliminates interference from background colors and surrounding objects.

    \item \textbf{Color Quantization:}  
    We perform k-means clustering with $k=5$ on the segmented object regions to extract the dominant colors. This step ensures a compact representation of the object’s color composition.

    \item \textbf{Color Matching:}  
    Among the extracted dominant colors, we identify the color that is the closest match to the target color based on the CIELAB distance($\Delta E_{CMC}$). To avoid small or noisy regions affecting the evaluation, we only consider colors that occupy at least 20\% of the object’s total pixels.
\end{enumerate}

After completing these steps, we compute three evaluation metrics for each color: \textbf{CIELAB Distance($\Delta E_{CMC}$)}, \textbf{RGB L2 Distance}, and \textbf{Accuracy} (as defined in the main paper). Further details on threshold selection for the accuracy metric are provided in Section~\ref{sec:accuracy_threshold}.

We exclude source–edited image pairs if either image fails to accurately depict the two target objects. Specifically, we use SAM to perform object segmentation on the generated images with respect to the original prompt's objects. We then filter images which do not have two object masks or masks with IOU larger than 0.5 to avoid cases of overlapping masks for the same object. This filtering may introduce minor variations in accuracy across models, but ensures a fair and reliable evaluation.

\begin{figure*}
    \centering
    \includegraphics[width=\textwidth]{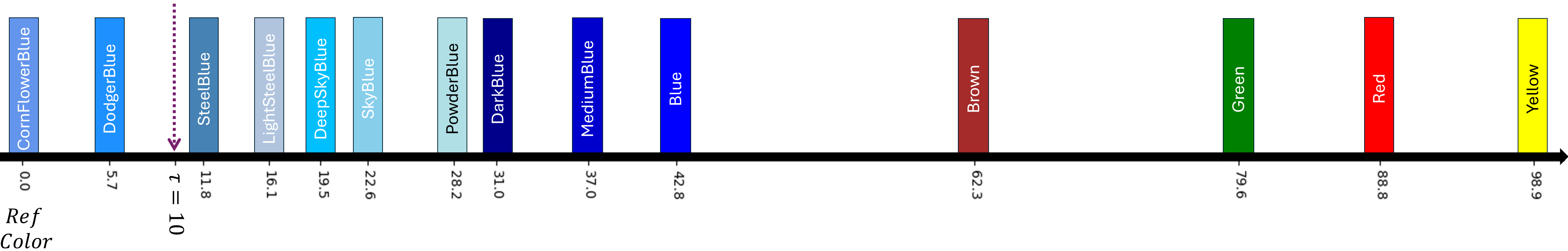} 
    \caption{Illustration of the accuracy threshold based on CIELAB distance using CrownFloweBlue as reference color. We visualize the accuracy threshold with a purple dotted line.}
    \label{fig:acc_threshold}
\end{figure*}

\bigbreak
\noindent\textbf{Additional metrics.} Beyond the primary fidelity metrics above, we also use:
\bigbreak

\textbf{Color Leakage Metric:}
In addition to these metrics, we introduce the \textbf{Color Leakage Metric} to measure how colors intended for one object shift towards another object’s color in multi-object prompts. 

To formalize this metric, we define:
\begin{itemize}
    \item \textbf{Main Object} (object1): The object whose generated color is evaluated.
    \item \textbf{Paired Object} (object2): The object associated with the secondary color (color2) used to formulate the compositional prompt.
\end{itemize}

Given a prompt \textit{``a \{color1\} colored \{object1\} and a \{color2\} colored \{object2\}''}, we compute the CIELAB distance of the generated color for \textit{object1} to both \textit{color1} and \textit{color2}, and take the minimum of these two distances as the CIELAB Distance of this sample, defined as:
\small
\begin{equation}
    d_{\text{shift}} = \min\left(\text{dist}(\text{color}_{\text{object1}}, \text{color1}),\, \text{dist}(\text{color}_{\text{object1}}, \text{color2})\right)
\end{equation}

\normalsize
For clarification, the original metric compares the distance of the generated color of \textit{object1} with the intended \textit{color1} for accuracy computation, while the \textbf{Color Leakage Metric} accounts for the minimum distance with the two colors in the prompt, giving credit (i.e., not penalizing performance) even when color leakage occurs.

This analysis helps diagnose model failures caused by color leakage. We observe significant performance improvement with this lenient evaluation metric compared to the original metric. This reinforces that color leakage is a primary source of error.

\bigbreak
\textbf{Clip Score (CS):} 
We calculate the Clip Score between the edited image and the original text prompt ("a \textit{color1} colored \textit{object1} and a \textit{color2} colored \textit{object2}"). A higher score indicates stronger alignment between the edited image and the input prompt.
\bigbreak
\textbf{Clip Directional Score (CDS):} 
We calculate the Clip Directional Score between the source image, the edited image, the simplified prompt ("a \textit{object1} and a \textit{object2}") and the full text prompt ("a \textit{color1} \textit{object1} and a \textit{color2} \textit{object2}"). Positive score indicates positive influence of the editing method. 
\bigbreak
\textbf{Visual-Question-Answering Score (VQA-Score):} We use a recent multimodal score to evaluate the alignment of the edited image with the input prompt.
We calculated VQA-Score~\cite{lin2024evaluatingtexttovisualgenerationimagetotext}  of the edited and initial images with respect to the original text prompt ("a \textit{color1} colored \textit{object1} and a \textit{color2} colored \textit{object2}"). High score indicates high similarity between the input prompt and the image. We
followed the setup in the original paper and use the highest-performing model \texttt{clip-flant5-xxl}.
\bigbreak
\textbf{Image-Reward.}
We calculated ImageReward~\cite{xu2023imagerewardlearningevaluatinghuman} of the edited image with respect to the original text prompt ("a \textit{color1} colored \textit{object1} and a \textit{color2} colored \textit{object2}"). ImageReward is a learned reward model trained on 137k expert-labeled human preference comparisons, which assigns higher scores to images that are both visually pleasing and semantically aligned with their text prompts.

\medskip
These metrics will be used throughout our experiments. Quantitative comparisons and analysis are reported in Section~\ref{sec:shift_table_differences}.

%% file: sec/supp_details.tex
\section{Implementation Details}
\label{sec:imp_supp}

\subsection{Our Method}  
\label{sec:supp_method}  

Our method operates during inference to enhance image generation by improving color fidelity and attribute binding.  
\subsubsection{Method Key Components}
\begin{itemize}  
    \item \textbf{Model:} We use the official Stable Diffusion v1.5 text-to-image model with 50 diffusion steps and a guidance scale of 7.5.  
    \item \textbf{Runtime:} The complete process (inversion + editing) takes approximately 1.5 minutes per example on a single NVIDIA RTX A5000 GPU.  
    \item \textbf{Hyperparameters:}  
    \begin{itemize}  
        \item \textbf{Stroop Attention Loss:} Step size \(\gamma_{AL}\) = 20, iterations per step \(J_{AL}\) = 10, with early stopping when \(L_a < 0.1\), and loss applied only during steps \(t \leq 35\).  
        \item \textbf{Color Loss:} Step size \(\lambda_{CL}\) = 112.5, one iteration per step, duration \(t \geq 25\).  
    \end{itemize}  
\end{itemize}  

\subsubsection{Editing Procedure}  
Our method takes as input a source image, a target text prompt, and reference RGB values for each object in the prompt. Following the approach described in the main paper, we construct two prompts:  

\begin{itemize}  
    \item \textbf{Simplified Text Prompt}: ``a \textit{object1} and a \textit{object2}"  
    \item \textbf{Full Text Prompt}: ``a \textit{color1} \textit{object1} and a \textit{color2} \textit{object2}"  
\end{itemize}  

To enhance optimization under Stroop Attention Loss, we ensure that color terms in the full prompt are represented by single-token names. Multi-token color names are replaced with their corresponding base color from the HTML color table (e.g., "HotPink" \(\rightarrow\) "Pink").  

We apply our editing method through inversion and backward diffusion, as described in the main paper:  
\begin{enumerate}  
    \item \textbf{Inversion:} We perform DDIM inversion to obtain the latent representation \(Z_T\) and all intermidate latents.
    \item \textbf{Querying Intermediate Latents:} During the backward diffusion process we follow~\cite{Cao_2023_ICCV, liu2024understandingcrossselfattentionstable} using the intermidate latents from the inversion process to improve reconstruction fidelity and editing precision.
    \item \textbf{Pseudo-GT Cross-Attention Mapping:} We run the backward process with the \textbf{Simplified Text Prompt} and store the resulting cross-attention maps as pseudo-ground truth.  
    \item \textbf{Optimization:} Using the \textbf{Full Text Prompt}, we optimize the latents via Stroop Attention Loss and Color Loss, guided by the stored pseudo-GT attention maps and the reference RGB values.  
    \item \textbf{Segmentation for Color Loss:} Our method requires segmentation masks for latent optimization. We provide two variants for segmentation (the second variant is only discussed in the supplementary material):  
    \begin{itemize}  
        \item \textbf{SAM-based segmentation}, we used SAM~\cite{ravi2024sam2} official GitHub repository, In cases where multiple instances of the same object class are present, we merge their segmentations into a single mask.  
        \item \textbf{K-Means clustering on self- and cross-attention maps}, following \cite{Patashnik_2023_ICCV}, with parameters \(num\_segments=8\) and \(background\_segment\_threshold=0.35\).  
    \end{itemize}
    \item \textbf{Self-Attention Replacement:} We followed~\cite{liu2024understandingcrossselfattentionstable} and perform self-attention replacements for the first 40 steps, replacing the self-attention maps in layers 4-14.
\end{enumerate}  

This structured approach ensures improved color consistency and accurate attribute binding in the generated images.
The total objective used in our inference-time approach is:

\begin{align}
    L_{Total} = \gamma_{AL}\cdot{L_{Attention}} + \lambda_{CL}\cdot{M^{simp}_{fg}}\cdot{L_{Color}}
\end{align}
And the corresponding optimization step per iteration is:
\begin{align}
     Z_t &= Z_t - \frac{dL_{Total}}{dZ_t} %
\end{align}
We use \(\gamma_{AL}\) = 20  when \(t<=35\) and  \(\lambda_{CL}\) = 112.5 when \(t>=25\) otherwise 0.
This means that the Stroop attention loss is only applied for the first 35 time steps, and the color loss is only applied in the later iterations after the first 25 steps.
The rationale is that the cross-attention maps should align with the relevant object’s pseudo-ground-truth maps \(A^{simp}_{object}\) early in the process; otherwise, the result may drift.
We stop the Attention Loss after 35 steps to allow flexibility to the diffusion process and to allow the color loss to be more dominant.
We start the color loss later because it depends on the interaction between the color and object cross-attention maps with respect to the relevant object on the image. 

Finally, we want to ensure that our modifications do not alter the structure of the original image. To preserve the structural integrity of the original image, we draw inspiration from prior work in diffusion-based image editing~\cite{hertz2022prompttopromptimageeditingcross,Cao_2023_ICCV, liu2024understandingcrossselfattentionstable} and replace the self-attention for the full prompt with the self-attention obtained using the simplified prompt. In addition, to maintain the background, we perform blending at time \(t_b\) similar  to \cite{Patashnik_2023_ICCV} in order to preserve the background appearance during the editing process (see Figure 3 in the main paper). Concretely, for each object, we use the object masks $M^{simp}_{object}$ introduced previously in Section 4.2.2 in main paper and take the union to obtain $M^{simp}_{fg}$. Then, we perform blending at time $t_b$ during the backward process, updating the latent as follows: %
\begin{align}
Z_{t_{b}} \rightarrow Z_{t_{b}} \cdot {M^{simp}_{fg}} + Z^{simp}_{t_{b}} \cdot ({1-M^{simp}_{fg}})
\label{eq:blend}
\end{align}
where \(Z_{t_{b}}\) is the latent with the full prompt and \(Z^{simp}_{t_{b}}\) is from the corresponding time step for the simplified prompt.

\subsection{Evaluation Baselines and Compared Methods}

 \label{sec:Baselines}

 We evaluate each model as described in the main paper. Here, we provide additional implementation details.

 \subsubsection{Pretrained Models} \begin{enumerate} \item \textbf{Stable Diffusion (SD):} Versions \textbf{1.4, 1.5, and 2.1}, official diffusers library defaults. \item \textbf{FLUX (Dev Version):} Resolution: $1024 \times 1024$, guidance scale: 3.5, inference steps: 28. \end{enumerate}

 \subsubsection{Inference-Time Models} All models use official GitHub implementations with default parameters. Specific notes include: \begin{enumerate} \item \textbf{A\&E}, \textbf{Structured-Diffusion} and \textbf{SynGen}: Based on SD 1.4. \item \textbf{RichText}: Based on SD 1.5, using a JSON config to describe the input prompt, we followed the official GitHub implementations and used the prompt "A \textit{object1} and a \textit{object2}" specifying RGB color attributes directly (without color names). 
 \item \textbf{Bounded-Attention}: Based on SD 1.5, using default bounding boxes provided by the authors. For single-object prompts, we used the first default bounding box. Additionally Bounded-Attention requires the input prompt to be split into subject token indices. We use the token division tool provided by~\cite{NEURIPS2023_0b08d733} to generate it.
\end{enumerate}

\subsubsection{Post-Generation Editing Methods}
In addition to the methods above, we further compare our editing method with AnySD~\cite{yu2025anyeditmasteringunifiedhighqualityz}, MagicBrush~\cite{zhang2024magicbrushmanuallyannotateddataset},  InstructPix2Pix\cite{brooks2023instructpix2pixlearningfollowimage}, FPE~\cite{liu2024understandingcrossselfattentionstable} and MasaCtrl~\cite{Cao_2023_ICCV}. We used the default setup from the official GitHub repository for them. Additionally, for MasaCtrl, we followed their recommendation to inject the reference latents from the inversion process to the backward process for better reconstruction and editing results. 
To enable multi-object color editing with AnySD. For AnySD and MagicBrush we followed the official instruction prompt from AnySD applying two consecutive single-object edits, one for each color-object pair. For InstructPix2Pix we follow their official editing prompt, applying two consecutive single-object edits, one for each color-object pair as it resulted in the best performance.

%% file: sec/supp_results.tex
\definecolor{ghostwhite}{rgb}{0.972, 0.972 1} 
\definecolor{royalblue}{rgb}{0.2549, 0.411 0.882} 
\definecolor{dimgray}{rgb}{0.41, 0.41, 0.41} 
\definecolor{aqua}{rgb}{0,1, 1} 
\definecolor{mintcream}{rgb}{0.96,1, 0.98} 
\definecolor{cornflowerblue}{rgb}{0.392,0.584, 0.929} 
\definecolor{hotpink}{rgb}{1, 0.411, 0.705} 
\definecolor{tomato}{rgb}{1, 0.388 0.278} 
\definecolor{paleturquoise}{rgb}{0.686, 0.933, 0.933} 

\section{Additional Results}
\label{sec:result-supp}

\subsection{ Analysis of Color Leakage}
\label{subsection:cross_color_swaping}

\begin{table}[t]
\centering

\textbf{LAB} \\[5pt]
\resizebox{\columnwidth}{!}{
\begin{tabular}{l|cc||cc}
\toprule
\textbf{Method} & \multicolumn{2}{c||}{\textbf{Close-Mean}} & \multicolumn{2}{c}{\textbf{Distant-Mean}} \\
 & \textbf{Original} & \textbf{w/Color Leakage} & \textbf{Original} & \textbf{w/Color Leakage} \\
\midrule
SD 1.4 & 16.80 & \textbf{12.16} & 22.02 & \textbf{13.35} \\
SD 1.5 & 16.53 & \textbf{11.80} & 22.54 & \textbf{11.94} \\
SD 2.1 & 18.09 & \textbf{13.72} & 25.38 & \textbf{12.27} \\
FLUX                 & 12.50 & \textbf{9.41} & 15.23 & \textbf{12.19} \\
A\&E  & 17.02 & \textbf{12.63} & 20.36 & \textbf{14.78} \\
StructDiff & 14.46 & \textbf{10.40} & 19.79 & \textbf{11.63} \\
SynGen          & 12.56 & \textbf{12.10} & 11.13 & \textbf{10.96} \\
RichText       & 18.60 & \textbf{12.74} & 21.66 & \textbf{16.35} \\
BA & 13.10 & \textbf{10.41} & 21.77 & \textbf{12.28} \\
\bottomrule
\end{tabular}
}

\vspace{10pt}

\textbf{RGB} \\[5pt]
\resizebox{\columnwidth}{!}{
\begin{tabular}{l|cc||cc}
\toprule
\textbf{Method} & \multicolumn{2}{c||}{\textbf{Close-Mean}} & \multicolumn{2}{c}{\textbf{Distant-Mean}} \\
 & \textbf{Original} & \textbf{w/Color Leakage} & \textbf{Original} & \textbf{w/Color Leakage} \\
\midrule
SD 1.4 & 98.07 & \textbf{83.80} & 126.35 & \textbf{91.14} \\
SD 1.5 & 90.71 & \textbf{79.30} & 127.35 & \textbf{80.35} \\
SD 2.1 & 99.11 & \textbf{84.15} & 135.04 & \textbf{83.44} \\
FLUX                 & 71.41 & \textbf{59.87} & 79.50 & \textbf{70.56} \\
A\&E  & 81.51 & \textbf{70.65} & 103.47 & \textbf{86.68} \\
StructDiff & 80.01 & \textbf{65.62} & 101.06 & \textbf{73.72} \\
SynGen          & 81.82 & \textbf{80.19} & 67.21 & \textbf{66.81} \\
RichText       & 78.40 & \textbf{68.30} & 84.99 & \textbf{74.22} \\
BA & 79.73 & \textbf{69.24} & 118.13 & \textbf{78.40} \\
\bottomrule
\end{tabular}
}

\vspace{10pt}

\textbf{Accuracy} \\[5pt]
\resizebox{\columnwidth}{!}{ %
\begin{tabular}{l|cc||cc}
\toprule
\textbf{Method} & \multicolumn{2}{c||}{\textbf{Close}} & \multicolumn{2}{c}{\textbf{Distant}} \\
 & \textbf{Original} & \textbf{w/Color Leakage} & \textbf{Original} & \textbf{w/Color Leakage} \\
\midrule
SD 1.4 & 0.38 & \textbf{0.58} & 0.28 & \textbf{0.45} \\
SD 1.5 & 0.36 & \textbf{0.54} & 0.30 & \textbf{0.48} \\
SD 2.1 & 0.33 & \textbf{0.53} & 0.26  & \textbf{0.50} \\
FLUX                 & 0.54 & \textbf{0.66} & 0.49 & \textbf{0.51} \\
A\&E  & 0.46 & \textbf{0.64} & 0.38 & \textbf{0.46} \\
StructDiff & 0.39 & \textbf{0.65} & 0.39 & \textbf{0.55} \\
SynGen          & 0.49 & \textbf{0.50} & 0.56 & \textbf{0.57} \\
RichText       & 0.35 & \textbf{0.57} & 0.30 & \textbf{0.32} \\
BA & 0.55 & \textbf{0.64} & 0.29 & \textbf{0.48} \\
\bottomrule
\end{tabular}
}
\caption{
Results for color fidelity and attribute leakage metrics (LAB, RGB, and Accuracy) across \textbf{Close} and \textbf{Distant} color sets, comparing the original evaluation and the additional \textbf{Color Leakage Metric} metric. The Color Leakage Metric reduces penalization from color blending across objects, improving overall scores and reducing the performance gap between close and distant color sets. \textbf{Bolded} values indicate better scores when comparing original and Color Leakage Metric.
}

\label{tab:shift_table_differences}
\end{table}

As noted in the main paper, pretrained models and most inference-time models perform worse on distant color sets than close sets.  We hypothesize that this discrepancy arises due to amplified penalties caused by \textbf{cross-object color leakage}, where colors intended for one object inadvertently blend or swap with those of another. When the two colors are visually more distinct (e.g., in the distant set), even small color shifts or blending between objects result in large perceptual differences. In contrast, in close sets, the inherent color similarities naturally reduce the perceived error.

To further analyze this phenomenon, we introduced an additional metric—the \textbf{Color Leakage Metric}—as detailed in Section~\ref{sec:Evaluation Metric}. Unlike the original evaluation, which strictly penalizes deviations from the intended color, the Color Leakage Metric credits models even when color leakage occurs, as long as the generated color closely matches either of the two colors specified in the prompt.

Results in Table~\ref{tab:shift_table_differences} substantiate this hypothesis: scores under the Color Leakage Metric are consistently higher across LAB, RGB, and Accuracy metrics, highlighting that cross-object color leakage is a primary contributor to model failures. Additionally, the significant reduction in the performance gap between close and distant color sets under this metric reinforces the idea that many errors previously classified as failures were actually instances of color leakage rather than complete misgeneration. These findings provide a complementary evaluation perspective, allowing for a more fine-grained understanding of model limitations in compositional color tasks.

An interesting observation is that SynGen’s performance remains nearly unchanged under the swap metric, showing minimal improvement across all metrics. As mentioned in main paper SynGen uses contrastive loss to separate different object's colors, this feature diminishes color leakage and swapping significantly and thus we see very little improvement  under the swap metric.

\subsection{Evaluation Using Different Metrics}\label{sec:shift_table_differences}
\input{tables_cvpr/full_metrics_table}

In Table~\ref{tab:full_metrics_comparison_models} we compare our editing method against all baselines across LAB, RGB mean, and ACC metrics. Our approach achieves consistent improvements across all models and metrics, highlighting its effectiveness and generality.

\input{tables_cvpr/image_reward}
In Table~\ref{tab:full_imagereward_comparison_models} we quantify image quality using Image Reward. Our evaluation shows that our model yields comparable performance  while achieving significantly more accurate color edits. The largest margin is observed in distant-color prompts, where color edits are more challenging and prone to quality degradation. These results suggest that our edits do not harm—and may often enhance—visual fidelity.

In Table~\ref{tab:CS_CDS_FID_VQA} we compare the editing performance of the different models using CLIP Score (CS), CLIP Directional Score (CDS), and VQAScore improvement (edited compare to initial), see Table~\ref{tab:CS_CDS_FID_VQA}.
Our results show that while our CS and CDS scores are lower than FPE and InstructPix2Pix, this does not necessarily indicate inferior performance. As discussed in Figure 2 of the main paper, these metrics are often coarse and struggle to accurately capture fine-grained semantics, particularly in tasks involving precise color control. Thus, their lower scores may not be fully reflective of the effectiveness of our approach in this specific color-focused evaluation.

To address the limitations of these traditional metrics, we also incorporate VQAScore~\cite{lin2024evaluatingtexttovisualgenerationimagetotext}, which considered the SOTA for measuring similarity between image and text. Our editing method outperforms or is comparable to AnySD, MagicBrush, InstructPix2Pix, FPE and MasaCtrl across all model datasets, except for FLUX-close, where the difference is minimal. However, even VQAScore, despite being a more advanced metric, still struggles to fully capture fine-grained semantic alignment, as illustrated in Figure 2 of the main paper.
The limitations of existing evaluation metrics further emphasize the importance of our explicit evaluation pipeline, which directly quantifies perceptible semantic differences in color alignment, providing a more reliable assessment of model performance in compositional color tasks.

\begin{table}[t]
\centering
\resizebox{\columnwidth}{!}{
\begin{tabular}{l|cccccc}
\toprule

\textbf{Method} & \textbf{Ours (CS\textbackslash CDS\textbackslash VQA)} & \textbf{AnySD} & \textbf{MagicBrush} & \textbf{InstructPix2Pix} & \textbf{FPE} & \textbf{MasaCtrl)} \\
\midrule
\multicolumn{7}{c}{\textbf{Close}} \\
\midrule
SD 1.4       &  0.0035\textbackslash 0.010\textbackslash \textbf{0.08}  & 0.0015\textbackslash -0.0031\textbackslash 0.02  & -0.00097\textbackslash 0.0084\textbackslash 0.05 & -0.00028\textbackslash \textbf{0.025}\textbackslash 0.036 & \textbf{0.0036}\textbackslash 0.018\textbackslash 0.01 &  0.0031\textbackslash 0.0080\textbackslash 0.01 \\
SD 1.5        &  \textbf{0.0036}\textbackslash 0.0088\textbackslash \textbf{0.08}  &  0.0021\textbackslash 0.0042\textbackslash 0.03   & -0.0018\textbackslash 0.015\textbackslash 0.054 & -0.0017\textbackslash \textbf{0.023}\textbackslash 0.05 & 0.0031\textbackslash 0.019\textbackslash 0.01 &  0.0026\textbackslash 0.0049\textbackslash 0.00 \\
SD 2.1     &  0.0029\textbackslash 0.016\textbackslash \textbf{0.10}  &  -0.00063\textbackslash 0.0055\textbackslash 0.03  & -0.0041\textbackslash 0.018\textbackslash 0.066 & -0.0025\textbackslash \textbf{0.030}\textbackslash 0.067 & \textbf{0.0038}\textbackslash 0.017 \textbackslash 0.02 &  0.0030\textbackslash 0.0044\textbackslash -0.01 \\
FLUX     &  \textbf{0.0074}\textbackslash 0.024\textbackslash 0.00  &  0.0021 \textbackslash 0.028 \textbackslash 0.00 & -0.0029\textbackslash \textbf{0.036} \textbackslash -0.054 & -0.0015\textbackslash \textbf{0.036} \textbackslash -0.069 & 0.0049\textbackslash 0.026\textbackslash \textbf{0.01} & -0.00081\textbackslash 0.015\textbackslash -0.06 \\
A\&E         &  0.0032\textbackslash 0.014\textbackslash \textbf{0.10}  &  0.0041\textbackslash 0.0058\textbackslash 0.03 & -0.0011\textbackslash 0.024\textbackslash 0.035 & -0.0028\textbackslash \textbf{0.033} \textbackslash 0.029 & \textbf{0.0049}\textbackslash 0.030\textbackslash 0.02 &  0.0022\textbackslash 0.014\textbackslash 0.00 \\
StructDiff         &  0.0046\textbackslash 0.012\textbackslash \textbf{0.10} &  \textbf{0.0069}\textbackslash 0.0034\textbackslash 0.03 & 0.0015\textbackslash 0.016\textbackslash 0.059 & 0.0064\textbackslash \textbf{0.024}\textbackslash 0.049 & 0.0038\textbackslash 0.022\textbackslash 0.02 & 0.0054\textbackslash 0.0088\textbackslash 0.01 \\
SynGen     &  0.0054\textbackslash 0.026\textbackslash \textbf{0.03}  &  0.00087\textbackslash -0.0040\textbackslash -0.01 &  0.0043\textbackslash 0.020\textbackslash -0.05 & 0.0013\textbackslash 0.023\textbackslash -0.035 & \textbf{0.0073}\textbackslash \textbf{0.038}\textbackslash \textbf{0.03}  &  0.0063\textbackslash 0.017\textbackslash 0.01 \\
RichText &  0.013\textbackslash 0.051\textbackslash \textbf{0.12}  &  \textbf{0.018}\textbackslash 0.014\textbackslash 0.07 & 0.016\textbackslash 0.060\textbackslash 0.084 & 0.021\textbackslash \textbf{0.070} \textbackslash 0.093 & 0.016\textbackslash 0.062 \textbackslash 0.07  &  0.011\textbackslash 0.029\textbackslash 0.00 \\
BA     &  0.0025\textbackslash 0.0081\textbackslash \textbf{0.04} &  0.0023\textbackslash 0.00035\textbackslash 0.01 & -0.021\textbackslash 0.0080\textbackslash -0.12 & -0.0046\textbackslash \textbf{0.024} \textbackslash -0.011 & \textbf{0.0034}\textbackslash 0.022\textbackslash 0.01 & -0.00025\textbackslash 0.0063\textbackslash -0.03 \\
\midrule
\multicolumn{7}{c}{\textbf{Distant}} \\
\midrule
SD 1.4        &  0.00095\textbackslash -0.0096\textbackslash \textbf{0.12} &  0.0015\textbackslash -0.0020\textbackslash 0.03 & -0.0061\textbackslash 0.0012\textbackslash 0.033 & -0.0083\textbackslash 0.014\textbackslash -0.04 &  \textbf{0.0032}\textbackslash \textbf{0.018}\textbackslash 0.00 &  0.0020\textbackslash 0.0076\textbackslash -0.01 \\
SD 1.5        &  0.0013\textbackslash -0.0022\textbackslash \textbf{0.13} &  0.0013\textbackslash 0.0046\textbackslash 0.03 & -0.0038\textbackslash 0.015\textbackslash 0.062 & -0.0045\textbackslash \textbf{0.026}\textbackslash -0.045 &  \textbf{0.0030}\textbackslash 0.018\textbackslash 0.00 &  0.0021\textbackslash 0.0053\textbackslash -0.01 \\
SD 2.1     &  0.0019\textbackslash 0.00018\textbackslash \textbf{0.14}  &  -0.00055\textbackslash 0.00020\textbackslash 0.04 & -0.0044\textbackslash 0.016\textbackslash 0.073 & -0.0066\textbackslash \textbf{0.023} \textbackslash -0.0025 & \textbf{0.0044}\textbackslash 0.022\textbackslash 0.01 &  0.0020\textbackslash 0.010\textbackslash -0.01 \\
FLUX     &  0.0080\textbackslash 0.026\textbackslash \textbf{0.03}   &  0.0050\textbackslash 0.029\textbackslash 0.01 & 0.00031\textbackslash 0.034\textbackslash -0.11 & -0.0012\textbackslash \textbf{0.041}\textbackslash -0.22 & \textbf{0.0082}\textbackslash \textbf{0.041}\textbackslash 0.00  &  0.0051\textbackslash 0.023\textbackslash -0.03 \\
A\&E        &  0.0024\textbackslash 0.014\textbackslash \textbf{0.13}   &  0.0033\textbackslash 0.011\textbackslash 0.02  & -0.0058\textbackslash 0.028\textbackslash -0.038 & -0.0072\textbackslash \textbf{0.030}\textbackslash -0.11 & \textbf{0.0051}\textbackslash \textbf{0.030}\textbackslash 0.01  &  0.0027\textbackslash 0.012\textbackslash -0.01 \\
StructDiff         &  0.0024\textbackslash 0.0084\textbackslash \textbf{0.16}  &  \textbf{0.0054}\textbackslash 0.01\textbackslash 0.03  & 0.0018\textbackslash 0.021\textbackslash 0.059 & 0.0034\textbackslash \textbf{0.029}\textbackslash 0.0016 & 0.0040\textbackslash 0.027\textbackslash 0.02 &   0.0041\textbackslash 0.014\textbackslash -0.01 \\
SynGen    &  0.0075\textbackslash 0.027\textbackslash \textbf{0.00} &  0.0011\textbackslash 0.0030\textbackslash \textbf{0.00} &  0.00054\textbackslash 0.026\textbackslash -0.14 & -0.0041\textbackslash 0.019\textbackslash -0.19 & \textbf{0.0078}\textbackslash \textbf{0.039}\textbackslash -0.02  &  0.0057\textbackslash 0.016\textbackslash \textbf{0.00} \\
RichText &  \textbf{0.022}\textbackslash 0.070\textbackslash \textbf{0.21} &  \textbf{0.022}\textbackslash 0.032\textbackslash 0.14 & 0.030\textbackslash 0.093\textbackslash 0.11 & 0.031\textbackslash \textbf{0.097}\textbackslash 0.027 &  0.021\textbackslash 0.079\textbackslash 0.13 &  0.011\textbackslash 0.028\textbackslash 0.03 \\
BA     &  0.000059\textbackslash -0.0064\textbackslash \textbf{0.13} &  0.0024\textbackslash 0.0048\textbackslash 0.04 & -0.025\textbackslash 0.0031\textbackslash -0.089 & -0.011\textbackslash 0.0089\textbackslash -0.13 &  \textbf{0.0035}\textbackslash \textbf{0.016}\textbackslash 0.00 &  0.0020\textbackslash 0.0059\textbackslash -0.01 \\
\bottomrule
\end{tabular}
}
\caption{Evaluation using additional metrics. We report performance using CLIP Similarity (CS), CLIP Directional Similarity (CDS), and VQAScore improvement, comparing our method with AnySD, MagicBrush, InstructPix2Pix, FPE and MasaCtrl (CS\textbackslash CDS\textbackslash VQA Score). Higher is better.}
\label{tab:CS_CDS_FID_VQA}
\end{table}

\subsection{Comparison with ColorPeel}
\label{sec:colorPeel}
We perform an additional comparison to ColorPeel~\cite{butt2024colorpeelcolorpromptlearning}, a text-to-image model that uses prompt learning to learn specific RGB colors embeddings. We followed their official GitHub repository and created four colors:\textcolor{red235}{\textless c1*\textgreater},\textcolor{green235}{\textless c2*\textgreater},\textcolor{blue235}{\textless c3*\textgreater},\textcolor{yellow235}{\textless c4*\textgreater}. We generated images using prompts inspired by their work. As illustrated in Figure~\ref{fig:colorpeel_fig} we see that even though ColorPeel specifically targeted the issue of correctly learning RGB color representation, it still fails to generate the colors correctly in the multi-color multi-object setting, as it is challenged by issues such as leakage and swapping of colors.
We see that our editing method is able to correctly bind the colors to the objects, resulting in edited images that correspond to the input prompt.  
\input{figures/colorpeel/fig_colorpeel}

\subsection{Runtime Comparison}
\label{sec:runtime}
\input{tables_cvpr/runtime_comparison}
In Table~\ref{tab:runtime_comparison} we added a runtime comparison between our method and the baselines. As our method optimizes the latents for performing each color edit, it is indeed a bit slower than prior approaches. However, our approach delivers substantially higher color fidelity in comparison to prior work, remaining practical in terms of runtime for editing scenarios where quality is prioritized over speed.

\medskip
\subsection{Ablation Studies}
\label{sec:ablations}

\subsubsection{Method Component Ablation}
We conduct ablation studies to isolate the contributions of key components in our method. We performed the ablation study on the Stable Diffusion 2.1 generations as it yielded the highest improvement considering both close and distant pairs. Specifically we conduct the following ablations: (1) \emph{Removing the Stroop Attention Loss} (w/o Stoop Attention Loss), which removes the attention loss from our inference-time approach. (2) \emph{Removing the Color Loss} (w/o Color Loss), which removes the color loss from our inference-time approach. (3) \emph{Removing the Simplifed Prompt} (w/o Simplifed), which removes the Simplified Prompt from our Stroop attention loss, such that the attention map of the color token is compared against the object token from the Full Prompt directly.

 The results are presented on Table \ref{tab:ablation}. Removing the \textit{Color Loss} term leads to a notable drop in color accuracy, particularly in the pair-color benchmark, as it directly enforces alignment between specified and generated colors. Similarly, removing the \textit{Stroop Attention Loss} results in increased attribute leakage, with color attributes often incorrectly associated with the wrong objects. Furthermore, as also illustrate in the table,  the simplified prompt (without colors) allows for achieving  more accurate cross-attention maps for image color-editing.
 
\begin{table}[t]
\centering
\resizebox{\columnwidth}{!}{
\begin{tabular}{l|ccc}
\toprule
\textbf{Method} & \textbf{LAB (Imp $\uparrow$)} & \textbf{RGB (Imp $\uparrow$)} & \textbf{Acc (Imp $\uparrow$)} \\
\midrule
\multicolumn{4}{c}{\textbf{Close}} \\
\midrule
w/o Stroop Attention Loss       &  5.63 &  43.49 &  0.13 \\
w/o Color Loss         & 3.73 & 15.41 & 0.04 \\
w/o Simplified Prompt        &  6.87 &  49.21 &  \textbf{0.22} \\
Ours (Full pipeline)         &  \textbf{7.43} &  \textbf{51.02} &  \textbf{0.22}\\
\midrule
\multicolumn{4}{c}{\textbf{Distant}} \\
\midrule
w/o Stroop Attention Loss       &  7.82 &  60.40&  0.13 \\
w/o Color Loss         &  6.99 & 39.82 &  0.10 \\
w/o Simplified Prompt        &  11.81 &  79.25 &  0.21 \\
Ours (Full pipeline)         &  \textbf{12.43} &  \textbf{79.42} &  \textbf{0.24}\\
\bottomrule
\end{tabular}
}
\caption{Ablation study of our ColorEdit approach conducted on images generated with SD 2.1; see Section \ref{sec:ablations} for additional details.} 
\label{tab:ablation}
\end{table}

\subsubsection{Inversion Method Ablation}
We evaluate our editing method under different inversion techniques, comparing DDIM inversion (which we use for all experiments)  with Null-Text inversion. As shown in Table~\ref{tab:ablation_inversion}, Null-Text inversion yields improved editing performance. %
However, due to its computational overhead—more than doubling the editing time—we do not adopt it in our main results. Nonetheless, it demonstrates that our approach can perform robustly over different inversion methods.
\input{rebuttal_files/inversion/inversion}

\subsection{Performance without using SAM masks}
\label{sec:comparison_without_sam}
As mentioned in the main paper, we use SAM~\cite{ravi2024sam2} to extract segmentation masks which are utilized in our editing approach. Next we demonstrate that our approach can also operate without these external masks. For this variant, we use K-means based on the self and cross-attention map during the backward process to extract the objects' masks, similarly to the approach proposed in~\cite{patashnik2023localizingobjectlevelshapevariations}. As can be seen on Table \ref{tab:comparison_without_sam}, the overall results are very similar indicating that our method is not highly-dependent on the segmentation masks obtained by SAM.

\begin{table}[t]

\centering
\resizebox{\columnwidth}{!}{
\begin{tabular}{l|cc}
\toprule
\textbf{Method} & \textbf{with SAM(Imp $\uparrow$)} & \textbf{without SAM (Imp $\uparrow$)} \\
\midrule
\multicolumn{3}{c}{\textbf{Close}} \\
\midrule
SD 1.4       &  \textbf{6.15\textbackslash 46.88\textbackslash 0.21} &  4.63\textbackslash 38.76\textbackslash 0.17  \\
SD 1.5        &  \textbf{6.04\textbackslash 38.39\textbackslash 0.18}  &  5.70\textbackslash 38.24\textbackslash0.15 \\
SD 2.1     &  \textbf{7.43\textbackslash 51.02\textbackslash 0.22}  &  6.75\textbackslash 43.02\textbackslash 0.19\\
FLUX     &  \textbf{4.11\textbackslash 30.12\textbackslash0.16}  & 3.58\textbackslash 26.84\textbackslash0.11\\
A\&E         &  6.40\textbackslash 33.16\textbackslash 0.15 &  \textbf{6.53\textbackslash 32.26\textbackslash 0.16} \\
StructDiff         &  \textbf{4.66\textbackslash 32.81\textbackslash 0.23} &  4.15\textbackslash 25.07\textbackslash 0.20 \\
SynGen     &  4.23\textbackslash 38.19\textbackslash 0.22   &  \textbf{3.77\textbackslash 37.36\textbackslash 0.23} \\
RichText &  \textbf{7.91\textbackslash 29.59\textbackslash 0.23}  & 6.18\textbackslash 18.25\textbackslash 0.21 \\
BA     &  \textbf{3.57\textbackslash 34.51\textbackslash 0.11}  &  \textbf{3.23\textbackslash 90.62\textbackslash 0.11}\\

\midrule
\multicolumn{3}{c}{\textbf{Distant}} \\
\midrule
SD 1.4        &  \textbf{9.52\textbackslash 70.14\textbackslash 0.20}  &  8.82\textbackslash 64.21\textbackslash 0.17\\
SD 1.5        &  \textbf{10.15\textbackslash 66.24\textbackslash 0.19}  &  8.62\textbackslash 60.19\textbackslash 0.18 \\
SD 2.1     &  \textbf{12.43\textbackslash 79.42\textbackslash 0.24}  &  10.27\textbackslash 70.91\textbackslash0.19\\
FLUX     &  \textbf{5.28\textbackslash 36.94\textbackslash 0.14}  &  4.74\textbackslash 32.37\textbackslash 0.13\\
A\&E       &  \textbf{9.75\textbackslash 50.13 \textbackslash0.23}   & 8.71\textbackslash 45.97\textbackslash 0.18\\
StructDiff         &  \textbf{7.98\textbackslash 44.88\textbackslash 0.18} &  6.45\textbackslash 35
93\textbackslash 0.15 \\
SynGen    &  \textbf{3.05\textbackslash 26.78\textbackslash 0.17}  &  2.61\textbackslash 22.80\textbackslash 0.16\\
RichText &  \textbf{8.17\textbackslash 28.53\textbackslash 0.24}  &  7.69\textbackslash 25.90\textbackslash 0.20 \\
BA    &  8.03\textbackslash 61.89\textbackslash 0.18  &  \textbf{7.99\textbackslash 57.54\textbackslash 0.19}\\
\bottomrule
\end{tabular}
}
\caption{ColorEdit performance (LAB\textbackslash RGB\textbackslash Acc) using different segmentation methods; see Section \ref{sec:comparison_without_sam} for additional details. %
}
\label{tab:comparison_without_sam}
\end{table}

\subsection{Inversion-free Performance}
\label{sec:no_inversion}

\input{tables_cvpr/no_inversion}

\input{figures/no_inversion/no_inversion}
In the previous section, we demonstrated that our method is robust to different inversion techniques. However, since inversion itself may occasionally fail, we further show that our approach does not depend on inversion at all. Specifically, we conduct an experiment where editing is performed without inversion: rather than inverting the image to obtain latents for editing, we generate an image from random noise using the input prompt and then apply our editing procedure on the same random initial latents. This experiment highlights the generality of our method, demonstrating its effectiveness even without inversion.
We perform this evaluation using SD1.5, which serves as the base model for our approach. SAM-based segmentation is not employed in this setting, since image generation and editing occur simultaneously. As shown in Table~\ref{tab:no_inversion}, our method achieves strong performance without inversion, surpassing the standard SD1.5 baseline and confirming both the generality and inversion-independence of our color editing framework.  Figure~\ref{fig:no_inversion} shows that our editing method performs well without inversion, editing the right colors to the balloons and the bench.

\subsection{Analysis of Overlapping Objects}
\input{tables_cvpr/overlap_objects}
\label{sec:overlapp_analysis}
We added an analysis of overlapping objects, considering multiple overlap levels. In Table~\ref{tab:overlap_objects} the analysis shows that our model’s performance remains relatively stable even at high overlap levels ($IOU \geq 0.1$), still consistently improving performance over the source images. For Bounded-Attention we see low overlap as we used the default bounding rects from the official implementation which forces non overlapping objects.

\subsection{Statistical Significance}
\label{sec:statistical_significance}
We performed a paired non-parametric analysis, comparing images edited by our method with both the source images and those produced by baseline approaches (AnySD, MagicBrush, InstructPix2Pix, FPE, and MasaCtrl). The Wilcoxon signed-rank test was applied to paired LAB distance values under the same set of prompts. The results show that our method achieves statistically significant improvements over all baselines ($p<0.05$, Wilcoxon signed-rank test).

\subsection{Accuracy Threshold Validation}
\label{sec:accuracy_threshold}
\input{figures/user_study_example}
\input{tables_cvpr/user_study_stats}
To validate that our selected threshold aligns with human perception, we conducted a user study. Participants were asked whether an object’s color was sufficiently similar to the target text or clearly different, see Figure~\ref{fig:user_study_example}. In Table~\ref{tab:user_study_stats} we compared their responses against multiple thresholds and found that a threshold of 10 provides the best balance between the true positive and true negative rates and best overall performance in terms of AUC, G-mean, and balanced accuracy.

\subsection{Additional Qualitative Results}
\subsubsection{Propriety Models Experiment}

\input{figures/SOTA/SOTA}
In addition to pre-trained and inference-time models, we present qualitative examples from three state-of-the-art (SOTA) models: OpenAI 4O, Gemini 2.5, and Grok 3, conditioned on the prompt depicted in Figure 1 (main paper). As illustrated in Figure~\ref{fig:SOTA_example}, even SOTA propriety models fail to generate the correct colors specified by the input prompt. By applying our editing method, the resulting images exhibit colors that are both accurate and consistent with the prompt.

\subsubsection{Multiple Color Editing}
Our editing approach can generalize to more than two objects. 
We create an extension of our paired-colors/objects dataset which contains three colors and objects. In this extended set, in addition to the close color we added a distant color corresponding to a third object which was chosen randomly from our object list to create prompts of the format: ``a \textit{\{color1\}} colored \textit{\{object1\}} and a \textit{\{color2\}} colored \textit{\{object2\}} and a \textit{\{color3\}} colored \textit{\{object3\}}''. We tested our results using 3 different models which excel in multiple objects prompts: FLUX and SynGen, we added SD 2.1 too as it was our ablation model. Results are reported in Table ~\ref{tab:reviewer3_3_colors}. We see that our editing approach is not limited by the number of objects, and that our method yields substantial improvements on the task of multiple color editing also over this extended set.
\input{rebuttal_files/3_colors/3_colors}

We demonstrate an example of editing three objects (each associated with a different color) in Figure~\ref{fig:three_colors}. Additional examples are shown in Figure 1 (main paper) of the main paper and in Figure~\ref{fig:SOTA_example}.
\input{figures/three_colors/three_colors}

\subsubsection{Real Images Editing}
We also present real image editing results using our method ColorEdit, comparing against AnySD, MagicBrush, InstructPix2Pix, FPE and MasaCtrl; see Figure \ref{fig:real_images}. The images were taken from the internet. 
As illustrated in Figure~\ref{fig:real_images} we successfully edit all real images with respect to the editing prompts, including challenging objects containing fine-grained details, \emph{e.g.}, a bridge. We can see that ColorEdit significantly outperforms AnySD, MagicBrush, InstructPix2Pix, FPE and MasCtrl in the task of real image editing.
\input{figures/real_images/real_images}
In addition we added real image editing of object with dominant intrinsic colors(e.g. banana is usually yellow or green). On Figure~\ref{fig:banana} we exemplify that our editing approach is not limited by object or color, showing multiple color editing of banana.
\input{rebuttal_files/banana/banana}

\subsection{Limitations}
While our method performs well in most scenarios and allows for consistently achieving more accurate color-based edits, it has certain limitations. First, we use segmentation masks to limit our color editing to the relevant objects and preserve image structure. In some cases the segmentation fails due to poor generation quality or challenging image content. In Figure~\ref{fig:sam_fail}, the image generated by SD 2.1 based on the prompt ``a blanched-almond colored hat and a white colored backpack" includes a backpack which is only partially visible, resulting in a segmentation that includes the woman's jacket and shirt as well. In such cases our approach operates over incorrect regions, leading to unintended edits to the jacket and shirt. %
\input{figures/sam_fail/sam_fail}

As our method optimizes the latents for performing each color edit, it is limited by the runtime required to bind the colors to the right object. We foresee future work accelerating these results, possibly using few-step diffusion models.
Third, noisy cross-attention maps could also result in poor performance in some cases. 
Our method uses the cross attention maps extracted using the simplified prompt, while in most cases the procedure work very well, in some cases where objects are generating in overlapping regions and share mutual semantic connection (e.g. chair usually contains a pillow) our simplified prompt struggles to correctly separate the objects (e.g. the pillow from the chair), resulting noisy cross-attention maps that degrade the effectiveness of the Stroop attention loss. In Figure~\ref{fig:sd_fail}, we see that for both simplified prompt and full prompt the cross attention maps are noisy, the pillow cross-attention map is including the chair too which results edited image where the chair is colored in a similar color to the pillow. %
\input{figures/sd_fail/sd_fail}

%% file: tables_cvpr/full_metrics_table.tex
\begin{table}[t]
\centering
\resizebox{\columnwidth}{!}{
\begin{tabular}{l|cccccc}
\toprule
\textbf{Method} & \textbf{Ours(Imp $\uparrow$)} & \textbf{AnySD (Imp $\uparrow$)} & \textbf{MagicBrush (Imp $\uparrow$)} & \textbf{InstructPix2Pix (Imp $\uparrow$)} & \textbf{FPE (Imp $\uparrow$)} & \textbf{MasaCtrl (Imp $\uparrow$)} \\
\midrule
\multicolumn{7}{c}{\textbf{Close}} \\
\midrule
SD 1.4       &  \textbf{6.15\textbackslash 46.88\textbackslash 0.21} &  0.35\textbackslash -0.51\textbackslash 0.03 & 3.35\textbackslash 18.72\textbackslash 0.10 & 0.79\textbackslash 11.94\textbackslash 0.00 &  1.42\textbackslash 8.26\textbackslash 0.03 &  0.36\textbackslash 0.21\textbackslash 0.01 \\
SD 1.5        &  \textbf{6.04\textbackslash 38.39\textbackslash 0.18} &  0.76\textbackslash -1.89\textbackslash -0.01 & 3.29\textbackslash 11.52\textbackslash 0.10 &  2.30\textbackslash 10.72\textbackslash 0.04 & 1.83\textbackslash 3.73\textbackslash 0.03 &  1.05\textbackslash -0.82\textbackslash -0.02 \\
SD 2.1     &  \textbf{7.43\textbackslash 51.02\textbackslash 0.22} &  1.63\textbackslash 3.48\textbackslash 0.03 & 3.54\textbackslash 17.59\textbackslash 0.11 & 3.74\textbackslash 21.84\textbackslash 0.04 & 2.12\textbackslash 7.94\textbackslash 0.03 &  0.57\textbackslash 1.82\textbackslash -0.02  \\
FLUX     &  \textbf{4.11\textbackslash 30.12\textbackslash 0.16} &  1.46\textbackslash 2.06\textbackslash 0.04  & 0.97\textbackslash 5.01\textbackslash 0.02 & -3.22\textbackslash -5.62\textbackslash -0.18 & 1.72\textbackslash 7.54\textbackslash 0.07 & 1.28\textbackslash 3.61\textbackslash 0.02  \\
A\&E        &  \textbf{6.40\textbackslash 33.16\textbackslash 0.15} &  0.47\textbackslash -2.18\textbackslash 0.00 & 5.00\textbackslash 14.62\textbackslash 0.11 & 2.30\textbackslash 3.56\textbackslash -0.05 &  2.43\textbackslash 5.42\textbackslash 0.02 &  0.80\textbackslash 0.38 \textbackslash-0.04 \\
StructDiff        &  \textbf{4.66\textbackslash 32.81\textbackslash 0.23} &  -0.18\textbackslash -4.00\textbackslash 0.03 & 2.93\textbackslash 13.75\textbackslash 0.14 & 1.82\textbackslash 9.12\textbackslash 0.07 &  1.38\textbackslash 5.09\textbackslash 0.06 &  0.80\textbackslash 0.96 \textbackslash 0.02 \\
SynGen     &  \textbf{4.23\textbackslash 38.19\textbackslash 0.22} &  0.17\textbackslash 1.01\textbackslash-0.04  & 1.94\textbackslash 13.07\textbackslash 0.10 & 0.51\textbackslash 15.80\textbackslash -0.01 &  2.89\textbackslash 18.83\textbackslash 0.16  &  1.49\textbackslash 7.37\textbackslash 0.08 \\
RichText &  \textbf{7.91\textbackslash 29.59\textbackslash 0.23} &  2.88\textbackslash 9.38\textbackslash 0.11  & 4.88\textbackslash 3.96\textbackslash 0.16 & 3.29\textbackslash 0.24\textbackslash 0.05 &  5.06\textbackslash 11.72\textbackslash 0.15  &  3.24\textbackslash 5.77\textbackslash 0.07 \\
BA     &  \textbf{3.57\textbackslash 34.51\textbackslash 0.11} &  0.48\textbackslash -1.13\textbackslash -0.03 & 1.47\textbackslash 12.77\textbackslash 0.01 & 0.29\textbackslash 5.65\textbackslash -0.09 &  1.04\textbackslash 8.56\textbackslash 0.02 & 0.34\textbackslash 3.34\textbackslash -0.01 \\

\midrule
\multicolumn{7}{c}{\textbf{Distant}} \\
\midrule
SD 1.4        &  \textbf{9.52\textbackslash 70.14\textbackslash 0.20} &  0.72\textbackslash 4.58\textbackslash 0.03 & 3.49\textbackslash 18.90\textbackslash 0.15 & -3.42\textbackslash -9.41\textbackslash -0.12 &  0.32\textbackslash 0.075\textbackslash -0.00 &  -0.17\textbackslash -2.56\textbackslash -0.03 \\
SD 1.5        &  \textbf{10.15\textbackslash 66.24\textbackslash 0.19} &  1.36\textbackslash 3.76\textbackslash 0.06  & 4.52\textbackslash 27.08\textbackslash 0.14 & -2.58\textbackslash -10.92\textbackslash -0.08 &  0.53\textbackslash -1.14\textbackslash 0.00 &  0.35\textbackslash -2.60\textbackslash -0.02 \\
SD 2.1     &  \textbf{12.43\textbackslash 79.42\textbackslash 0.24} &  2.43\textbackslash 8.47\textbackslash 0.07 & 7.84\textbackslash 37.50\textbackslash 0.20 & -2.56\textbackslash -10.15\textbackslash -0.06 &  1.51\textbackslash 1.93\textbackslash 0.03 &  0.29\textbackslash -0.56\textbackslash 0.00  \\
FLUX     &  \textbf{5.28\textbackslash 36.94\textbackslash 0.14} &  1.93\textbackslash 0.94\textbackslash 0.06 & -2.03\textbackslash -12.73\textbackslash -0.02 & -13.18\textbackslash -66.85\textbackslash -0.29 &  1.44\textbackslash 3.96\textbackslash 0.05  &  1.23\textbackslash 1.61\textbackslash 0.04 \\
A\&E        &  \textbf{9.75\textbackslash 50.13 \textbackslash 0.23} &  0.82\textbackslash -0.93\textbackslash 0.02  & 4.65\textbackslash 5.64\textbackslash 0.13 & -4.53\textbackslash -31.02\textbackslash -0.13 & 1.27\textbackslash 1.89\textbackslash 0.04  &  1.20\textbackslash 0.28\textbackslash-0.01 \\
StructDiff        &  \textbf{7.98\textbackslash 44.88\textbackslash 0.18} &  -0.81\textbackslash -7.70\textbackslash -0.02 & 3.92\textbackslash 2.97\textbackslash 0.04 & -4.80\textbackslash -27.28\textbackslash -0.16 &  0.1\textbackslash -2.43\textbackslash -0.03 &  -0.60\textbackslash -6.66 \textbackslash -0.03 \\
SynGen    &  \textbf{3.05\textbackslash 26.78\textbackslash 0.17} &  -0.29\textbackslash 0.80\textbackslash 0.04 & -1.44\textbackslash -1.80\textbackslash 0.01 & -9.62\textbackslash -38.78\textbackslash -0.25 &  1.85\textbackslash 9.56\textbackslash 0.12  &  0.95\textbackslash 5.52\textbackslash 0.09 \\
RichText &  \textbf{8.17\textbackslash 28.53\textbackslash 0.24} &  4.15\textbackslash 3.94\textbackslash 0.09 & 5.96\textbackslash 5.36\textbackslash 0.16 & -1.66\textbackslash -32.58\textbackslash -0.04 &  5.95\textbackslash 7.61\textbackslash 0.10 &  2.07\textbackslash 2.48\textbackslash 0.02 \\
BA     &  \textbf{8.03\textbackslash 61.89\textbackslash 0.18} &  2.91\textbackslash 13.29\textbackslash 0.06 & 1.29\textbackslash 13.47\textbackslash 0.12 & -9.09\textbackslash -44.34\textbackslash -0.15 &  0.43\textbackslash -0.95\textbackslash -0.01 &  0.032\textbackslash-3.20 \textbackslash-0.04 \\
\bottomrule
\end{tabular}
}
\caption{Color editing evaluation, comparing our approach to AnySD, MagicBrush, InstructPix2Pix, FPE and MasaCtrl. We report the absolute improvement in each metric (LAB\textbackslash RGB\textbackslash ACC); higher is better, and 0 means no improvement.
}
\label{tab:full_metrics_comparison_models}

\end{table}

%% file: tables_cvpr/image_reward.tex
\begin{table}[t]
\centering
\resizebox{\columnwidth}{!}{
\begin{tabular}{l|cccccc}
\toprule
\textbf{Method} & \textbf{Ours} & \textbf{AnySD} & \textbf{MagicBrush} & \textbf{InstructPix2Pix} & \textbf{FPE} & \textbf{MasaCtrl} \\
\midrule
\multicolumn{7}{c}{\textbf{Close}} \\
\midrule
SD 1.4       &  0.48 &  \textbf{0.51} & 0.38 & 0.38 & 0.48 &  0.45\\\
SD 1.5        &  0.63&  \textbf{0.67} & 0.41 & 0.50 & 0.62 &  0.53\\
SD 2.1     &  0.84&  \textbf{0.88} & 0.62 & 0.37 & 0.86 &  0.74 \\
FLUX     &  1.26 & 1.38 & 0.93 & 0.98 & \textbf{1.39} & 0.91 \\
A\&E        &  0.48 &  \textbf{0.51} & 0.14 & 0.11 & 0.50 &  0.37\\
StructDiff        &  0.53 &  \textbf{0.59}& 0.43 & 0.45 & 0.56 & 0.52\\
SynGen     & 0.69 &  0.50  & 0.28 & 0.24 & \textbf{0.71} &  \textbf{0.71}\\
RichText &  \textbf{0.29} &  0.21 & 0.06 & 0.23 & 0.25 &  -0.00\\
BA     &  0.47 & 0.57 & -0.57 & 0.16 & \textbf{0.59} & 0.44\\
\hline
Close-Mean & 0.63 & 0.65 & 0.30 & 0.38 & \textbf{0.66} & 0.52\\

\midrule
\multicolumn{7}{c}{\textbf{Distant}} \\
\midrule
SD 1.4        &  0.60 &  \textbf{0.66} & 0.13 & 0.15 & 0.61 &  0.54\\
SD 1.5        &  \textbf{0.65} &  0.58 & 0.18 & 0.18 &  0.60 & 0.53\\
SD 2.1     &  \textbf{0.79} &  0.78 & 0.41  & 0.40 & 0.79 &  0.61\\
FLUX     & 1.36 &  \textbf{1.45} & 0.87  & 0.73 & 1.44 &  1.22\\
A\&E        &  \textbf{0.71} &  0.61 & 0.02 & 0.00 & 0.66 &  0.58\\
StructDiff        &  0.59 &  \textbf{0.60} & 0.22 & 0.32 &  0.57&  0.52 \\
SynGen    &  0.92 &  0.80 & 0.27 &  0.06 & \textbf{0.98}&  0.95\\
RichText &  \textbf{0.38} &  0.15 & -0.02 & 0.01 & 0.22 &  -0.15\\
BA     &  0.53 &  0.58 &  -0.83 & -0.19&  \textbf{0.61} &  0.45\\
\hline
Distant-Mean & \textbf{0.73} & 0.69 & 0.14 & 0.18 & 0.72 & 0.58\\
\bottomrule
\end{tabular}
}
\caption{Evaluation using image reward, comparing our method with AnySD, MagicBrush, InstructPix2Pix, FPE and MasaCtrl. We report the image reward mean score; higher is better.
}
\label{tab:full_imagereward_comparison_models}

\end{table}

%% file: figures/colorpeel/fig_colorpeel.tex
\definecolor{firebrick}{rgb}{0.698, 0.133 0.133} 
\definecolor{tan}{rgb}{0.823, 0.705 0.549} 
\definecolor{hotpink}{rgb}{1, 0.411, 0.705} 
\definecolor{tomato}{rgb}{1, 0.388 0.278} 
\definecolor{ghostwhite}{rgb}{0.972, 0.972 1} 
\definecolor{paleturquoise}{rgb}{0.686, 0.933, 0.933} 
\definecolor{silver}{rgb}{0.752, 0.752, 0.752} 
\definecolor{lavenderblush}{rgb}{1, 0.94, 0.949} 
\definecolor{cornflowerblue}{rgb}{0.392, 0.584, 0.929} 
\definecolor{linen}{rgb}{0.98, 0.94, 0.90} 
\definecolor{powderblue}{rgb}{0.69, 0.878, 0.9} 
\definecolor{teal}{rgb}{0, 0.5, 0.5} 
\definecolor{aqua}{rgb}{0, 1, 1} 
\definecolor{blanchedalmond}{rgb}{1, 0.92, 0.8} 
\definecolor{whitesmoke}{rgb}{0.96, 0.96, 0.96} 
\definecolor{lightblue}{rgb}{0.678, 0.847, 0.9}

\begin{figure*}[htbp]
    \raggedleft 
    \begin{minipage}{0.32\textwidth}
        \centering
        \hspace{-0.2cm}
        \centering{\small{Initial}}
        \hspace{1.7cm}
        \centering{\small{+Ours}}
        
        \includegraphics[width=0.48\textwidth]{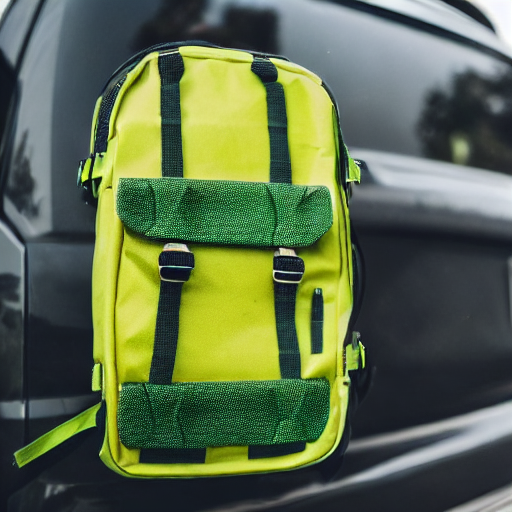}
        \includegraphics[width=0.48\textwidth]{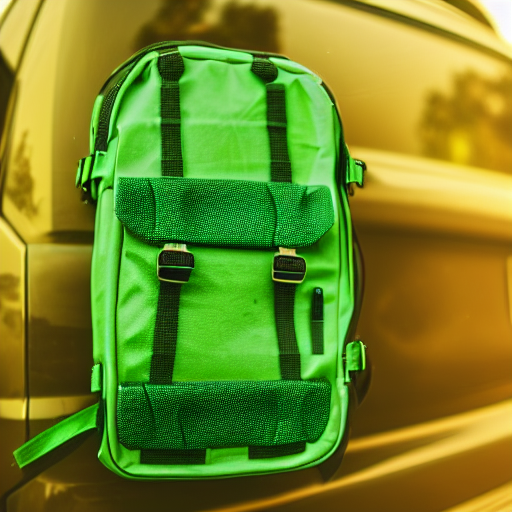} %
       {\footnotesize {"a backpack in \colorbox{green235}{\textbf{\textless c2*\textgreater}} color and a car in \colorbox{yellow235}{\textbf{\textless c4*\textgreater}} color"}} %
    \end{minipage}
    \begin{minipage}{0.32\textwidth}
        \centering
        \hspace{-0.2cm}
        \centering{\small{Initial}}
        \hspace{1.7cm}
        \centering{\small{+Ours}}
        
        \includegraphics[width=0.48\textwidth]{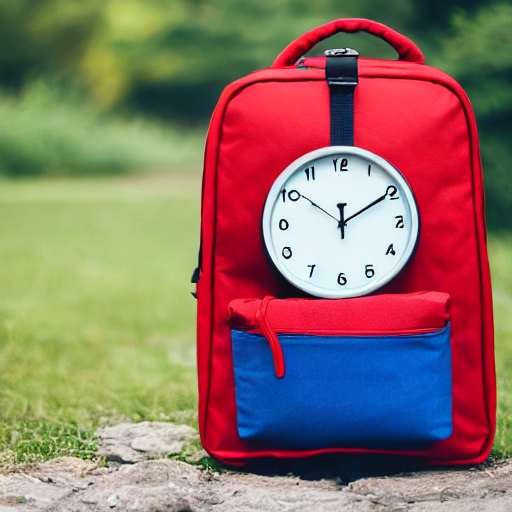}
        \includegraphics[width=0.48\textwidth]{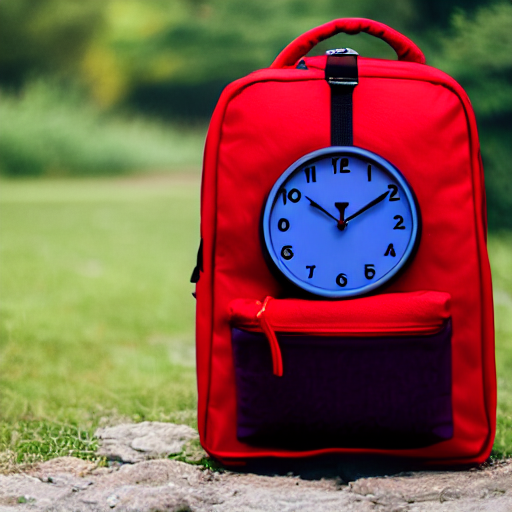} %
       {\footnotesize {"a backpack in \colorbox{red235}{\textbf{\textless c1*\textgreater}} color and a clock in \colorbox{blue235}{\textcolor{white}{\textbf{\textless c3*\textgreater}}} color"}} %
    \end{minipage}
    \begin{minipage}{0.32\textwidth}
        \centering
        \hspace{-0.2cm}
        \centering{\small{Initial}}
        \hspace{1.7cm}
        \centering{\small{+Ours}}
        
        \includegraphics[width=0.48\textwidth]{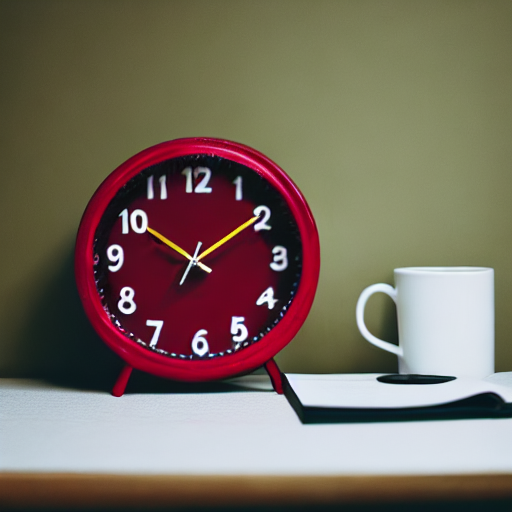} 
        \includegraphics[width=0.48\textwidth]{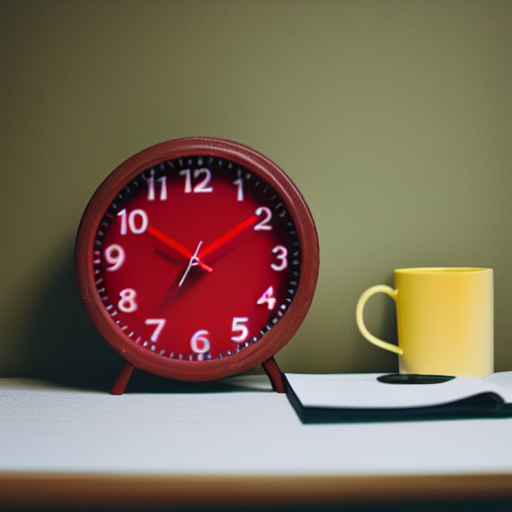} 
       {\footnotesize {"a clock in \colorbox{red235}{\textbf{\textless c1*\textgreater}} color and a mug in \colorbox{yellow235}{\textbf{\textless c4*\textgreater}} color"}} %
    \end{minipage}
\vspace{-8pt}
     \caption{Images generated using ColorPeel, before and after being edited by our approach. As can be observed, ColorPeel is challenged by our multi-color multi-object setting; see Section \ref{sec:colorPeel} for additional details. }
    \label{fig:colorpeel_fig}
\end{figure*}

%% file: tables_cvpr/runtime_comparison.tex
\begin{table}[t]
\centering
\resizebox{\columnwidth}{!}{
\begin{tabular}{cccccc}
\hline
\textbf{ColorEdit-Ours} & \textbf{AnySD} & \textbf{MagicBrush} & \textbf{InstructPix2Pix} & \textbf{FPE} & \textbf{MasaCtrl}\\
\hline
70 & 11 & 25 & 25 & 20 & 22 \\
\bottomrule
\end{tabular}
}
\caption{Runtime comparison (in seconds) between our method and AnySD, MagicBrush, InstructPix2Pix, FPE and
MasaCtrl.}
\label{tab:runtime_comparison}
\end{table}

%% file: rebuttal_files/inversion/inversion.tex
\begin{table}[t]
\centering
\resizebox{\columnwidth}{!}{
\begin{tabular}{lcc}
\hline
\textbf{Method} & \textbf{Close-LAB\textbackslash RGB\textbackslash ACC(Imp $\uparrow$)} & \textbf{Distant}\\
\hline
w Null-Text       &  \textbf{8.39\textbackslash54.24\textbackslash0.29} & \textbf{12.91\textbackslash79.57\textbackslash0.28}\\
Ours (DDIM)         &  7.43\textbackslash51.02\textbackslash0.22 & 12.43\textbackslash 79.42\textbackslash0.24\\
\hline
\end{tabular}
}
\caption{Influence of different inversion methods on our method performance, conducted on images generated by SD 2.1.}
\label{tab:ablation_inversion}
\end{table}

%% file: tables_cvpr/no_inversion.tex
\begin{table}[t]

\centering
\resizebox{\columnwidth}{!}{
\begin{tabular}{l|cc}
\toprule
\textbf{Method} & \textbf{Inversion(Imp $\uparrow$)} & \textbf{w/o Inversion(Imp $\uparrow$)} \\
\midrule
\multicolumn{3}{c}{\textbf{Close}} \\
\midrule
SD 1.5        &  \textbf{6.04\textbackslash 38.39}\textbackslash 0.18  & 3.75\textbackslash 37.07\textbackslash\textbf{0.21} \\
\midrule
\multicolumn{3}{c}{\textbf{Distant}} \\
\midrule
SD 1.5        &  \textbf{10.15\textbackslash 66.24\textbackslash 0.19}  &  7.00\textbackslash 54.97\textbackslash 0.18 \\
\bottomrule
\end{tabular}
}
\caption{ColorEdit performance (LAB\textbackslash RGB\textbackslash ACC) with and without inversion; see Section \ref{sec:no_inversion} for additional details. %
}
\label{tab:no_inversion}
\end{table}

%% file: figures/no_inversion/no_inversion.tex
\begin{figure}[t]
     \centering
    \begin{minipage}{0.48\columnwidth}
        \centering
        {SD1.5}
        \includegraphics[width=0.98\columnwidth]{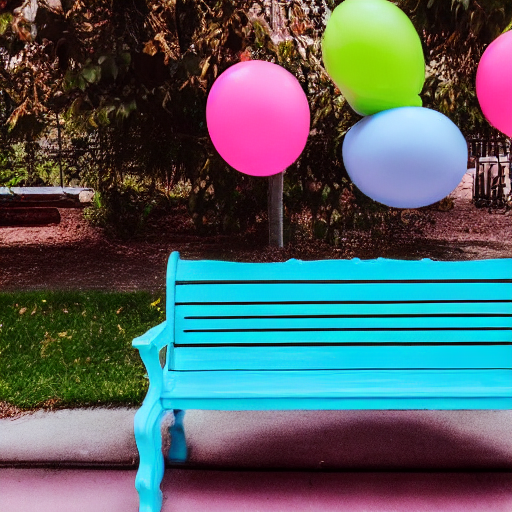} 
    \end{minipage}
    \begin{minipage}{0.48\columnwidth}
        \centering
        {+Ours w/o Inversion}
        \includegraphics[width=0.98\columnwidth]{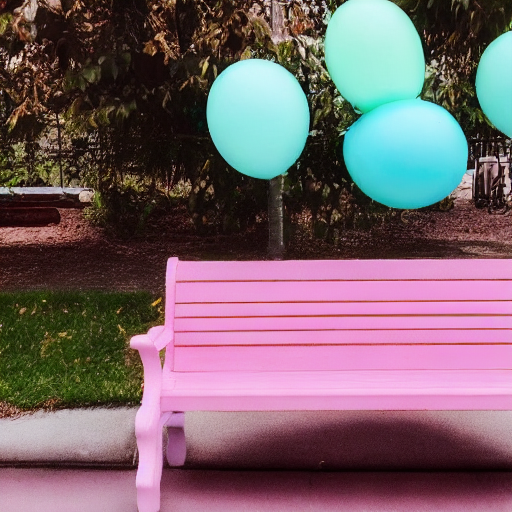}  %
    \end{minipage}
    \vspace{0.1cm}
    \hspace{0.1cm}
    {\footnotesize {"a \colorbox{paleturquoise}{pale-turquoise} colored balloon and a \colorbox{hotpink}{{\textcolor{white}{hot-pink}}} colored bench"}} %
    \caption{Editing results using our method without inversion. Images are generated with SD1.5, showing that our approach remains effective even without inversion.}
    \hspace{0.3cm}
    \label{fig:no_inversion}
\end{figure}

%% file: tables_cvpr/overlap_objects.tex
\captionsetup[table]{font=footnotesize}
\begin{table}[t]
\scriptsize
\centering
\resizebox{\columnwidth}{!}{
\begin{tabular}{l|ccc}
\toprule
\textbf{Method} & \textbf{Iou$>$ 0.5}& \textbf{0.1$<$ Iou $\leq$ 0.5} & \textbf{Iou $\leq$ 0.1}\\
\midrule
\multicolumn{4}{c}{\textbf{Close}} \\
\midrule
SD 1.4       &  50.00 \textcolor{darkgreen}{13.00\%$\uparrow$} \textbackslash(0.12) & 58.62 \textcolor{darkgreen}{18.39\%$\uparrow$} \textbackslash(0.44) & 61.80 \textcolor{darkgreen}{25.84\%$\uparrow$} \textbackslash(0.45) \\
SD 1.5       &  41.18 \textcolor{darkgreen}{0.00\%$\uparrow$} \textbackslash(0.09) & 51.69 \textcolor{darkgreen}{17.98\%$\uparrow$} \textbackslash(0.49) & 60.81 \textcolor{darkgreen}{22.97\%$\uparrow$} \textbackslash(0.41)\\
SD 2.1       &  50.00 \textcolor{darkgreen}{27.78\%$\uparrow$} \textbackslash(0.08) & 57.65 \textcolor{darkgreen}{18.82\%$\uparrow$} \textbackslash(0.36) & 54.62 \textcolor{darkgreen}{23.85\%$\uparrow$} \textbackslash(0.56)\\
FLUX     &  76.92 \textcolor{darkgreen}{7.69\%$\uparrow$}\textbackslash(0.04)  & 70.91 \textcolor{darkgreen}{11.82\%$\uparrow$}\textbackslash(0.35) & 68.42 \textcolor{darkgreen}{18.95\%$\uparrow$}\textbackslash(0.61)\\
A\&E     &  42.31 \textcolor{darkgreen}{0.00\%$\uparrow$}\textbackslash(0.08)  & 54.92 \textcolor{darkgreen}{6.56\%$\uparrow$}\textbackslash(0.37) & 67.21 \textcolor{darkgreen}{22.40\%$\uparrow$}\textbackslash(0.55)\\
StructDiff     &  58.82 \textcolor{darkgreen}{17.65\%$\uparrow$}\textbackslash(0.09)  & 63.11 \textcolor{darkgreen}{19.42\%$\uparrow$}\textbackslash(0.53) & 62.67 \textcolor{darkgreen}{29.33\%$\uparrow$}\textbackslash(0.38)\\
SynGen     &  63.89 \textcolor{darkgreen}{16.67\%$\uparrow$}\textbackslash(0.13)  & 70.93 \textcolor{darkgreen}{25.58\%$\uparrow$}\textbackslash(0.63) & 76.56 \textcolor{darkgreen}{17.19\%$\uparrow$}\textbackslash(0.24) \\
RichText     &  - -\textbackslash-  & 47.56 \textcolor{darkgreen}{20.73\%$\uparrow$}\textbackslash(0.57) & 70.97 \textcolor{darkgreen}{25.81\%$\uparrow$}\textbackslash(0.43) \\
BA     &  0.00 \textcolor{darkgreen}{0.00\%$\uparrow$}\textbackslash(0.003)  & 67.86 \textcolor{darkgreen}{32.14\%$\uparrow$}\textbackslash(0.08) & 66.47 \textcolor{darkgreen}{9.79\%$\uparrow$}\textbackslash(0.92) \\

\midrule
\multicolumn{4}{c}{\textbf{Distant}} \\
\midrule
SD 1.4       &  31.25 \textcolor{darkgreen}{6.25\%$\uparrow$} \textbackslash(0.08) & 50.52 \textcolor{darkgreen}{24.74\%$\uparrow$} \textbackslash(0.48) & 47.78 \textcolor{darkgreen}{16.67\%$\uparrow$} \textbackslash(0.44) \\
SD 1.5       &  53.85 \textcolor{darkgreen}{30.77\%$\uparrow$} \textbackslash(0.06) & 44.55 \textcolor{darkgreen}{14.85\%$\uparrow$} \textbackslash(0.45) & 51.38 \textcolor{darkgreen}{21.10\%$\uparrow$} \textbackslash(0.49)\\
SD 2.1       &  56.25 \textcolor{darkgreen}{18.75\%$\uparrow$} \textbackslash(0.06) & 50.00 \textcolor{darkgreen}{22.22\%$\uparrow$} \textbackslash(0.40) & 48.61 \textcolor{darkgreen}{25.69\%$\uparrow$} \textbackslash(0.54)\\
FLUX     &  75.00 \textcolor{darkgreen}{12.50\%$\uparrow$}\textbackslash(0.04)  & 66.01 \textcolor{darkgreen}{13.07\%$\uparrow$}\textbackslash(0.40) & 59.33 \textcolor{darkgreen}{14.83\%$\uparrow$}\textbackslash(0.55)\\
A\&E     &  52.00 \textcolor{darkgreen}{8.00\%$\uparrow$}\textbackslash(0.07)  & 65.93 \textcolor{darkgreen}{17.78\%$\uparrow$}\textbackslash(0.36) & 59.55 \textcolor{darkgreen}{27.73\%$\uparrow$}\textbackslash(0.58)\\
StructDiff     & 43.75 \textcolor{darkgreen}{0.00\%$\uparrow$}\textbackslash(0.07)  & 58.26 \textcolor{darkgreen}{15.65\%$\uparrow$}\textbackslash(0.52) & 57.30 \textcolor{darkgreen}{23.60\%$\uparrow$}\textbackslash(0.40)\\
SynGen     &  75.47 \textcolor{darkgreen}{13.21\%$\uparrow$}\textbackslash(0.16)  & 73.99 \textcolor{darkgreen}{17.94\%$\uparrow$}\textbackslash(0.66) & 66.13 \textcolor{darkgreen}{14.52\%$\uparrow$}\textbackslash(0.18) \\
RichText     &  0.00 \textcolor{darkgreen}{0.00\%$\uparrow$}\textbackslash(0.01)  & 43.48 \textcolor{darkgreen}{17.39\%$\uparrow$}\textbackslash(0.54) & 67.53 \textcolor{darkgreen}{32.47\%$\uparrow$}\textbackslash(0.45) \\
BA     & 100.0 \textcolor{darkgreen}{0.00\%$\uparrow$}\textbackslash(0.003)  & 62.16 \textcolor{darkgreen}{27.02\%$\uparrow$}\textbackslash(0.10) & 45.70 \textcolor{darkgreen}{17.51\%$\uparrow$}\textbackslash(0.90)\\
\bottomrule
\end{tabular}
}
\vspace{-8pt}
\caption{ColorEdit evaluation with respect to different overlap levels between the objects. We report the absolute improvement in Accuracy(ACC); higher is better, and 0 means no
improvement. In brackets we report the relative amount of images with such overlap.
}
\label{tab:overlap_objects}

\end{table}

%% file: figures/user_study_example.tex
\begin{figure}[t]
     \centering
    \begin{minipage}{0.96\columnwidth}
        \centering
        \includegraphics[width=0.98\columnwidth]{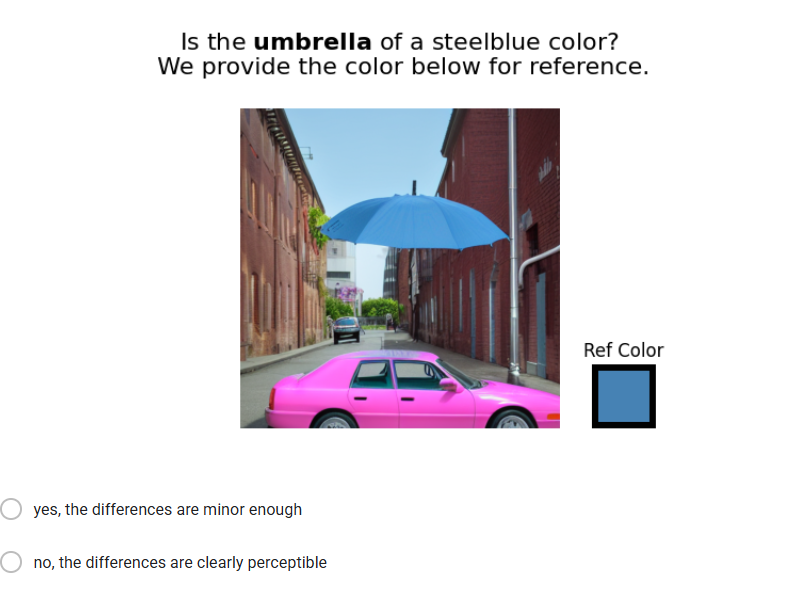} 
    \end{minipage}
    \caption{Example user study question: participants were asked whether the difference between the reference object’s color and the color shown on the right was minor or clearly perceptible.}
    \label{fig:user_study_example}
\end{figure}

%% file: tables_cvpr/user_study_stats.tex
\begin{table}[t]
\centering
\resizebox{\columnwidth}{!}{
\begin{tabular}{l|ccccc}
\hline
\textbf{Metric} & \textbf{Th-5} & \textbf{Th-10} & \textbf{Th-15} & \textbf{Th-20} & \textbf{Th-25}\\
\hline
TPR \textbackslash TNR & 0.29\textbackslash \textbf{1.0} & 0.64\textbackslash 0.69 & 0.79 \textbackslash 0.38 & 0.79 \textbackslash 0.31 & \textbf{0.86} \textbackslash 0.25  \\
AUC & 0.64 & \textbf{0.67} & 0.58 & 0.55 & 0.55\\
G-mean & 0.53 &  \textbf{0.66} & 0.54 & 0.5 & 0.46 \\
Balanced Accuracy & 0.64 &  \textbf{0.67} & 0.58 & 0.55 & 0.55\\
\hline
\end{tabular}
}
\caption{User study statistics in comparison with different thresholds, we see that threshold 10 is providing the best statistics in terms of AUC, G-mean and Balanced Accuracy.}
\label{tab:user_study_stats}
\end{table}

%% file: figures/SOTA/SOTA.tex
\begin{figure*}[t]
    \raggedright
       \hspace{-0.17cm} 
      \raisebox{-0.9cm}{\rotatebox{90}{\textcolor{black}{\bfseries OpenAI 4o}}} %
    \begin{minipage}{0.135\textwidth}
        \centering
    \raisebox{0.1cm}{\textbf{Source}} %
        \includegraphics[width=\textwidth]{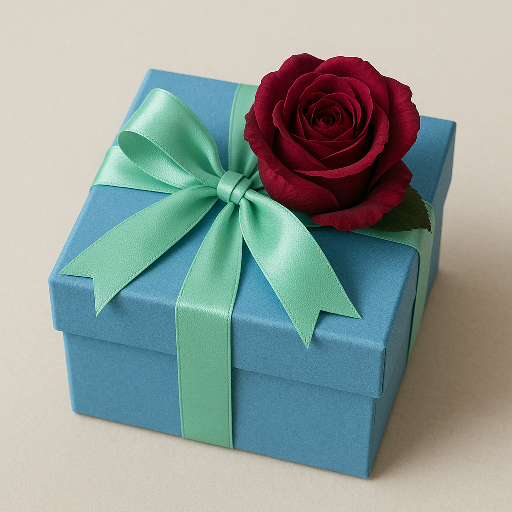} 
    \end{minipage}
    \hspace{0.05cm} 
    \begin{minipage}{0.135\textwidth}
        \centering
       \raisebox{0.1cm}{\textbf{{+Ours}}} %
        \includegraphics[width=\textwidth]{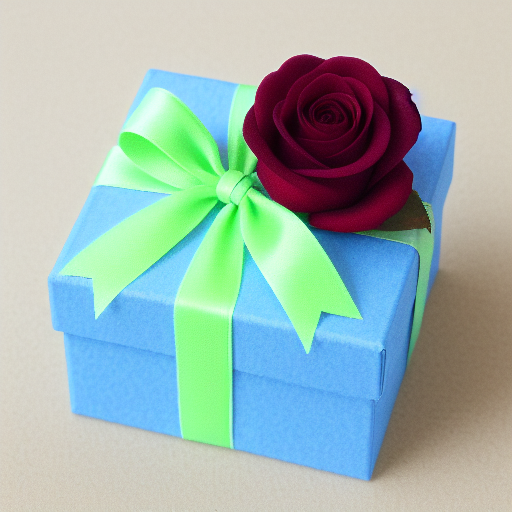}
    \end{minipage}
    \begin{minipage}{0.135\textwidth}
        \centering
        \textbf{+AnySD}
        \vspace{-0.04cm}
        \includegraphics[width=\textwidth]{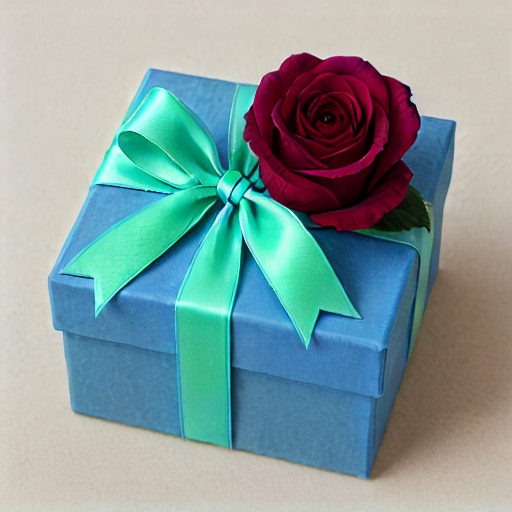}
    \end{minipage}
    \begin{minipage}{0.135\textwidth}
        \centering
        \textbf{+MagicBrush}
        \vspace{-0.04cm}
        \includegraphics[width=\textwidth]{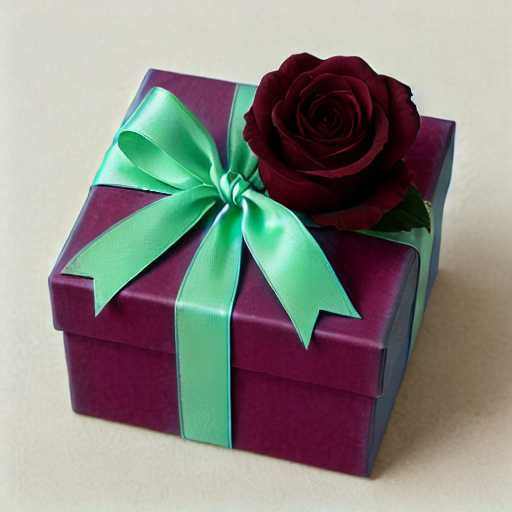}
    \end{minipage}
    \begin{minipage}{0.135\textwidth}
        \centering
        \raisebox{0.1cm}{\textbf{+InstPix2Pix}} 
        \includegraphics[width=\textwidth]{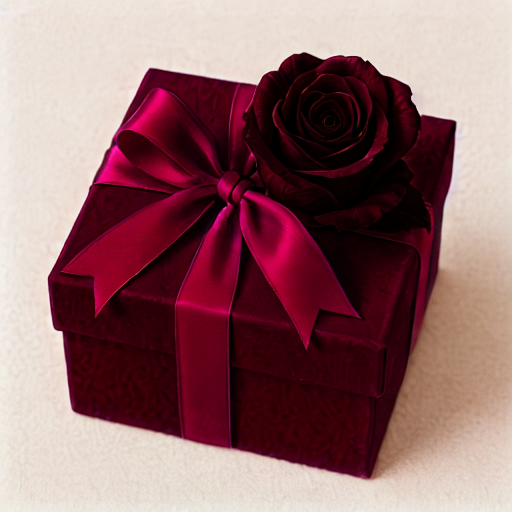}
    \end{minipage}
    \begin{minipage}{0.135\textwidth}
        \centering
        \raisebox{0.1cm}{\textbf{+FPE} }      
        \includegraphics[width=\textwidth]{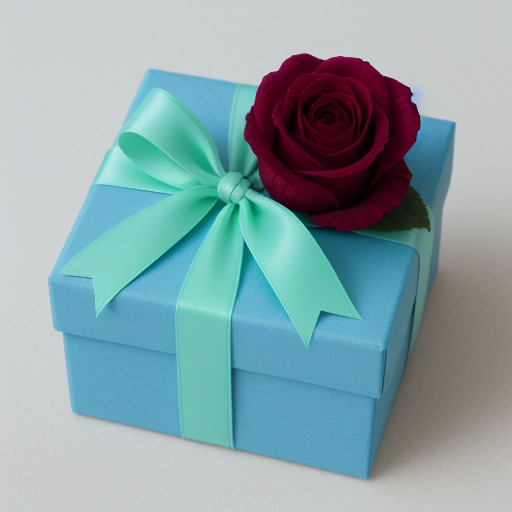}
    \end{minipage}
    \vspace{-0.01cm}
    \begin{minipage}{0.135\textwidth}
        \centering
        \raisebox{0.1cm}{\textbf{+MasaCtrl}}
        \includegraphics[width=\textwidth]{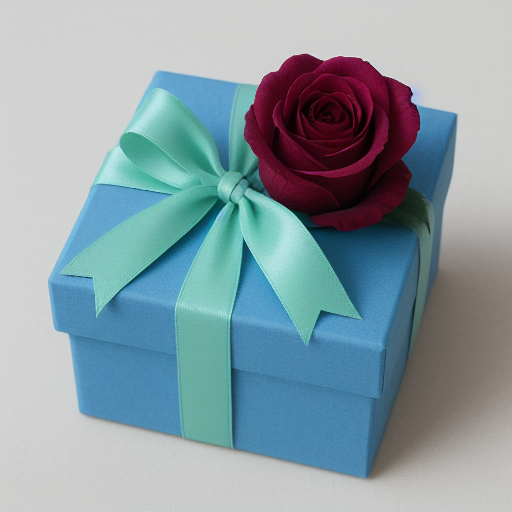}
    \end{minipage}
    \vspace{-0.015cm}
    \hspace{-0.1cm} 
    \raisebox{-0.7cm}{\rotatebox{90}{\textcolor{black}{\bfseries Gemini2.5}}} %
    \begin{minipage}{0.135\textwidth}
        \centering
        \vspace{+0.05cm}
        \includegraphics[width=\textwidth]{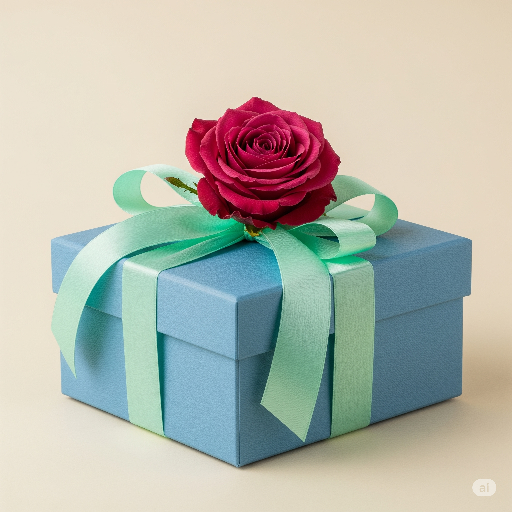} 
    \end{minipage}
    \hspace{0.05cm} 
    \begin{minipage}{0.135\textwidth}
        \centering
        \vspace{+0.05cm}
        \includegraphics[width=\textwidth]{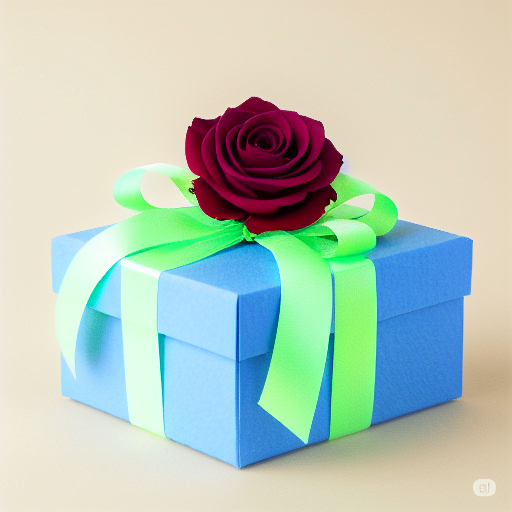} 
    \end{minipage}
    \begin{minipage}{0.135\textwidth}
        \centering
        \vspace{+0.05cm}
        \includegraphics[width=\textwidth]{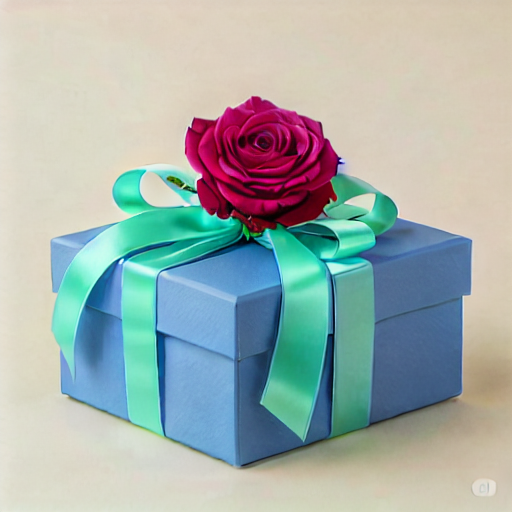} 
    \end{minipage}
    \begin{minipage}{0.135\textwidth}
        \centering
        \vspace{+0.05cm}
        \includegraphics[width=\textwidth]{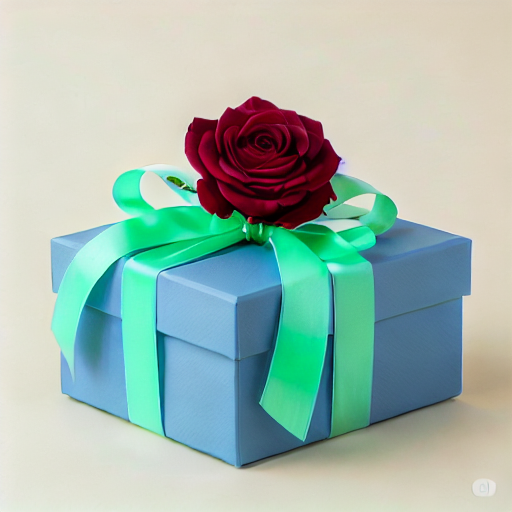} 
    \end{minipage}
    \begin{minipage}{0.135\textwidth}
        \centering
        \vspace{+0.05cm}
        \includegraphics[width=\textwidth]{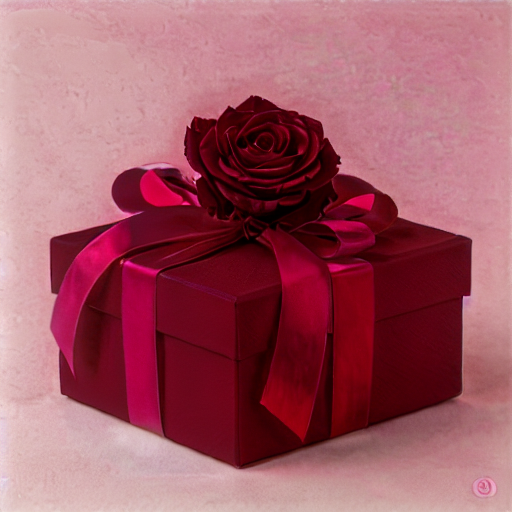} 
    \end{minipage}
    \begin{minipage}{0.135\textwidth}
        \centering
        \vspace{+0.05cm}
        \includegraphics[width=\textwidth]{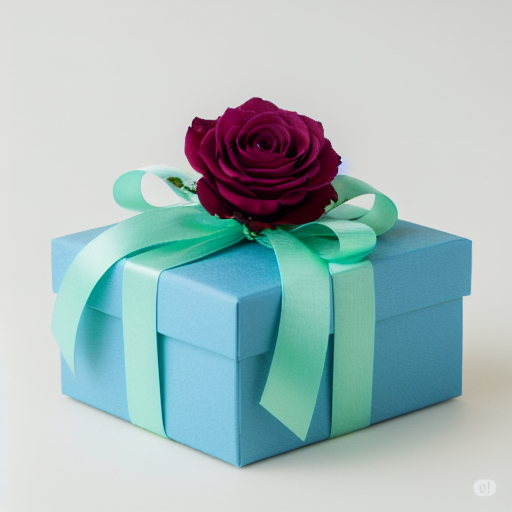} 
    \end{minipage}
    \begin{minipage}{0.135\textwidth}
        \centering
        \vspace{+0.05cm}
        \includegraphics[width=\textwidth]{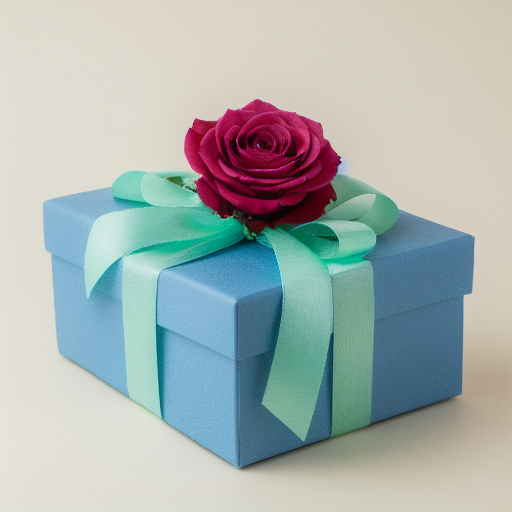}
    \end{minipage}
    \vspace{-0.015cm}
    \hspace{-0.1cm} 
      \raisebox{-0.4cm}{\rotatebox{90}{\textcolor{black}{\bfseries Grok3}}} %
    \begin{minipage}{0.135\textwidth}
        \centering
        \vspace{+0.05cm}
        \includegraphics[width=\textwidth]{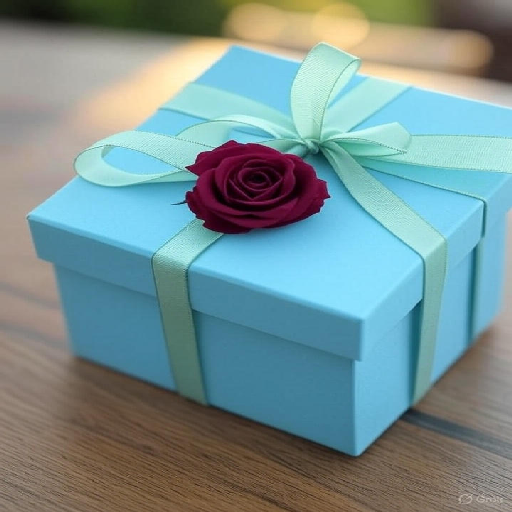}
    \end{minipage}
        \hspace{0.05cm} 
    \begin{minipage}{0.135\textwidth}
        \centering
        \vspace{+0.05cm}
        \includegraphics[width=\textwidth]{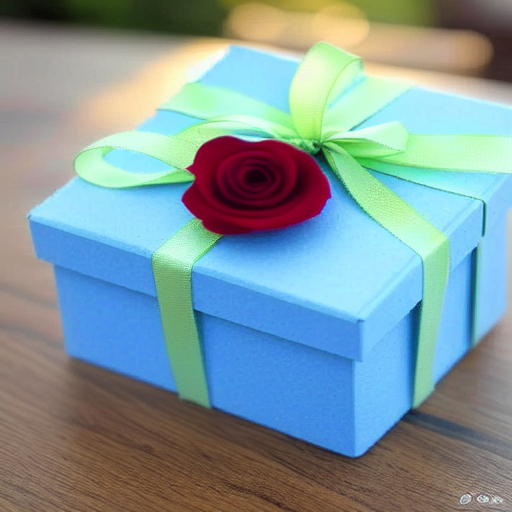}
    \end{minipage}
    \begin{minipage}{0.135\textwidth}
        \centering
        \vspace{+0.05cm}
        \includegraphics[width=\textwidth]{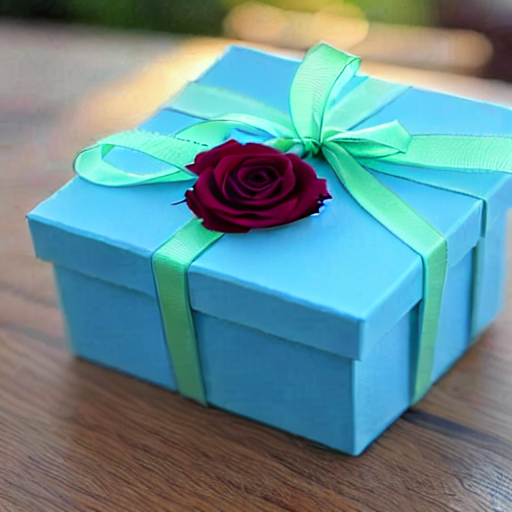}
    \end{minipage}
    \begin{minipage}{0.135\textwidth}
        \centering
        \vspace{+0.05cm}
        \includegraphics[width=\textwidth]{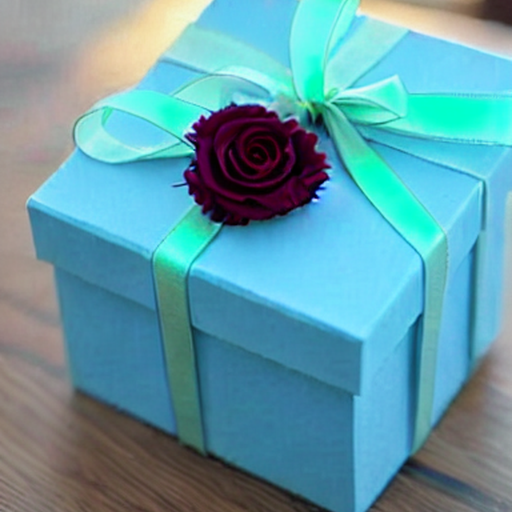}
    \end{minipage}
    \begin{minipage}{0.135\textwidth}
        \centering
        \vspace{+0.05cm}
        \includegraphics[width=\textwidth]{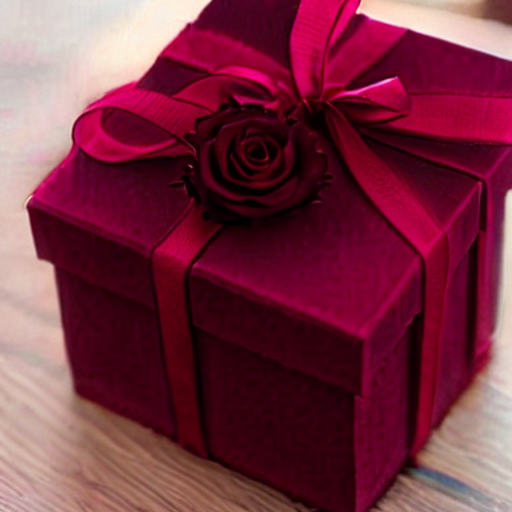}
    \end{minipage}
    \begin{minipage}{0.135\textwidth}
        \centering
        \vspace{+0.05cm}
        \includegraphics[width=\textwidth]{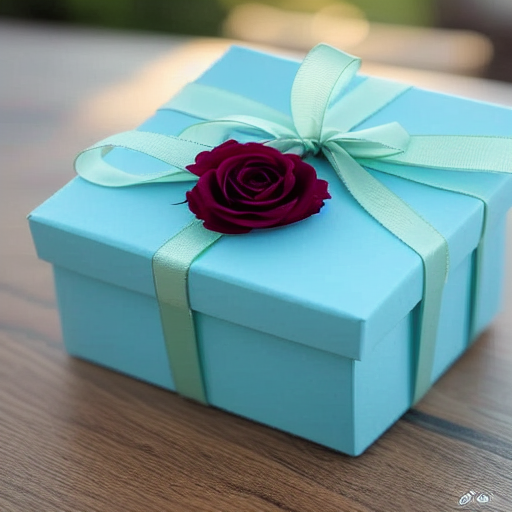}
    \end{minipage}
    \begin{minipage}{0.135\textwidth}
        \centering
        \vspace{+0.05cm}
        \includegraphics[width=\textwidth]{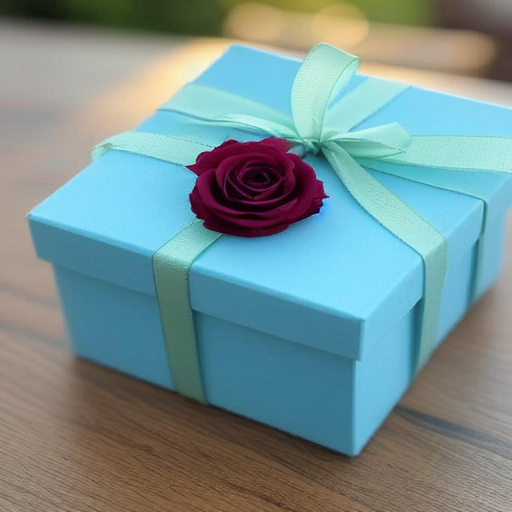}
    \end{minipage}
    \vspace{0.1cm}

    \centering
    {\footnotesize {"a \colorbox{cornflowerblue}{\textcolor{white}{cornflower-blue}} gift box tied with a \colorbox{mintgreen}{\textcolor{white}{mint-green}} ribbon and a \colorbox{burgundy}{\textcolor{white}{burgundy}} rose"}} %
    \caption{Propriety model performance on the prompt shown in Figure 1 (main paper). As shown, even the latest VLMs, including OpenAI 4O, Gemini 2.5, and Grok 3, often struggle to capture the correct semantics associated with multiple colors. Our method consistently outperforms existing editing approaches such as AnySD~\cite{yu2025anyeditmasteringunifiedhighqualityz}, MagicBrush~\cite{zhang2024magicbrushmanuallyannotateddataset}, InstructPix2Pix~\cite{brooks2023instructpix2pixlearningfollowimage}, FPE~\cite{liu2024understandingcrossselfattentionstable}, and MasaCtrl~\cite{Cao_2023_ICCV}, which exhibit limitations in addressing such semantic misalignment. MagicBrush and InstructPix2Pix process the input image differently resulting scale change compare to other models with respect to grok3 image.}
    \label{fig:SOTA_example}
\end{figure*}

%% file: rebuttal_files/3_colors/3_colors.tex
\captionsetup[table]{font=footnotesize}
\begin{table}[t]
\centering
\resizebox{\columnwidth}{!}{
\begin{tabular}{lcc}
\hline
\textbf{Method} & \textbf{Base - Close \& Distant} & \textbf{Edited} \\
\hline
SD 2.1 & 25.39\textbackslash 135.36\textbackslash 0.23 & \textbf{12.39\textbackslash 51.79\textbackslash 0.52} \\
FLUX & 13.96\textbackslash 73.97\textbackslash 0.48 & \textbf{10.13\textbackslash 39.11\textbackslash 0.65} \\
SynGen & 13.11\textbackslash 70.12\textbackslash 0.56 &  \textbf{9.99\textbackslash 43.47\textbackslash 0.64} \\
\hline
\end{tabular}
}
\caption{Performance over three colors prompts, containing close and distant colors, LAB($\downarrow$)\textbackslash RGB($\downarrow$)\textbackslash ACC($\uparrow$) metrics.}
\label{tab:reviewer3_3_colors}
\end{table}

%% file: figures/three_colors/three_colors.tex
\begin{figure}[t]
     \centering
    \begin{minipage}{0.48\columnwidth}
        \centering
        {FLUX}
        \includegraphics[width=0.98\columnwidth]{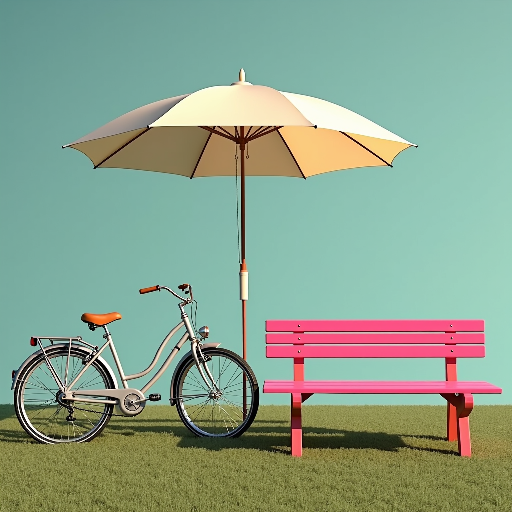} 
    \end{minipage}
    \begin{minipage}{0.48\columnwidth}
        \centering
        {+Ours}
        \includegraphics[width=0.98\columnwidth]{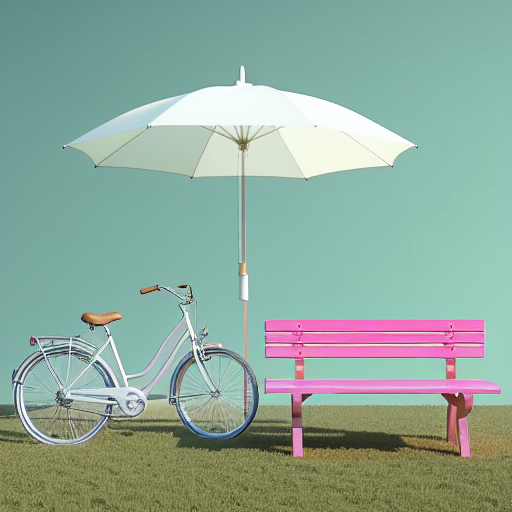}  %
    \end{minipage}
    \vspace{0.1cm}
    \hspace{0.1cm}
    {\footnotesize {"a \colorbox{ivory}{ivory} colored umbrella and a \colorbox{silver}{{\textcolor{white}{silver}}} colored bicycle and a \colorbox{hotpink}{{\textcolor{white}{hot-pink}}} colored bench"}} %
    \caption{Our editing approach with a full text prompt containing three objects. The initial image is generated using FLUX.}
    \hspace{0.3cm}
    \label{fig:three_colors}
\end{figure}

%% file: figures/real_images/real_images.tex
\definecolor{firebrick}{rgb}{0.698, 0.133 0.133} 
\definecolor{tan}{rgb}{0.823, 0.705 0.549} 
\definecolor{hotpink}{rgb}{1, 0.411, 0.705} 
\definecolor{tomato}{rgb}{1, 0.388 0.278} 
\definecolor{ghostwhite}{rgb}{0.972, 0.972 1} 
\definecolor{paleturquoise}{rgb}{0.686, 0.933, 0.933} 
\definecolor{silver}{rgb}{0.752, 0.752, 0.752} 
\definecolor{lavenderblush}{rgb}{1, 0.94, 0.949} 
\definecolor{cornflowerblue}{rgb}{0.392, 0.584, 0.929} 
\definecolor{linen}{rgb}{0.98, 0.94, 0.90} 
\definecolor{powderblue}{rgb}{0.69, 0.878, 0.9} 
\definecolor{teal}{rgb}{0, 0.5, 0.5} 
\definecolor{aqua}{rgb}{0, 1, 1} 
\definecolor{blanchedalmond}{rgb}{1, 0.92, 0.8} 
\definecolor{whitesmoke}{rgb}{0.96, 0.96, 0.96} 
\definecolor{lightblue}{rgb}{0.678, 0.847, 0.9} 
\definecolor{green128}{rgb}{0, 0.5, 0}

\begin{figure*}[htbp]
    \centering 
    \begin{minipage}{0.0695\textwidth}
        \centering
                             \centering
              \raisebox{0.05cm}{\tiny Source Image}
         \vspace{+0.5cm}
         \includegraphics[width=0.95\textwidth]{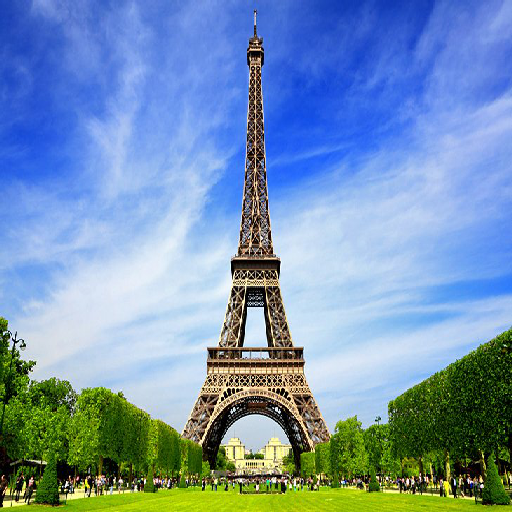} 
    \end{minipage}
    \hspace{0.05cm}
    \begin{minipage}{0.435\textwidth}
        \begin{minipage}{0.155\textwidth}
            \centering
              \raisebox{0.05cm}{\tiny +Ours}
            \includegraphics[width=0.98\textwidth]{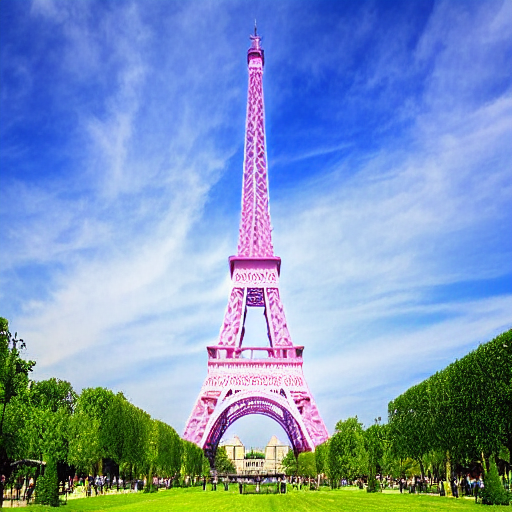} 
        \end{minipage}
        \begin{minipage}{0.155\textwidth}
            \centering
             \raisebox{0.05cm}{\tiny{+AnySD}}       \includegraphics[width=0.98\textwidth]{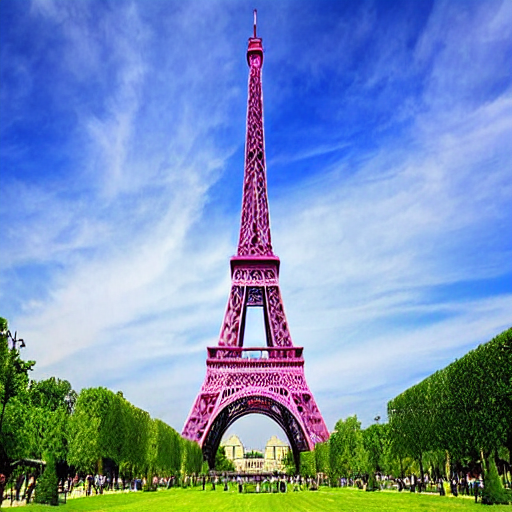} 
        \end{minipage}
                \begin{minipage}{0.155\textwidth}
            \centering
             \raisebox{0.05cm}{\tiny{+MagicBrush}}       \includegraphics[width=0.98\textwidth, height=0.98\textwidth]{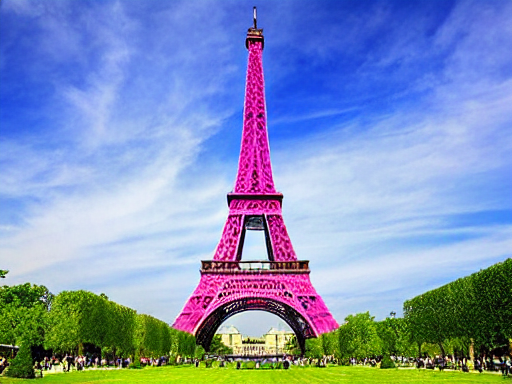} 
        \end{minipage}
                \begin{minipage}{0.155\textwidth}
            \centering
             \raisebox{0.05cm}{\tiny{+InstPix2Pix}}       \includegraphics[width=0.98\textwidth, height=0.98\textwidth]{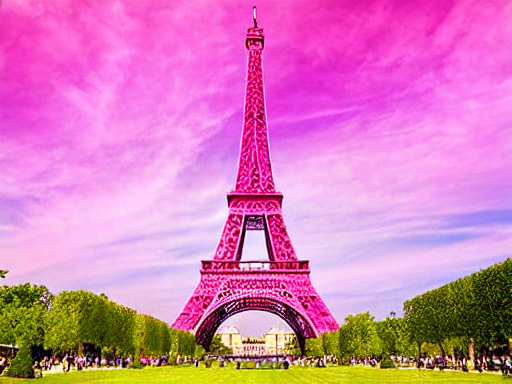} 
        \end{minipage}
        \begin{minipage}{0.155\textwidth}
            \centering
              \raisebox{0.05cm}{\tiny +FPE}
            \vspace{-0.01cm}\includegraphics[width=0.98\textwidth]{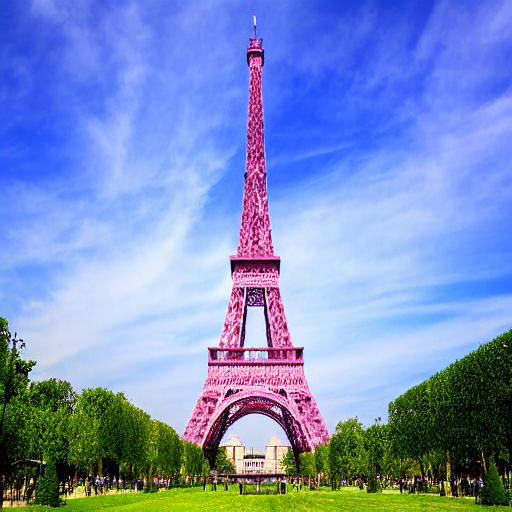} 
        \end{minipage}
        \begin{minipage}{0.155\textwidth}
            \centering
              \raisebox{0.05cm}{\tiny +MasaCtrl}
            \vspace{-0.05cm}\includegraphics[width=0.98\textwidth]{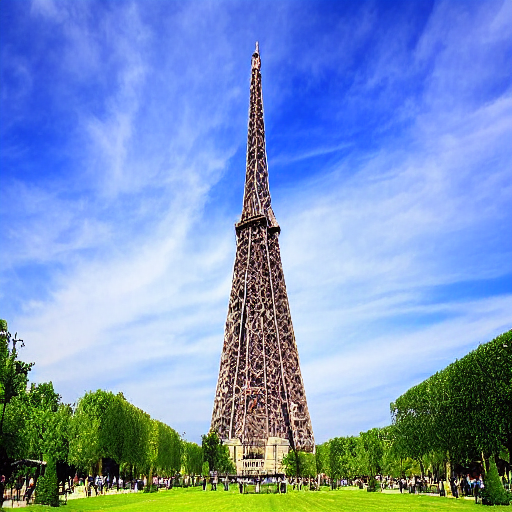} 
        \end{minipage}
        \centering
        {\footnotesize {"a \colorbox{pink}{pink} tower"}}
    \end{minipage}
    \hspace{0.05cm}
    \begin{minipage}{0.435\textwidth}
        \begin{minipage}{0.155\textwidth}
            \centering
              \raisebox{0.05cm}{\tiny +Ours}
            \includegraphics[width=0.98\textwidth]{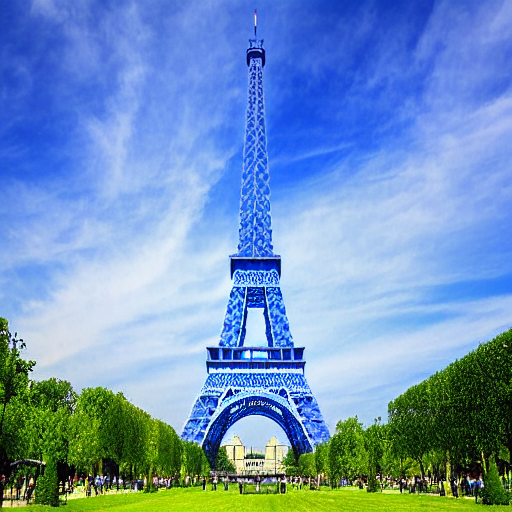} 
        \end{minipage}
        \begin{minipage}{0.155\textwidth}
            \centering
             \raisebox{0.05cm}{\tiny +AnySD}
            \includegraphics[width=0.98\textwidth]{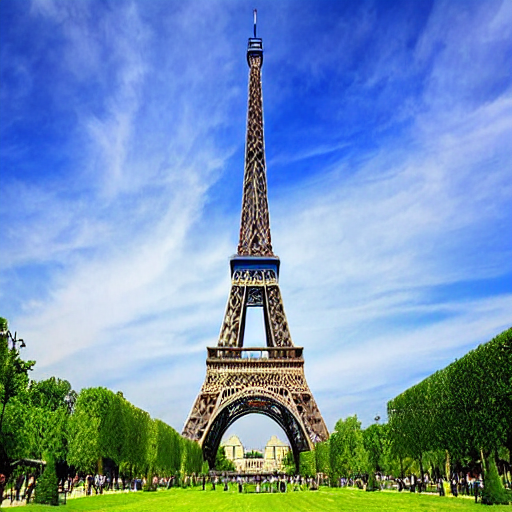} 
        \end{minipage}
                \begin{minipage}{0.155\textwidth}
            \centering
             \raisebox{0.05cm}{\tiny +MagicBrush}
            \includegraphics[width=0.98\textwidth, height=0.98\textwidth]{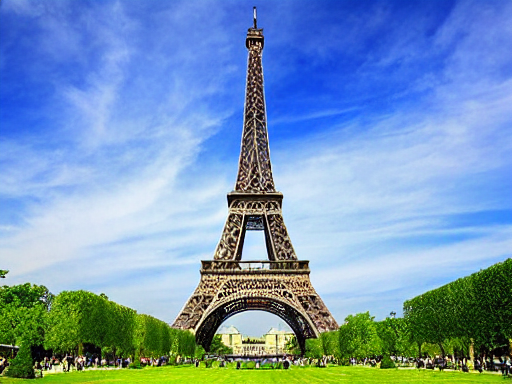} 
        \end{minipage}
                \begin{minipage}{0.155\textwidth}
            \centering
             \raisebox{0.05cm}{\tiny +InstPix2Pix}
            \includegraphics[width=0.98\textwidth, height=0.98\textwidth]{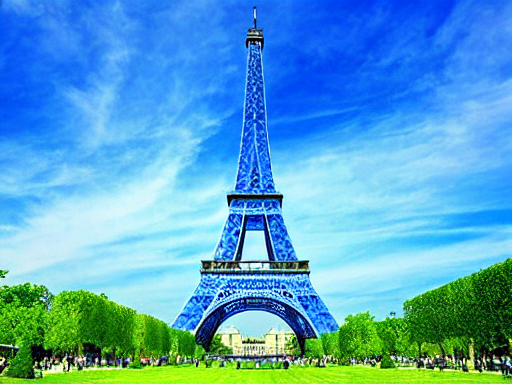} 
        \end{minipage}
        \begin{minipage}{0.155\textwidth}
            \centering
              \raisebox{0.05cm}{\tiny +FPE}
            \vspace{-0.01cm}\includegraphics[width=0.98\textwidth]{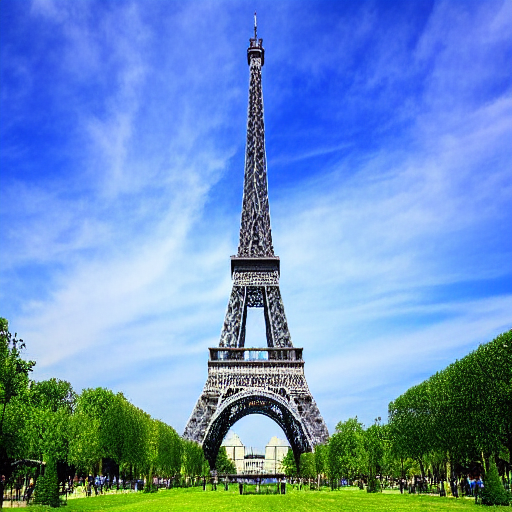} 
        \end{minipage}
        \begin{minipage}{0.155\textwidth}
            \centering
              \raisebox{0.05cm}{\tiny +MasaCtrl}
            \vspace{-0.01cm}\includegraphics[width=0.98\textwidth]{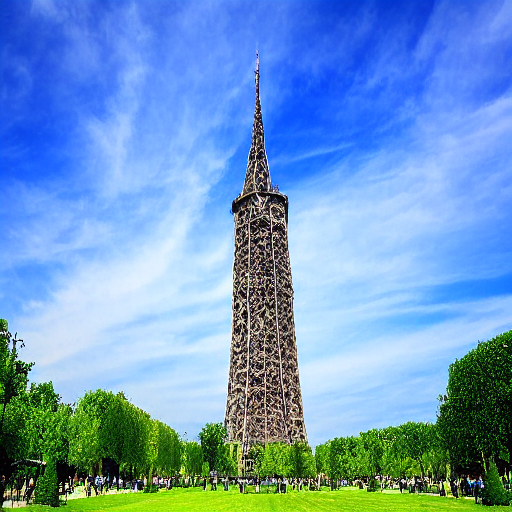} 
        \end{minipage}
        \centering
        {\footnotesize {"a \colorbox{blue}{\textcolor{white}{blue}} tower"}}
    \end{minipage}

    \vspace{0.25cm}
 \hspace{-0.0625cm}
    \begin{minipage}{0.0695\textwidth}
        \centering
         \vspace{-0.0cm}
                 \vspace{-0.45cm}
\includegraphics[width=0.95\textwidth]{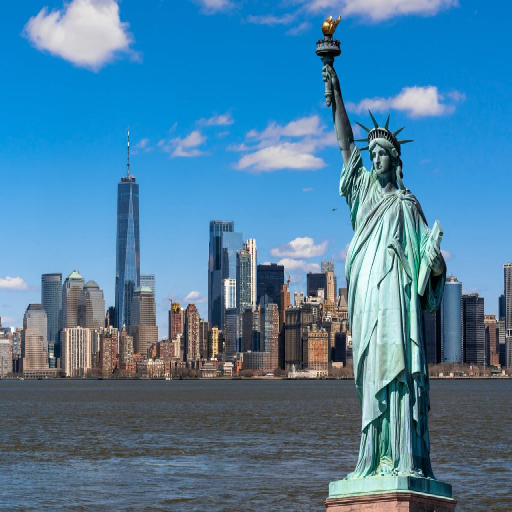} 
    \end{minipage}
    \hspace{0.05cm}
    \begin{minipage}{0.435\textwidth}
        \begin{minipage}{0.155\textwidth}
            \centering
            \includegraphics[width=0.98\textwidth]{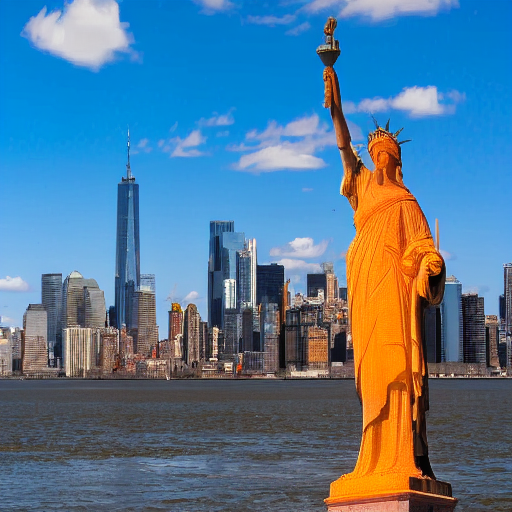} 
        \end{minipage}
                \begin{minipage}{0.155\textwidth}
        \centering
            \includegraphics[width=0.98\textwidth]{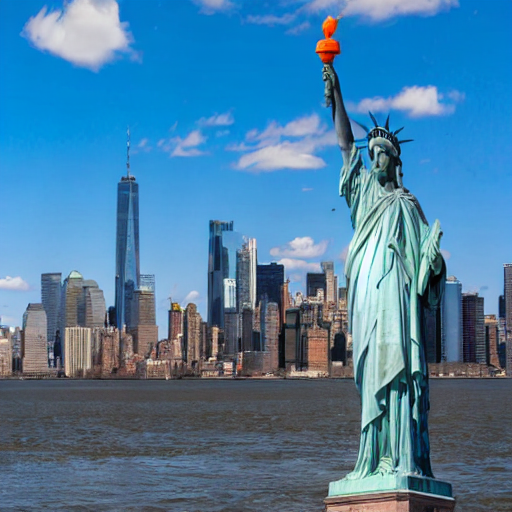} 
        \end{minipage}
                        \begin{minipage}{0.155\textwidth}
        \centering
            \includegraphics[width=0.98\textwidth, height=0.98\textwidth]{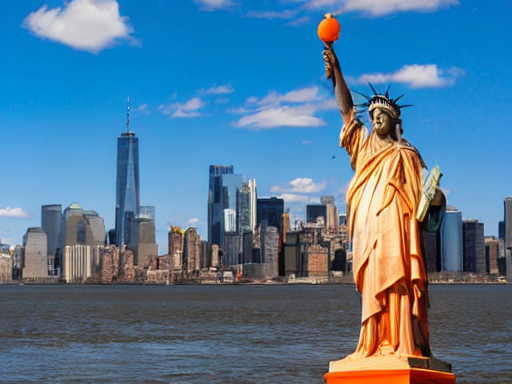} 
        \end{minipage}
                        \begin{minipage}{0.155\textwidth}
        \centering
            \includegraphics[width=0.98\textwidth, height=0.98\textwidth]{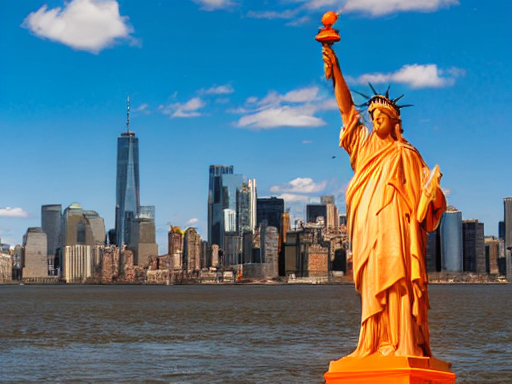} 
        \end{minipage}
        \begin{minipage}{0.155\textwidth}
            \centering
            \includegraphics[width=0.98\textwidth]{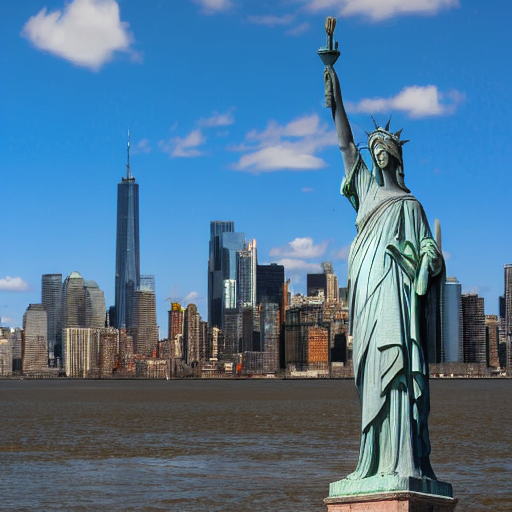} 
        \end{minipage}
        \begin{minipage}{0.155\textwidth}
            \centering
            \includegraphics[width=0.98\textwidth]{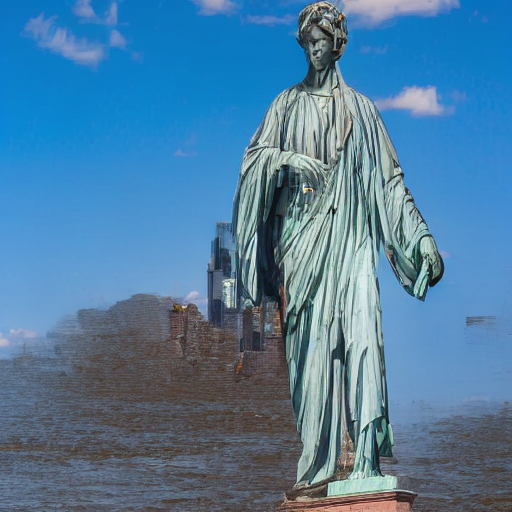} 
        \end{minipage}
        \centering
        {\footnotesize {"a \colorbox{orange}{orange} statue"}}
    \end{minipage}
    \hspace{0.05cm}
    \begin{minipage}{0.435\textwidth}
        \begin{minipage}{0.155\textwidth}
            \centering
            \includegraphics[width=0.98\textwidth]{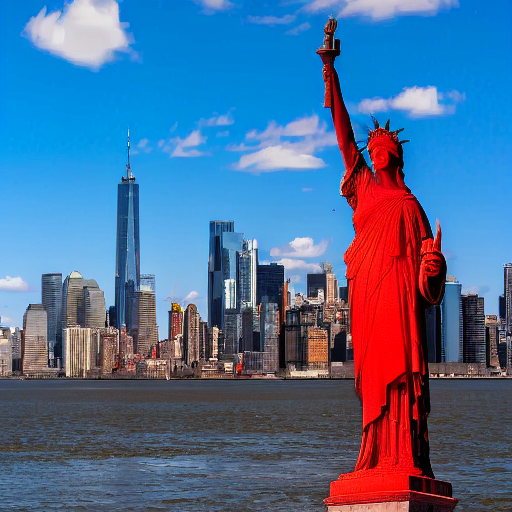} 
        \end{minipage}
        \begin{minipage}{0.155\textwidth}
            \centering
            \includegraphics[width=0.98\textwidth]{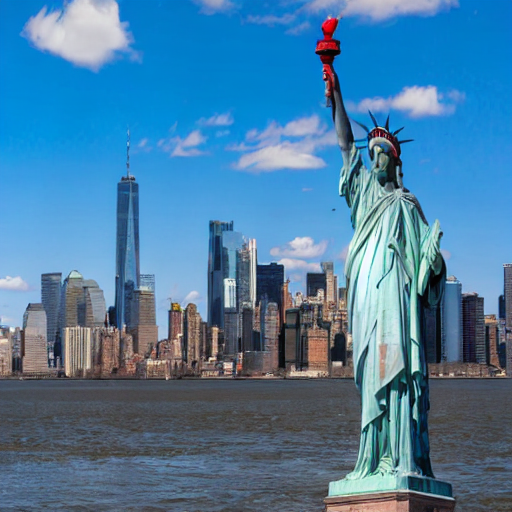} 
        \end{minipage}
                \begin{minipage}{0.155\textwidth}
            \centering
            \includegraphics[width=0.98\textwidth, height=0.98\textwidth]{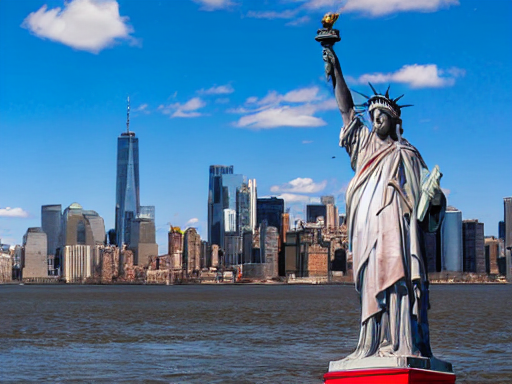} 
        \end{minipage}
                \begin{minipage}{0.155\textwidth}
            \centering
            \includegraphics[width=0.98\textwidth, height=0.98\textwidth]{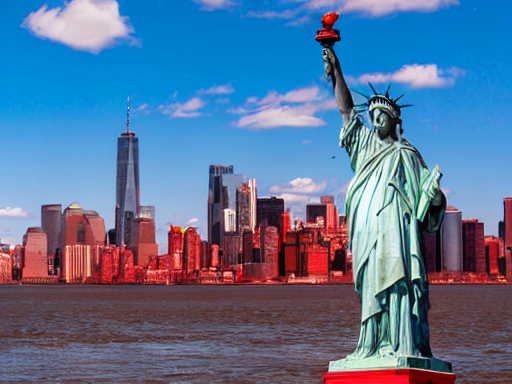} 
        \end{minipage}
        \begin{minipage}{0.155\textwidth}
            \centering
            \includegraphics[width=0.98\textwidth]{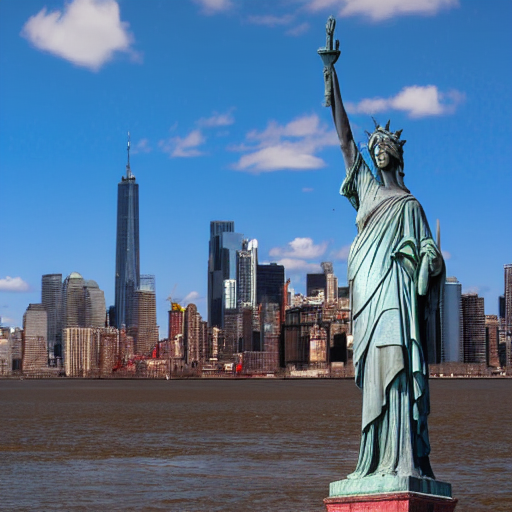} 
        \end{minipage}
        \begin{minipage}{0.155\textwidth}
            \centering
            \includegraphics[width=0.98\textwidth]{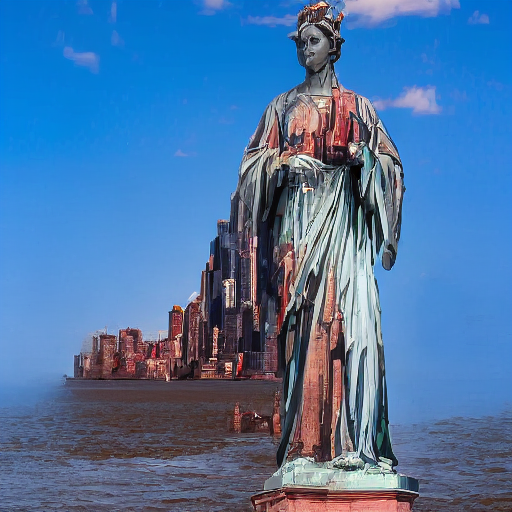} 
        \end{minipage}
        \centering
        {\footnotesize {"a \colorbox{red}{red} statue"}}
    \end{minipage}
    
    \vspace{0.25cm}

    \hspace{-0.05cm}
    \begin{minipage}{0.0695\textwidth}
        \centering
         \vspace{-0.0cm}
                  \vspace{-0.45cm}
         \includegraphics[width=0.95\textwidth]{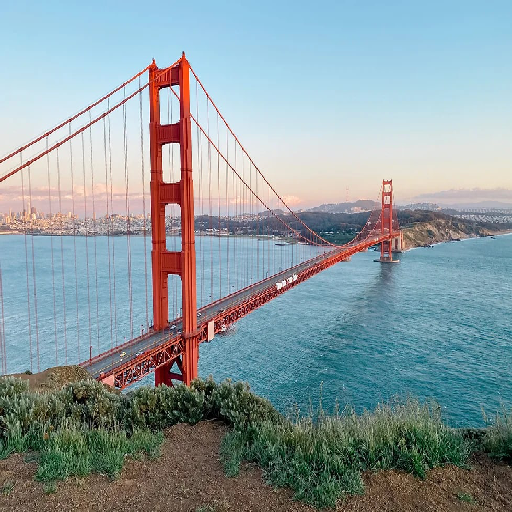} 
    \end{minipage}
    \hspace{0.03cm}
    \begin{minipage}{0.435\textwidth}
        \begin{minipage}{0.155\textwidth}
            \centering
            \vspace{0.02cm}\includegraphics[width=0.98\textwidth]{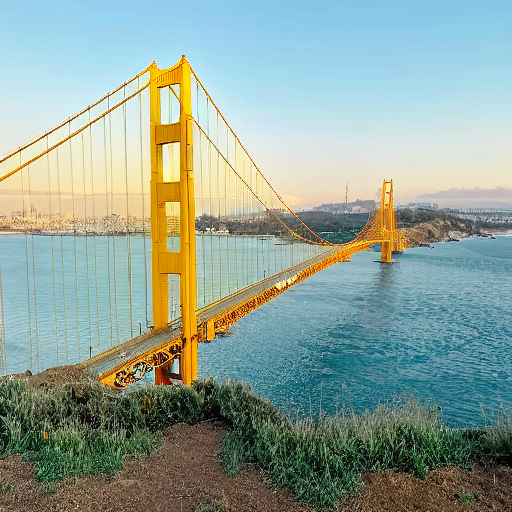} 
        \end{minipage}
        \begin{minipage}{0.155\textwidth}
            \centering
            \includegraphics[width=0.98\textwidth]{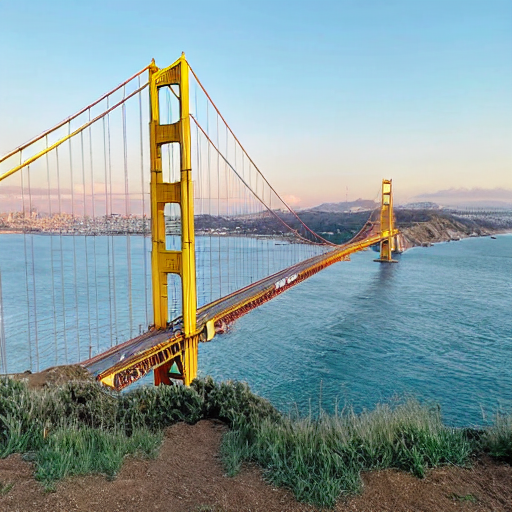} 
        \end{minipage}
                \begin{minipage}{0.155\textwidth}
            \centering
            \includegraphics[width=0.98\textwidth, height=0.98\textwidth]{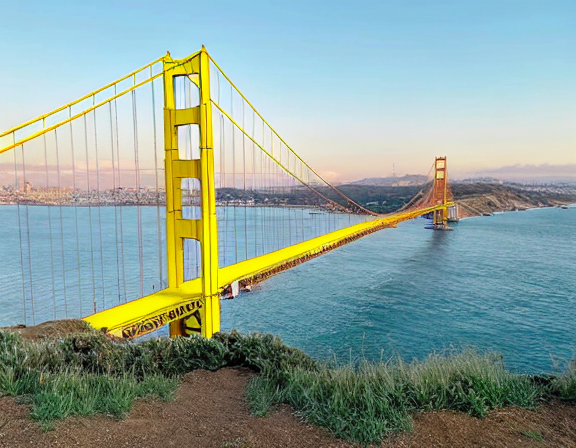} 
        \end{minipage}
                \begin{minipage}{0.155\textwidth}
            \centering
            \includegraphics[width=0.98\textwidth, height=0.98\textwidth]{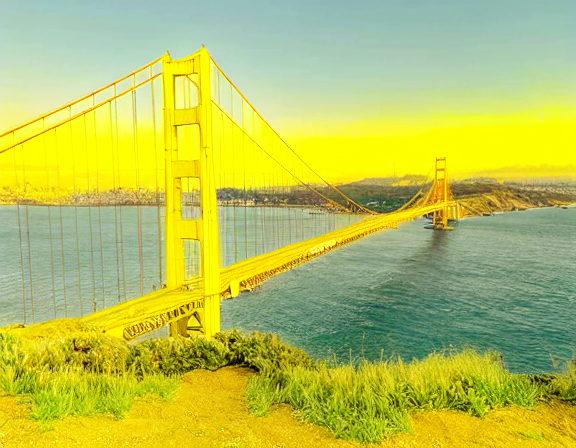} 
        \end{minipage}
        \begin{minipage}{0.155\textwidth}
            \centering
            \includegraphics[width=0.98\textwidth]{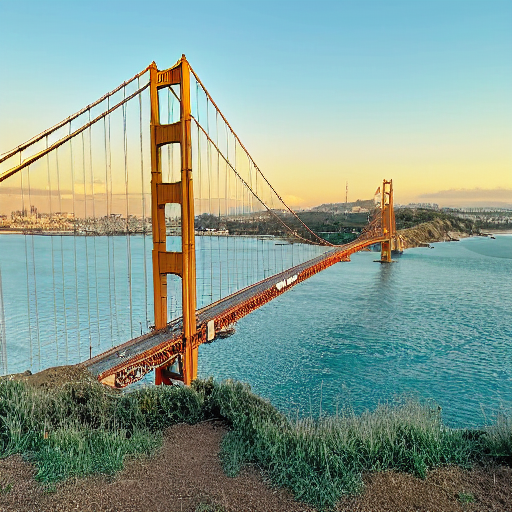} 
        \end{minipage}
        \begin{minipage}{0.155\textwidth}
            \centering
            \includegraphics[width=0.98\textwidth]{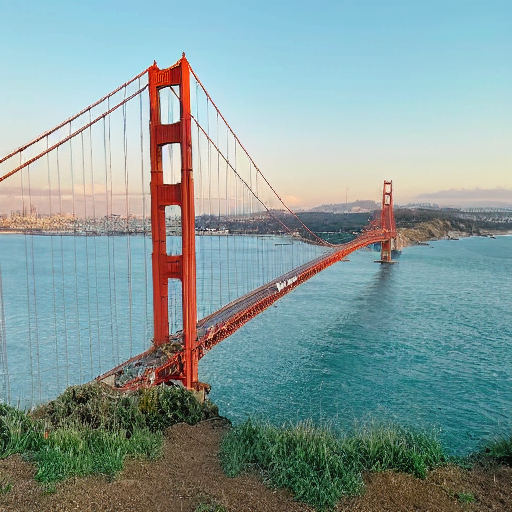} 
        \end{minipage}
        \centering
        {\footnotesize {"a \colorbox{yellow}{yellow} bridge"}}
    \end{minipage}
    \hspace{0.05cm}
    \begin{minipage}{0.435\textwidth}
        \begin{minipage}{0.155\textwidth}
            \centering
            \includegraphics[width=0.98\textwidth]{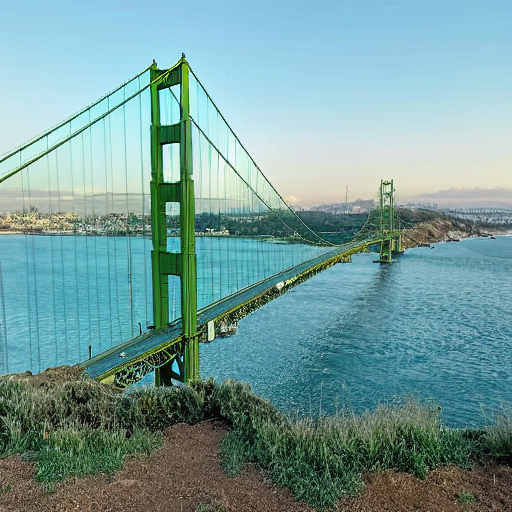} 
        \end{minipage}
        \begin{minipage}{0.155\textwidth}
            \centering
            \includegraphics[width=0.98\textwidth]{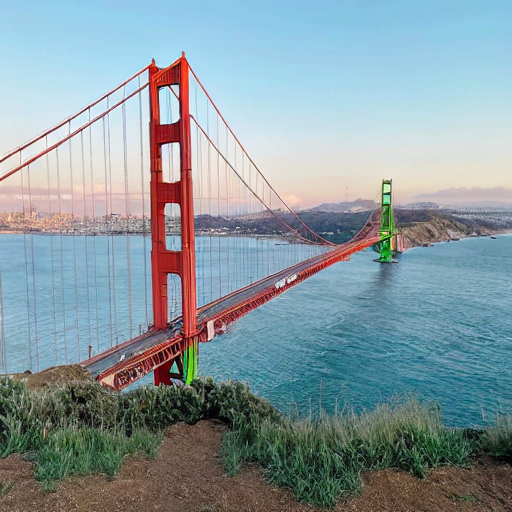} 
        \end{minipage}
        \begin{minipage}{0.155\textwidth}
            \centering
            \includegraphics[width=0.98\textwidth, height=0.98\textwidth]{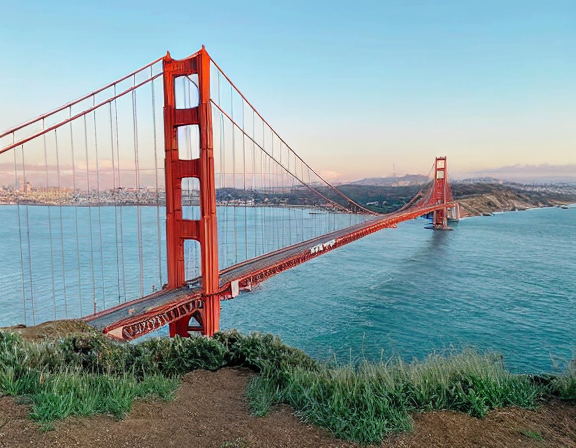} 
        \end{minipage}
        \begin{minipage}{0.155\textwidth}
            \centering
            \includegraphics[width=0.98\textwidth, height=0.98\textwidth]{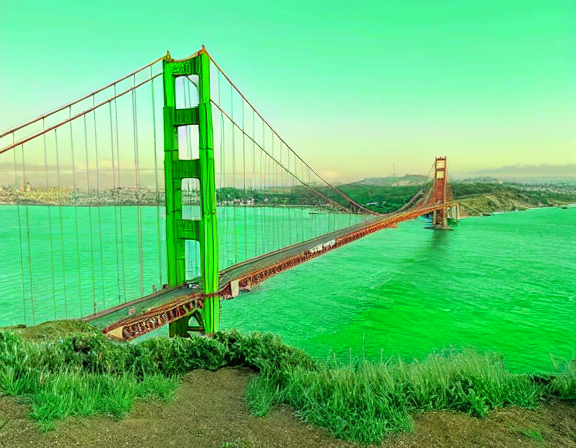} 
        \end{minipage}
        \begin{minipage}{0.155\textwidth}
            \centering
            \includegraphics[width=0.98\textwidth]{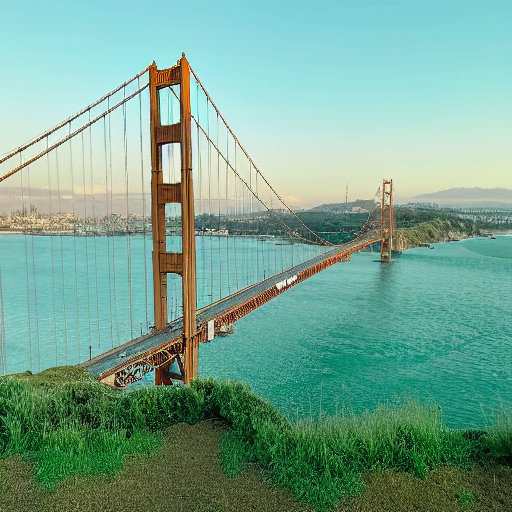} 
        \end{minipage}
        \begin{minipage}{0.155\textwidth}
            \centering
            \includegraphics[width=0.98\textwidth]{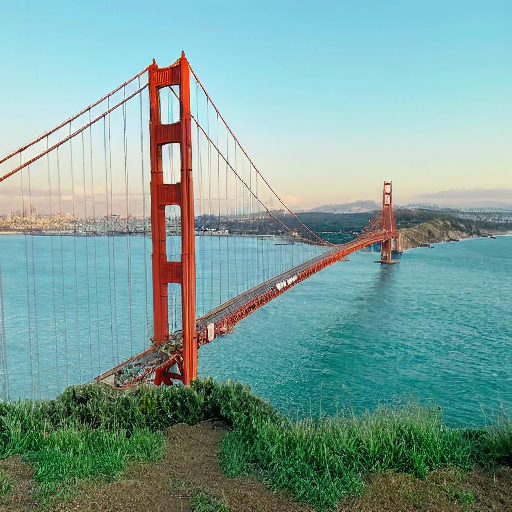} 
        \end{minipage}
        \centering
        {\footnotesize {"a \colorbox{green128}{green} bridge"}}
    \end{minipage}

     \caption{Above, we show edits obtained using our method ColorEdit, AnySD, MagicBrush, InstructPix2Pix, FPE and MasaCtrl over real images, demonstrating that our method can modify the colors of objects in real images, outperforming baseline methods significantly.}
    \label{fig:real_images}
\end{figure*}

%% file: rebuttal_files/banana/banana.tex
\definecolor{green128}{rgb}{0.0,0.5,0.0}
\definecolor{purple128}{rgb}{0.5,0.0,0.5}

\begin{figure}[t]
     \centering
    \begin{minipage}{0.155\columnwidth}
        \centering
        \includegraphics[width=0.98\columnwidth]{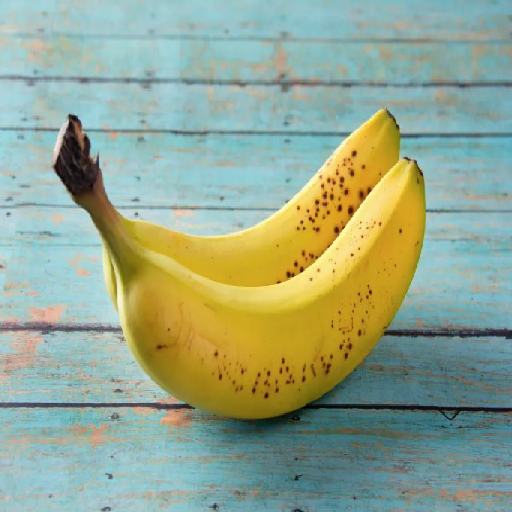} 
        {Source}
    \end{minipage}
    \hspace{0.01cm}
    \begin{minipage}{0.155\columnwidth}
        \centering
        \vspace{+0.07cm}
        \includegraphics[width=0.98\columnwidth]{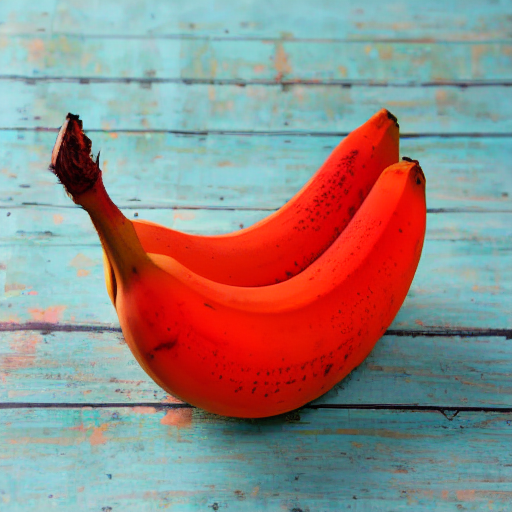}  %
        \small{\colorbox{red}{\textcolor{white}{Red}}}
    \end{minipage}
    \begin{minipage}{0.155\columnwidth}
        \centering
        \vspace{+0.07cm}
        \includegraphics[width=0.98\columnwidth]{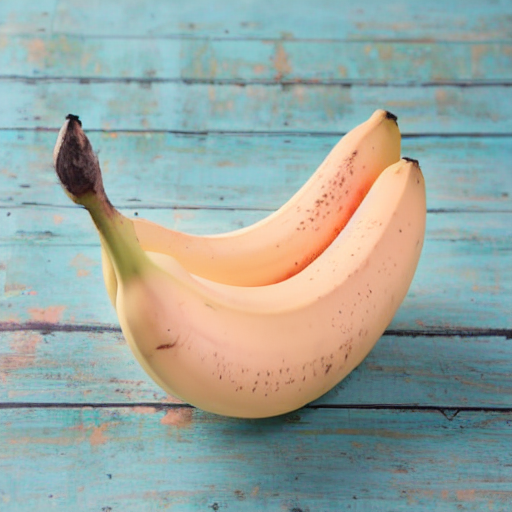}   %
        \small{\colorbox{pink}{\textcolor{white}{Pink}}}
    \end{minipage}
    \begin{minipage}{0.155\columnwidth}
        \centering
        \vspace{+0.07cm}
        \includegraphics[width=0.98\columnwidth]{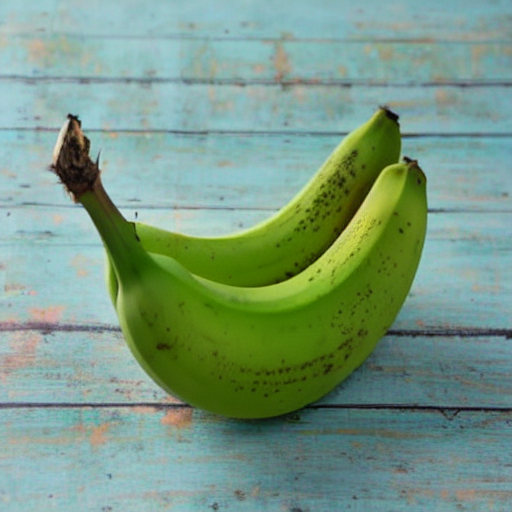}   %
        \small{\colorbox{green128}{\textcolor{white}{Green}}}
    \end{minipage}
    \begin{minipage}{0.155\columnwidth}
        \centering
        \vspace{+0.14cm}
        \includegraphics[width=0.98\columnwidth]{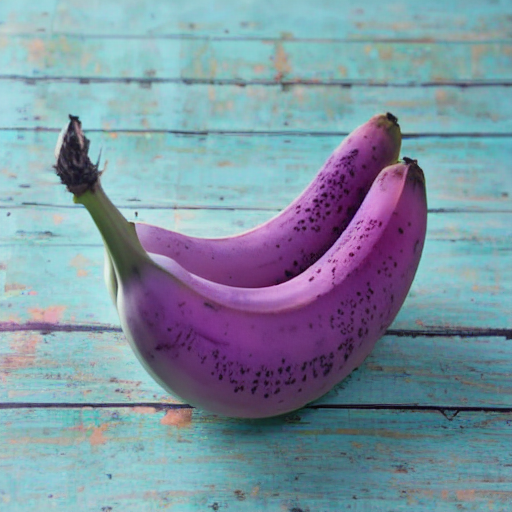}   %
        \small{\colorbox{purple128}{\textcolor{White}{Purple}}}
    \end{minipage}
    \begin{minipage}{0.155\columnwidth}
        \centering
        \vspace{+0.07cm}
        \includegraphics[width=0.98\columnwidth]{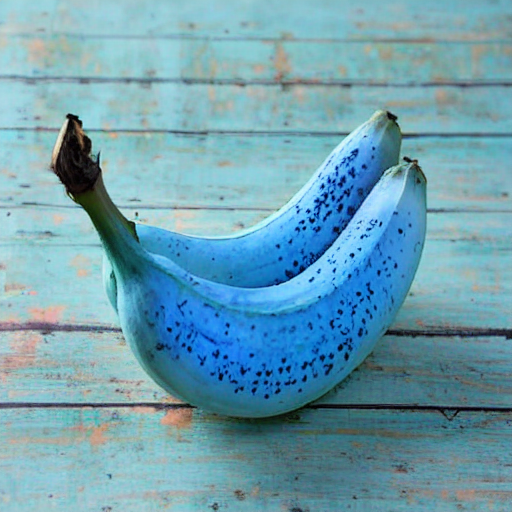}   %
        \small{\colorbox{blue}{\textcolor{white}{Blue}}}
    \end{minipage}
    \caption{Real image color editing of dominant intrinsic color object (i.e., banana is usually yellow).}
    \label{fig:banana}
\end{figure}

%% file: figures/sam_fail/sam_fail.tex
\begin{figure}[t]
     \centering
    \begin{minipage}{0.32\columnwidth}
        \centering
        {SD 2.1}
        \includegraphics[width=0.98\columnwidth]{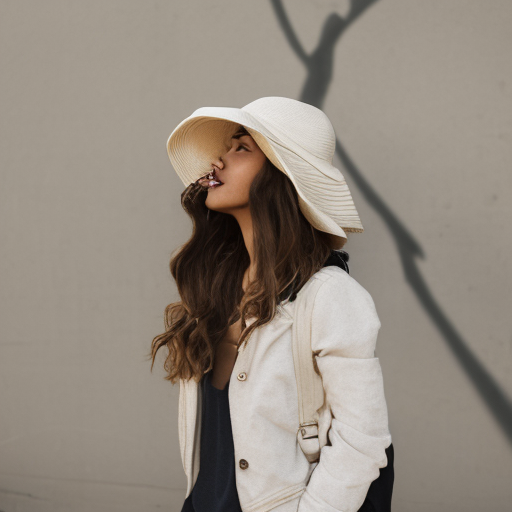} 
    \end{minipage}
    \begin{minipage}{0.32\columnwidth}
        \centering
        {Backpack Seg.}
        \includegraphics[width=0.98\columnwidth]{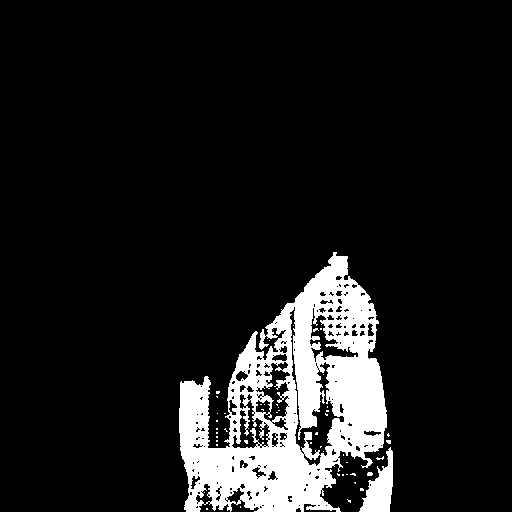} 
    \end{minipage}
    \begin{minipage}{0.32\columnwidth}
        \centering
        {SD 2.1 Edited}
        \includegraphics[width=0.98\columnwidth]{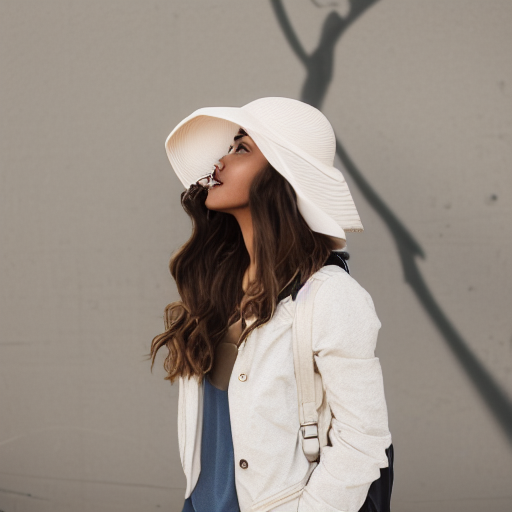}  %
    \end{minipage}
    \vspace{0.1cm}
    \hspace{0.1cm}
    {\footnotesize {"a \colorbox{blanchedalmond}{blanched-almond} colored hat and a \colorbox{white}{white} colored backpack"}} %
    \caption{Example of a failure case, due to an erroneous segmentation map. The backpack segmentation includes the woman's jacket and shirt, resulting in these regions being edited as well.}
    \hspace{0.3cm}
    \label{fig:sam_fail}
\end{figure}

%% file: figures/sd_fail/sd_fail.tex
\begin{figure}[t]
     \centering
    \begin{minipage}{0.48\columnwidth}
        \centering
        {FLUX}
        \includegraphics[width=0.98\columnwidth]{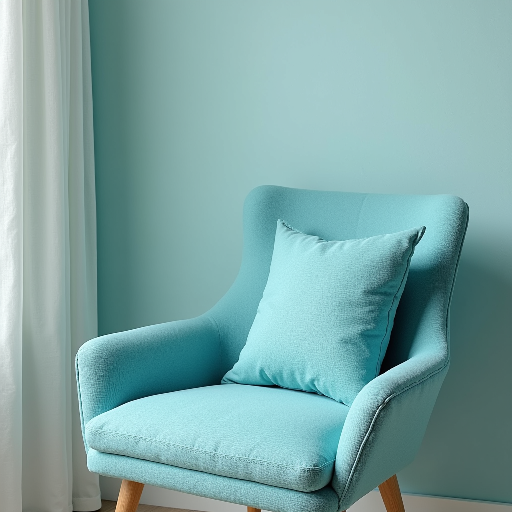} 
    \end{minipage}
    \begin{minipage}{0.48\columnwidth}
        \centering
        {+Ours}
        \includegraphics[width=0.98\columnwidth]{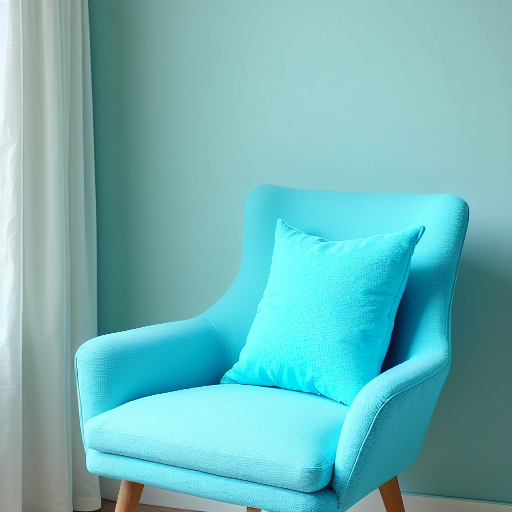}  %
    \end{minipage}
    \vspace{0.3cm}
    \begin{minipage}{0.7\columnwidth}
        \centering
            \vspace{0.3cm}
                {$P_{simp}$ Maps}
        \includegraphics[width=0.98\columnwidth]{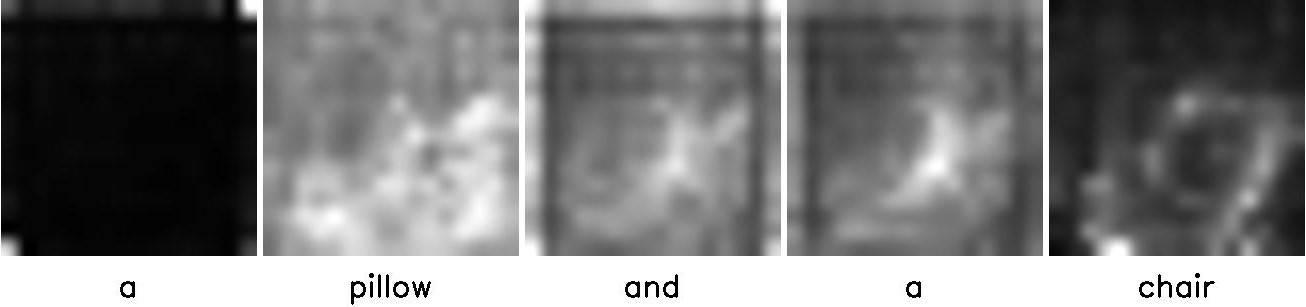} 

    \end{minipage}
    \vspace{0.3cm}
    \begin{minipage}{0.98\columnwidth}
        \centering
            \vspace{-0.1cm}
                {$P_{full}$ Maps}
        \includegraphics[width=0.98\columnwidth]{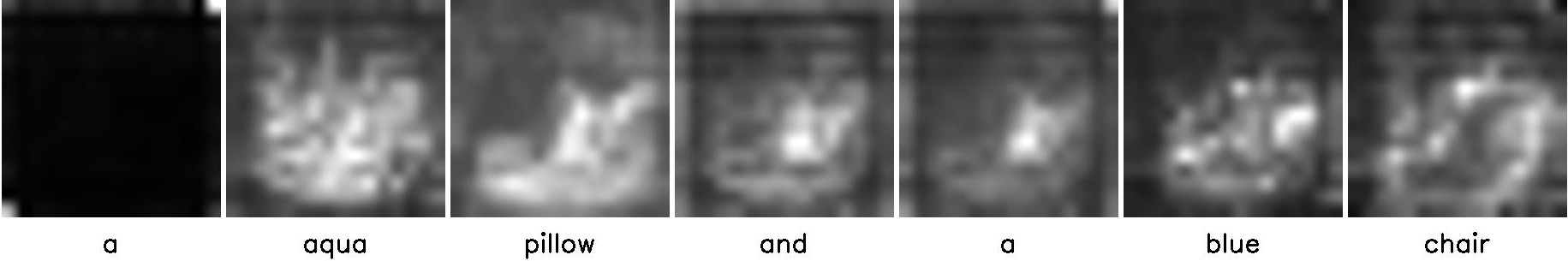} 

    \end{minipage}
    \hspace{0.1cm}
    {\footnotesize {"a \colorbox{aqua}{aqua} colored pillow and a \colorbox{skyblue}{sky-blue} colored chair"}} %
    \caption{An example of noisy cross attention using simplified and full prompts of semantically similar objects (e.g. a pillow and a chair), this results suboptimal performance as both objects are colored with the aqua color.}
    \hspace{0.3cm}
    \label{fig:sd_fail}
\end{figure}